\documentclass[10pt,journal,compsoc]{IEEEtran}
\newif\ifpeerreview

\peerreviewfalse
\usepackage[table]{xcolor}
\input{mlpreamble}
\usepackage[nocompress]{cite}
\usepackage{url}
\usepackage{amsmath,amssymb,graphicx}
\usepackage{amsthm}
\usepackage{times}
\usepackage{epsfig}
\usepackage{graphicx}
\usepackage{amsmath}
\usepackage{amssymb}
\usepackage{booktabs}
\usepackage{here}
\usepackage{bm}
\usepackage{url}
\usepackage[subrefformat=parens]{subcaption}
\usepackage{adjustbox}
\usepackage{algorithm}
\usepackage[noend]{algpseudocode}
\usepackage{multirow}
\usepackage{siunitx}
\usepackage{mathtools}
\usepackage{wrapfig}
\usepackage{empheq}
\usepackage[whole]{bxcjkjatype}
\usepackage[switch]{lineno}
\usepackage{arydshln}
\usepackage{colortbl}
\usepackage{makecell}
\usepackage{pifont}
\usepackage{tabularx} 
\usepackage{wrapfig} 
\usepackage{xcolor}
\usepackage{color, colortbl}
\usepackage[colorlinks]{hyperref} 
\usepackage{diagbox}
\usepackage{cleveref} 
\usepackage{xcolor}
\usepackage{color, colortbl}
\definecolor{Gray}{gray}{0.85}
\definecolor{LightGray}{gray}{0.95}
\definecolor{LightRed}{rgb}{0.98,0.859,0.85}
\definecolor{LightYellow}{rgb}{0.99,0.95,0.81}
\definecolor{LightOrange}{rgb}{0.99,0.92,0.816}
\definecolor{DarkRed}{rgb}{0.8,0,0}
\definecolor{LightBlue}{rgb}{0,0.44,0.753}

\def\onedot{.~}
\def\eg{\emph{e.g.},~} 
\def\ie{\emph{i.e.},~} 
 
\def\st{s.t\onedot}

\def\iow{i.o.w.,~}
\def\etal{\emph{et al}\onedot}

\newcommand{\normal}{\V{n}}
\newcommand{\m}{\ensuremath{m}}

\newcommand{\vs}{\V{s}}
\newcommand{\vx}{\V{x}}

\newcommand{\point}{\vx}

\newcommand{\vunknown}{\V{e}}

\newcommand{\light}[2]{\vs^{#2}_{#1}}     
\newcommand{\dist}[2]{e^{#2}_{#1}}     
\newcommand{\im}[2]{m^{#2}_{#1}}     
\newcommand{\lo}{\vs_{o}}
\newcommand{\npairs}{\ensuremath{n_\mathrm{pairs}}}

\newcommand{\neqs}{\ensuremath{n_\mathrm{eqs}}}

\newcommand{\eqia}{\ensuremath{\mathrm{1A}}}
\newcommand{\eqiia}{\ensuremath{\mathrm{2A}}}
\newcommand{\eqib}{\ensuremath{\mathrm{1B}}}
\newcommand{\eqiib}{\ensuremath{\mathrm{2B}}}
\newcommand{\Veqia}{\V{A'}_{\eqia}}
\newcommand{\Veqiia}{\V{A'}_{\eqiia}}
\newcommand{\Veqib}{\V{A'}_{\eqib}}
\newcommand{\Veqiib}{\V{A'}_{\eqiib}}

\newcommand{\pparag}[1]{\vspace{3mm}\noindent\textbf{#1:}}

\newtheorem{condition}{Condition}

\newtheorem{proposition}{Proposition}

\newcommand{\colorbar}[3]{
\begin{tabular}[t]{@{}l@{}l@{}}
	\includegraphics[height=#1\linewidth,width=0.5em]{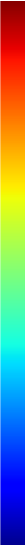} & 
	\begin{tabular}[b]{@{}l}
		#2 \vspace{#3\linewidth}\\
		$0$
        \vspace{0.5pt}
	\end{tabular}
\end{tabular}
}
\newcommand{\rotatedColorbar}[4]{
	\centering
	\begin{tabular}[t]{@{}l@{}c@{}l@{}}
		& #1 & \\
		& \rotatebox[origin=c]{270}{\includegraphics[height=#2\linewidth,width=0.5em]{figures/colorbar.pdf}} & 
		\vspace{#4pt}
		\\
		$0$ & & #3 \\
	\end{tabular}
}
\newcommand{\raisedtarget}[1]{%
  \raisebox{\fontcharht\font}[0pt][0pt]{\hypertarget{#1}{}}%
}

\newcommand{\paperID}{38}

\title{Near-light Photometric Stereo
\\ with Symmetric Lights}
\author{
Lilika~Makabe,~\IEEEmembership{Student Member,~IEEE,}
Heng~Guo,
Hiroaki~Santo,~\IEEEmembership{Member,~IEEE,}
Fumio~Okura,~\IEEEmembership{Member,~IEEE,} 
and Yasuyuki~Matsushita,~\IEEEmembership{Senior Member,~IEEE}
 \IEEEcompsocitemizethanks{\IEEEcompsocthanksitem L. Makabe, H. Guo, H. Santo, F. Okura, and Y. Matsushita are with Graduate School of Information Science and Technology, Osaka University, Japan. \protect\\
  E-mail: \{lilika.makabe, heng.guo, santo.hiroaki, okura, yasumat\}@ist.osaka-u.ac.jp
 }
}

\begin{document}

\IEEEtitleabstractindextext{%
\begin{abstract}
This paper describes a linear solution method for near-light photometric stereo by exploiting symmetric light source arrangements. Unlike conventional non-convex optimization approaches, by arranging multiple sets of symmetric nearby light source pairs, our method derives a closed-form solution for surface normal and depth without requiring initialization. In addition, our method works as long as the light sources are symmetrically distributed about an arbitrary point even when the entire spatial offset is uncalibrated. Experiments showcase the accuracy of shape recovery accuracy of our method, achieving comparable results to the state-of-the-art calibrated near-light photometric stereo method while significantly reducing requirements of careful depth initialization and light calibration.
\end{abstract}

}

\maketitle
\thispagestyle{empty}

\IEEEraisesectionheading{
  \section{Introduction}\label{sec:introduction}
}

\IEEEPARstart{P}{hotometric} stereo aims to estimate surface normals from images captured under varying lighting conditions. Most conventional settings employ the distant light source assumption, where point light sources are infinitely far away. Under a Lambertian assumption, this simplification allows a linear image formation model, leading to a closed-form solution for surface normals given three or more images taken under different lighting conditions~\cite{woodham1980}.
In practice, light sources are located near the target scene, and the distant light source assumption breaks down, resulting in a \emph{near-light} photometric stereo problem. Near-light photometric stereo remains challenging due to the variation of incident light directions and light fall-off effect, both depending on the scene point location. Even with the Lambertian reflectance assumption, the problem becomes non-linear since the surface normals and surface point locations need to be jointly determined.

\begin{figure}[t]
    \centering
    \includegraphics[width=\linewidth]{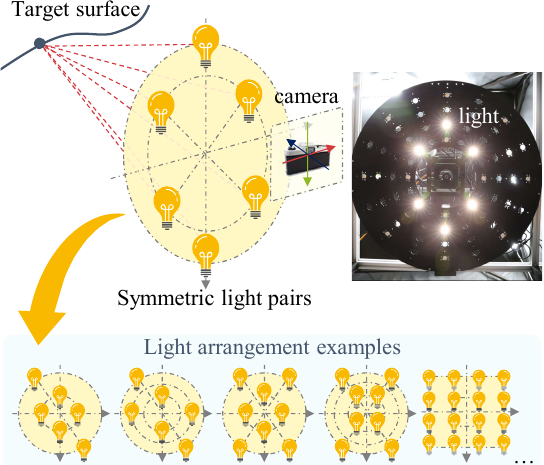}
    \caption{Examples of light source arrangement for our method. At least three pairs of symmetric point light sources are arranged to the image plane with a shared center. Our near-light photometric stereo achieves global minima in shape recovery using the proposed symmetric near-light configuration.}
    \label{fig:figure1}
    \vspace*{-20pt}
\end{figure}

\begin{table*}[t]
	\caption{Comparison of photometric stereo methods under Lambertian reflectance. Our method assumes a more practical light model, requires less light calibration effort, and provides convex and linear optimization to achieve the global optima of surface normal and depth.}
	\centering
	\resizebox{\textwidth}{!}{
		\begin{tabular}{cccccl}
			\toprule
			Light model & Method & Light calibration & Light arrangement & Optimization & Additional requirements \\
			\midrule
			\multirow{4}{*}{Distant} & \cite{woodham1980} & Calibrated & N.A & Convex & N.A\\
			& \cite{Minami2022} & Partially calibrated & Axis-aligned symmetric & Convex & Constant albedo \\
			& \cite{Zhou2010}, \cite{Chandraker2012, Shiradkar2014}& Partially calibrated & Ring & Convex & Integrability \\
			& \cite{favaro2012closed}& Uncalibrated & N.A. & Convex &  Integrability, diffuse maxima detection \\
			\midrule
			\multirow{5}{*}{Near} & \cite{Yvain2018}, \cite{mecca2014near, mecca2015realistic, mecca2016unifying}  & Calibrated & N.A & Non-convex & Depth initialization\\
			& \cite{Liu2018}& Calibrated & Ring & Non-convex &  Depth initialization \\
			& \cite{Sakue2011}& Calibrated & N.A. & Non-convex &  Relaxation of light fall-off \\
			& \cite{Papadhimitri2014}& Uncalibrated & N.A. & Non-convex &  Depth initialization, normal integration \\
            & \cite{lichy2022fast}& Uncalibrated & N.A. & Non-convex &  Data prior \\
			\rowcolor[gray]{0.8}
			& Ours & Partially calibrated & Symmetric pairs & Convex/Closed-form &  Relaxation of light fall-off \\
			\midrule
			Universal & \cite{ikehata2022universal}  & Uncalibrated & N.A. & Non-convex &  Data prior \\
			\bottomrule
		\end{tabular}
	}
	\label{tab:method_comparison}
    \vspace*{-10pt}
\end{table*}

Prior approaches to near-light photometric stereo rely on non-convex optimization due to the non-linear nature of the problem~\cite{Logothetis2020cnn,lichy2022fast, Yvain2018,mecca2014near,mecca2015realistic,mecca2016unifying}. These methods commonly require the initial guess of the scene point locations or a depth map with a calibrated camera. While these methods yield plausible surface shape recovery, the result is naturally susceptible to the initial guess because of the non-convexity. Liu~\etal~\cite{Liu2018} proposes a ring-light setting for near-light photometric stereo to obtain a reliable initial guess of the scene point locations. However, even with the improved initial guess, the issue of being affected by the initial guess still remains. Thus, it is wanted a stable near-light photometric stereo method that is unaffected by the initial guess.

This paper shows that a symmetric configuration of nearby point light sources yields novel constraints for making the problem tractable. Specifically, by positioning light sources on a plane that is perpendicular to the optical axis of the camera and symmetrically around an arbitrary central point on the plane, as shown in~\fref{fig:figure1}, we derive a set of linear constraints for determining surface normals and scene point locations based on the differences of the symmetric image measurements. More specifically, we show that three pairs of symmetric point lights, with at least one pair having a different radius than others (example configurations shown in \fref{fig:figure1} bottom), yield the constraints that are useful for uniquely reaching the optimal solution. Together with a linear relaxation of light fall-off effects~\cite{Sakue2011}, we develop a linear solution method for surface normals, scene point locations, and surface albedos, which is free from the initial guess.
In addition, with our method, the light sources are only required to be partially calibrated, \ie their absolute positions are unneeded, but only the ratios of radii of symmetric light source pairs and angles between them are assumed to be known while the offset from the camera's optical center can be unknown. 
One of our application scenarios is taking the devices from place to place for recovering different target surfaces, which requires frequent assembly and disassembly of the capture setup. Despite maintaining the relative position of lights, inevitable shifts in the absolute positions can occur, requiring careful light calibration. Our method bypasses the re-calibration step by using a symmetric light setup, which can be widely found in capture devices in previous works.

To summarize, this paper contributes to near-light photometric stereo by introducing a linear closed-form solution method that achieves global minima in surface shape recovery under the light fall-off relaxation, utilizing symmetric point light pairs. Furthermore, our method reduces the calibration effort of near light position, as it only requires the radii ratios and angles of the symmetric light source pairs, rather than precise point light positions.

\section{Related work}
Our method is related to near-light Lambertian photometric stereo methods and also photometric stereo methods that exploit specific light arrangements. We provide a brief summary of related works in these two categories, as summarized in \Tref{tab:method_comparison}. For non-Lambertian photometric stereo methods, we recommend referring to recent surveys and benchmark evaluations, such as DiLiGenT~\cite{DiLiGent} and DiLiGenT$10^2$~\cite{ren2022diligent102} for distant light configurations, and LUCES~\cite{mecca2021luces} for near-light configurations.

\subsection{Near-light photometric stereo}
Near-light photometric stereo has gained interest since Iwahori~\etal\cite{iwahori1990reconstructing} 
due to its practical and accurate modeling of nearby point lights compared to the conventional distant-light photometric stereo~\cite{woodham1980}.
However, under a near-light condition, different surface points are illuminated by different light directions and attenuation~(\ie light fall-off) depending on the relative positions between point lights to the surface points, leading to a non-linear relationship between the image observations and the surface normals.
Due to this non-linearity, the problem of near-light photometric stereo is inherently non-convex, and there is no closed-form solution.

Under calibrated light settings, variational methods~\cite{Yvain2018, mecca2014near, mecca2015realistic, mecca2016unifying} formulate the problem as non-linear PDEs based on image ratios~(ratio of two images under different illuminations) and iteratively update depth and its partial derivatives. Alternative approaches involving optimization of surface normals and depth are widely applied in \cite{ahmad2014improved,bony2013tridimensional,collins20123d,huang2015near,nie2016novel} as well as an uncalibrated near-light photometric stereo method~\cite{Papadhimitri2014}. Surface normals are first calculated from image observations with depth initialization, and depth is then recovered through normal integration~\cite{queau2018normal}. Due to the non-convex nature of near-light photometric stereo, these iterative optimization pipelines typically require a careful depth initialization. This depth initialization is also required in the most recent learning-based near-light photometric stereo methods~\cite{lichy2022fast,logothetis2022cnn, santo2020near,ikehata2022universal} addressing non-Lambertian reflectance.

To avoid depth initialization in non-convex optimization, Sakaue and Sato~\cite{Sakue2011} introduce light fall-off relaxation and derive a linear solution for near-light photometric stereo. However, their method treats surface normal and depth as independent variables, leading to inconsistent normal and depth map estimates. 
Our method differs from previous non-convex and light calibration-reliant approaches, providing a linear solution for near-light photometric stereo using partially-calibrated symmetric near lights and light fall-off relaxation~\cite{Sakue2011}. Our approach ensures that the estimated surface normal and depth achieve global optima under light fall-off relaxation.

\subsection{Photometric stereo with specific light arrangement}

Reducing the effort of light calibration is highly desirable in photometric stereo. While existing uncalibrated photometric stereo methods, such as Favaro and Papadhimitri~\cite{favaro2012closed}, do not require known light directions, they suffer from ambiguity in their estimation. As a result, additional constraints like surface integrability and detection of local diffuse reflectance maxima~\cite{favaro2012closed} are necessary to resolve the ambiguity. 
Few studies~\cite{lichy2022fast,ikehata2022universal} have explored uncalibrated photometric stereo under near-light settings. They adopt a learning-based approach, which eliminates the need for light source calibration. However, these methods require a large number of lights to obtain stable estimations.

An alternative approach to minimizing geometric light calibration effort is the utilization of specific light arrangements, which is referred to as ``partially-calibrated'' in \Tref{tab:method_comparison}.

Under distant light setting, Zhou and Tan~\cite{Zhou2010}, adopt ring light to disambiguate the GBR ambiguity~\cite{Chandraker2005} in the normal estimates, where the light directions lie on a camera view-centered cone. 
Chandraker~\etal\cite{Chandraker2012} further extends the ring light setting to differential light and achieves surface normal estimation for surfaces with isotropic reflectances.
Minami~\etal\cite{Minami2022} proposes to use a simplified version of the ring light, where four directional lights are fixed in vertical and horizontal symmetry (\ie axis-aligned), which is a special case of the symmetric light arrangement. 
The observations obtained under symmetric lights provide azimuth angles of surface normals without assuming a given radius for the lights. To uniquely determine the full surface normal, their approach requires at least two scene points with constant albedo.

Few methods explore light arrangement in near-light photometric stereo. Liu~\etal\cite{Liu2018} utilize a ring point light configuration to obtain a reasonable depth initialization for their non-linear photometric stereo optimization; however, light calibration for these ring point lights remains necessary. Our approach introduces a symmetric near-light arrangement, offering more flexible configurations, as shown in \fref{fig:figure1}, compared to the distant and axis-aligned symmetric light arrangement~\cite{Minami2022}. By employing the proposed symmetric light pairs, we not only eliminate the need for geometric light calibration but also enable the resolution of near-light photometric stereo using linear and convex optimization.


\section{Problem statement}
We start by presenting the image formation model under a near-light setting. We then outline our approach for estimating the scene shape and albedos using the proposed symmetric near point light pairs.

\label{sec:image_formation_model}
Assuming Lambertian reflectance and the equal light radiance for all the lights, the intensity measurement $m \in \mathbb{R}_+$ of a surface point $\vx \in \mathbb{R}^3$ illuminated by a point light source can be described as:
\begin{align}
    \m &= \rho \frac{1}{\norm{\vs - \vx}_2^2} \frac{\left(\vs - \vx\right)\transp\normal}{\norm{\vs - \vx}_2},
    \label{eq:image_formation}
\end{align}
where $\rho \in \doubleR_{+}$ is the albedo of the surface point, which is scaled by the strength of the light source, $\vs \in \doubleR^3$ is the point light source position, and $\normal \in \doubleS^2 \subset \doubleR^3$ is a surface normal.
The factor $\frac{1}{\norm{\vs - \vx}_2^2}$ represents the light fall-off effect proportional to the squared distance from light source $\V{s}$ to surface point $\V{x}$.
For the simplicity of notations, we define $d \triangleq \norm{\vs - \vx}_2$ and $e \triangleq \rho^{-1} d^3$. The \eref{eq:image_formation} can be rewritten as
\begin{align}
    e \m = \left(\vs - \vx\right)\transp\normal.
    \label{eq:image_formation_simple}
\end{align}
The scalar $e$ can be interpreted as a light-to-surface distance scaled by the albedo and light fall-off, and we hereafter call it a scaled distance.
 Given the measurements $\{m\}$ under a set of point light sources $\{\V{s}\}$, our goal is to determine the surface normal $\V{n}$, surface point $\V{x}$, and albedo $\rho$ at each surface point.

\begin{figure}[t]
	\centering
	\includegraphics[width=\linewidth]{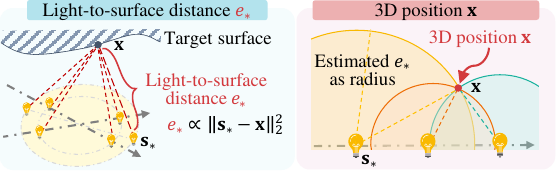}
	\caption{Overview of the proposed method.
		We consider the differences and summations of observations under the symmetric light pairs and construct a linear optimization system to estimate light-to-surface (scaled) distance $e_*$. From the scaled distance $e_*$, we compute the surface position (up to scale if radii are unknown) and normal.}
	\label{fig:overview_1}
\end{figure}

\section{Proposed method}
 As shown in \fref{fig:overview_1}, instead of directly solving for all the unknowns, our strategy is to first determine the scaled distance $e$ by exploiting the constraints derived by our symmetric light configuration. Once the scaled distance $e$ is obtained in a per-pixel manner, we can then solve for surface normal $\V{n}$ and surface point $\V{x}$, and albedo $\rho$ up to scale. 
   In the following, we first introduce the linear constraint derived from the symmetric light arrangement, then discuss the surface normal and depth estimation.

\subsection{Symmetric-light constraints}
\label{sec:scaled_dist_estimation1}

\begin{wrapfigure}{l}{0.5\linewidth}
	\vspace{-10pt}
	\centering
	\includegraphics[keepaspectratio,width=\linewidth]{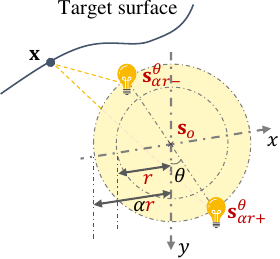} 
	\caption{\mbox{Light coordinate system.}}
	\label{fig:light_coordinate}
\end{wrapfigure}

As illustrated in \fref{fig:light_coordinate}, we have a pair of origin symmetric point lights whose 3D coordinates, $\light{r+}{\theta}$ and $\light{r-}{\theta}$, are:
\begin{equation}
    \left\{
    \begin{aligned}
        \light{r+}{\theta} &= \left[+r\sin{\theta}, +r\cos{\theta}, 0\right]\transp + \lo \\
        \light{r-}{\theta} &= \left[-r\sin{\theta}, -r\cos{\theta}, 0\right]\transp + \lo \\
    \end{aligned}
    \right.,
\end{equation}
in which $r \in \mathbb{R}_+$ is the radius of the symmetric lights, \hbox{$\lo = \left[s_x,s_y,s_z\right]\transp$} is the offset of the set of symmetric lights.
Hereafter, we call this origin-symmetric light pair parameterized by the same radius and angle as the symmetric pair. 

The observations under the symmetric pair of lights can be written using \eref{eq:image_formation_simple} as:
\begin{equation}
    \left\{
    \begin{array}{rcl}
        \dist{r+}{\theta} \im{r+}{\theta} &=& \left(\light{r+}{\theta} - \vx\right)\transp\normal \\
        \dist{r-}{\theta} \im{r-}{\theta} &=& \left(\light{r-}{\theta} - \vx\right)\transp\normal \\
      \end{array} 
      \right., \nonumber
\end{equation}
\begin{wrapfigure}{l}{\linewidth}
	\vspace{-10pt}
	\centering
	\includegraphics[keepaspectratio,width=\linewidth]{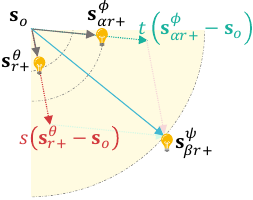} 
	\caption{\small{Relative vector relation.}}
	\label{fig:eq5}
	\vspace{-20pt}   
\end{wrapfigure}
where surface normal \hbox{$\normal = \left[n_x, n_y, n_z \right]\transp$}, surface position $\point = \left[x, y, z\right]\transp$, and scaled distances $e_{*}$ are unknown.
In a similar manner to~\cite{Minami2022}, we consider the difference of observations under symmetric pair and construct equations for unknown scaled distances $e_{*}$.
First, we consider the difference of observations under one pair to obtain the following equation:
\begin{equation}
        \dist{r+}{\theta} \im{r+}{\theta} - \dist{r-}{\theta} \im{r-}{\theta} = \left(\light{r+}{\theta} - \light{r-}{\theta}\right) \transp\normal
        = 2 \left(\light{r+}{\theta} - \lo\right) \transp\normal.\\
      \label{eq:difference}
\end{equation}

When we have at least two symmetric pair of lights, whose angles are different from each other, any other relative vectors $\light{*}{*} - \lo$ can be written as a linear combination of the two bases as:
\begin{equation}
    \begin{aligned}
        \light{\beta r+}{\psi} - \lo 
    &= s \left(\light{r+}{\theta} - \lo\right) \\
    &+ t \left(\light{\alpha r+}{\phi} - \lo\right),
    \end{aligned}
  \label{eq:relative_position_using_basis}
\end{equation} 
where the ratio of the radii $\alpha$ and $\beta$ are known in our setting, and coefficients $s,t \in \mathbb{R}$ can be parameterized by the known ratio~$\alpha:\beta$ and angles $\theta$, $\phi$, and $\psi$.
By substituting \eref{eq:relative_position_using_basis} into \eref{eq:difference}, we can obtain the following equation:
\begin{equation}
    \begin{array}{ll}
        \dist{\beta r+}{\psi} \im{\beta r+}{\psi} - \dist{\beta r-}{\psi} \im{\beta r-}{\psi} & \\ 
       = 2 \left(\light{\beta r+}{\psi} - \lo\right) \transp\normal\\ 
       = 2s \left(\light{r+}{\theta} - \lo\right)\transp\normal + 2t \left(\light{\alpha r+}{\phi} - \lo\right)\transp\normal\\ 
       = s \left(\dist{r+}{\theta} \im{r+}{\theta} - \dist{r+}{\theta} \im{r+}{\theta}\right)+ t \left(\dist{\alpha r+}{\phi} \im{\alpha r+}{\phi} - \dist{\alpha r-}{\phi} \im{\alpha r-}{\phi}\right).
    \end{array}
    \label{eq:linear_constraints1} 
\end{equation}
When we have $\npairs$ pairs of the symmetric light sources, we can obtain $\npairs - 2$ equations using two basis pairs, \eg $\left(r, \theta\right)$ and $\left(\alpha r, \phi\right)$. See the supplementary material for further detail on the number of linearly independent constraints. 

Further, by considering the summation of the symmetric pairs, 
we can obtain the following equations:
\begin{equation}
    \begin{array}{rl}
        \dist{r+}{\theta} \im{r+}{\theta} + \dist{r-}{\theta} \im{r-}{\theta}
        & = 2 \left(\lo - \point\right) \transp\normal\\
        & = \dist{\alpha r+}{\phi} \im{\alpha r+}{\phi} + \dist{\alpha r-}{\phi} \im{\alpha r-}{\phi},
      \end{array}
  \label{eq:linear_constraints2}
\end{equation}
which leads to another linear constraints on $e_{*}$. Given $\npairs$ pairs of symmetric lights, we can obtain $\npairs-1$ linear independent equations.
        
To sum up, we have two kinds of equations from the difference and summation of the symmetric pairs~(Eqs.~(\ref{eq:linear_constraints1}) and (\ref{eq:linear_constraints2})) for unknown scaled distances \hbox{$\vunknown = \left[ e_{*} \right]\transp \in \doubleR_+^{2\npairs}$}. 
Let \hbox{$\V{A} \in \mathbb{R}^{\neqs \times 2\npairs}$} be a coefficient matrix formed by putting together Eqs.~(\ref{eq:linear_constraints1}) and (\ref{eq:linear_constraints2}), where $\neqs$ is the total number of equations. 
If we can solve the homogeneous system,
\begin{equation}
\V{A} \V{e} = \V{0},
\end{equation} 
for the scaled distance vector $\V{e}$, it can be converted to the surface position $\V{x}$ as will be described in \sref{para:relative_depth_estimation}.
Unfortunately though, the right null space of the matrix $\V{A}$ is three-dimensional, \ie $\rank{\V{A}} = 2\npairs - 3$ in the general case.
 To determine the scaled distance vector $\V{e}$ up to scale, it requires two more independent constraints so that $\rank{\V{A}} = 2\npairs - 1$.

\subsection{Linear constraints by light fall-off relaxation}
\label{sec:scaled_dist_estimation2}
To derive additional constraints, we approximate the light fall-off proportional to the distance, instead of using the squared distance. Existing studies~\cite{Sakue2011,Papadhimitri2014} use the same approximation and show that its merit in alleviating difficult nonlinearity overtakes the effect of the approximation error. 
With the relaxation, the scaled distance $e$ is approximated as
\begin{align}
	e = \rho^{-1} d^3 \approx \rho^{-1} d^2 = \rho^{-1} \| \V{s} - \point \|^2_2 .
	\label{eq:falloff_relax}
\end{align}

Given this approximated scaled distance, we can obtain four additional \emph{linear} constraints on the scaled distance vector $\vunknown$, derived from its differences and sums, similar to the previous section.

First, \eref{eq:falloff_relax} of the symmetric pair can be rewritten using a shifted surface point $\point ' = \point - \lo$ as: 
\begin{equation}
	\left\{
	\begin{aligned}
		\dist{r+}{\theta} 
        &= \rho ^{-1}\norm{\light{r+}{\theta} - \point}_2^2 \\
		& =\rho ^{-1}\norm{\left(\light{r+}{\theta} - \lo\right) - \left(\point - \lo\right)}_2^2 \\
		&= \rho ^{-1}\left(
		r^2 
		+ {\point '}\transp{\point '}
		- 2{{\point}'} \transp\left(\light{r+}{\theta} - \lo\right)
		\right)\\
		\dist{r-}{\theta} &= \rho ^{-1}\left(
		r^2 
		+ {\point '}\transp{\point '}
		- 2{{\point}'} \transp\left(\light{r-}{\theta} - \lo\right)
		\right)\\
	\end{aligned}
	\right. .
	\label{eq:scaled_dist_def}
\end{equation}
By summing ${e}_*$, we have
\begin{equation}
	\begin{aligned}
		\dist{r+}{\theta} + \dist{r-}{\theta}
		&= 2\rho ^{-1}\left(
		r^2 
		+ {\point '}\transp{\point '}
		\right). 
	\end{aligned}
	\label{eq:sum_of_scaled_dist}
\end{equation}
Similarly, by taking a difference we can eliminate the quadratic terms as
\begin{equation}
	\begin{aligned}
		\dist{r+}{\theta} - \dist{r-}{\theta} 
		= -4 \rho^{-1}{\point'} \transp\left(\light{r+}{\theta} - \lo\right).  \\
	\end{aligned}
	\label{eq:diff_of_scaled_dist}
\end{equation}
 
\pparag{Additional constraint 1A} 
When we have at least three symmetric pairs whose radii are different from each other, the following homogeneous constraints can be obtained from \eref{eq:sum_of_scaled_dist} by eliminating albedo scaled radius:
\begin{equation}
    \begin{aligned}
    \rho^{-1} r^2 &= \frac{\left(\dist{\beta r+}{\psi} + \dist{\beta r-}{\psi}\right) - \left(\dist{r+}{\theta} + \dist{r-}{\theta}\right)}{2\left(\beta ^ 2 -1 \right)} \\
    &= \frac{\left(\dist{\alpha r+}{\phi} + \dist{\alpha r-}{\phi}\right) - \left(\dist{r+}{\theta} + \dist{r-}{\theta}\right)}{2\left(\alpha ^ 2 -1 \right)} 
    \end{aligned}
    \label{eq:e_constraints1a} 
\end{equation}
for $\npairs-2$ combinations.

\pparag{Additional constraint 1B}
As a special case of the above constraint, when we have symmetric pairs whose radii are the same as each other, we can obtain a constraint from \eref{eq:sum_of_scaled_dist},
\begin{equation}
    \dist{r+}{\theta} + \dist{r-}{\theta} = \dist{r+}{\phi} + \dist{r-}{\phi},
    \label{eq:e_constraints1b} 
\end{equation}
for $\npairs-1$ combinations.

\pparag{Additional constraint 2A}
Since \eref{eq:diff_of_scaled_dist} includes the relative light position \mbox{$\light{r+}{\theta} - \lo$} similarly in \eref{eq:difference}, we obtain the following equations by similar manner to \eref{eq:linear_constraints1}:
\begin{equation}
    \dist{\beta r+}{\psi} - \dist{\beta r-}{\psi}  
    = s \left(\dist{r+}{\theta} - \dist{r-}{\theta} \right)
    + t \left(\dist{\alpha r+}{\phi}  - \dist{\alpha r-}{\phi} \right),
    \label{eq:e_constraints2a} 
\end{equation}
assuming $\light{\beta r+}{\psi}$ can be represented using the two basis $\light{r+}{\theta}$ and $\light{\alpha r+}{\phi}$ (\eref{eq:relative_position_using_basis}). Considering all the possible pairs leads to \mbox{$\npairs - 2$} constraints, similar to \eref{eq:linear_constraints1}. 

\pparag{Additional constraint 2B}
The above three constraints, Eqs.~(\ref{eq:e_constraints1a})-(\ref{eq:e_constraints2a}), hold for arbitrary global offset $\lo$.
In the specific case where the lights only have the global offset along $z$-axis, \ie \hbox{$\lo=\left[0,0,s_z\right]$}, there are additional homogeneous constraints instead of \eref{eq:e_constraints2a} as:
\begin{equation}
    \begin{aligned}
        {\V{p}}\transp
        \left(\frac{\light{r+}{\theta}-\lo}{r}\right)
        &\left(\dist{\alpha r+}{\phi} - \dist{\alpha r-}{\phi}\right) \\
        &= {\V{p}}\transp
        \left(\frac{\light{\alpha r+}{\phi}-\lo}{r} \right)
        \left(\dist{r+}{\theta} - \dist{r-}{\theta}\right)
    \end{aligned},
    \label{eq:e_constraints2b} 
\end{equation}
where $\left(\light{r+}{\theta}-\lo\right) / r = \left[\sin{\theta}, \cos{\theta}, 0\right]$, and $\V{p} = \left[u',v', 1\right]$ is a normalized camera coordinate, which is parameterized as 
\begin{equation}
    \begin{aligned}
        {\V{p}}
        = \left[\frac{u-c_x}{f_x}, \frac{v-c_y}{f_y}, 1\right]\transp,
    \end{aligned}
    \label{eq:normalized_cam_coord} 
\end{equation}
where $\left(u,v\right)$,  $\left(f_x,f_y\right)$, and $\left(c_x,c_y\right)$ are the pixel coordinate, focal length, and optical center in a pixel unit, respectively.
This constraint comes from \eref{eq:diff_of_scaled_dist} and \hbox{${\point'} \transp\left(\light{r+}{\theta} - \lo\right) = z'{\V{p}} \transp\left(\light{r+}{\theta} - \lo\right)$}, where \hbox{${\point'} \triangleq \left[x', y', z'\right]\transp $}.

Considering all the combinations of the symmetric pairs, $\npairs - 1$ linearly independent constraints can be obtained.

\vspace{3mm}
In summary, the equations in Eqs.~(\ref{eq:e_constraints1a})-(\ref{eq:e_constraints2b}) are the additional constraints obtained by the relaxation. Putting them in a matrix $\V{A}' \in \mathbb{R}^{\neqs' \times 2\npairs}$, we have a new homogeneous system \mbox{$\V{A}' \V{e} = \V{0}$}, where $\neqs'$ is the number of all the equations from the relaxation.
Putting the constraints together, we solve the following minimization problem by vertically stacking constraint matrices $\V{A}$ and $\V{A'}$:
\begin{equation}
   \begin{gathered}
   {\vunknown}^* = \argmin_{{\vunknown}}{\norm{
   \left[
       \begin{array}{c}
   \V{A} \\
   \V{A}'
   \end{array} 
   \right]\vunknown~}_2^2} \quad
   \mathrm{\st} \quad \V{e}\transp \V{e} = 1. 
   \end{gathered}
   \label{eq:e_objective_func}
\end{equation}
The problem is equivalent to finding the right null space of the stacked constraint matrix. We solve the problem by the singular value decomposition (SVD) and ensure that the sign of the estimated $\V{e}$ becomes positive because they are scaled distances. While conventional SVD does not guarantee $\V{e} \ge \V{0}$, we have empirically observed that the elements of the estimated vector $\V{e}$ always have the same signs.

\subsection{Surface normal and position estimation}
\label{para:relative_depth_estimation}
From scaled distances $\V{e}$ estimated in the previous section, we here estimate the surface position $\V{x}$ up to scale and surface normal $\normal$. If we know the radius $r$ of the symmetric lights, the surface position $\V{x}$ can be recovered without the scaling ambiguity.

From \eref{eq:e_constraints1a}, we can parameterize $\rho^{-1}r^2$ using $\V{e}$, which we know the exact value from the estimated $\V{e}$.
Let us define a scaled and shifted surface position $\point_r'$ as
\begin{equation}
   {\point_r'} \triangleq \frac{\point'}{r} = 
   \left[x_r', y_r', z_r'\right]\transp = \left[\frac{x-s_x}{r},\frac{y-s_y}{r}, \frac{z-s_z}{r}\right]\transp, 
\end{equation}
Since \eref{eq:diff_of_scaled_dist} can be rewritten using $r$ and $\theta$ as 
\begin{equation}
   \begin{aligned}
    \dist{r+}{\theta} - \dist{r-}{\theta} 
        = -4 \rho^{-1}r^2 {\point'_r} \transp
        \left[\sin{\theta}, \cos{\theta}, 0\right]\transp,
       \label{eq:scaled_x}
   \end{aligned}
\end{equation}
we can derive $x_r'$ from \eref{eq:diff_of_scaled_dist} as 
\begin{equation}
   \begin{aligned}
        x_r' = \frac{
            \left(\dist{r+}{\theta} - \dist{r-}{\theta}\right)\alpha \cos{\phi}
            - \left(\dist{\alpha r+}{\phi} - \dist{\alpha r-}{\phi}\right) \cos{\theta}
            }{-4\rho^{-1}r^2\alpha\sin{\left(\theta - \phi\right)}}.
       \label{eq:scaled_x}
   \end{aligned}
\end{equation}
Similarly, $y'$ can be obtained as 
\begin{equation}
   \begin{aligned}
    y_r' = \frac{
        \left(\dist{r+}{\theta} - \dist{r-}{\theta}\right)\alpha \sin{\phi}
        - \left(\dist{\alpha r+}{\phi} - \dist{\alpha r-}{\phi}\right) \sin{\theta}
        }{-4\rho^{-1}r^2\alpha\sin{\left(\phi - \theta\right)}}.
   \end{aligned}
\end{equation}
Substituting $x'$, $y'$, and $\rho ^{-1}r^2$ into \eref{eq:scaled_dist_def} yields 
\begin{equation}
   \begin{aligned}
       z'_r &= \sqrt[]{\frac{\dist{r+}{\theta}}{2\rho ^{-1}r^2} - \left(x_r'^2 + y_r'^2 + 1\right) + 2 \left(\frac{\point'}{r}\right)\transp\left(\frac{\light{r+}{\theta}-\lo}{r}\right)},
       \label{eq:scaled_shifted_z}
 \end{aligned}
\end{equation}
assuming the camera is looking toward the $+z$ direction. Note that the $z$ element of the relative light position $\light{r+}{\theta}-\lo$ is always $0$, and thus the third term of \eref{eq:scaled_shifted_z} can be parameterized by $x'_r$, $y'_r$ and the known angle $\theta$. The same can be computed for the other combination of lights. In our method, we take the mean estimate from using all the equations from all the light pairs, and as a result, we have the scaled and shifted surface point $\point'_r$.

From the relaxed image formation model introduced in the previous section, we have
\begin{align}
   \m_* \approx \rho \frac{\left(\vs - \vx\right)\transp\normal}{\norm{\vs - \vx}_2^2}
   = \rho r^{-1} \frac{\left(\vs'_r - \point'_r\right)\transp\normal}{\norm{\vs'_r - \point'_r}_2^2},
\end{align}
where $\vs_r' = (\vs - \lo)/r$, \eg $\light{\alpha r+}{\phi} =  \alpha\left[\sin{\phi},\cos{\phi},0\right]\transp$, which are the known relative positions of symmetric lights. 
Since only unknowns are the scalar $\rho r^{-1}$ and surface normal $\normal$, and the norm of normal $\norm{\V{n}}^2_2$ should be always $1$, once the scaled and shifted surface point $\point'_r$ is obtained, the surface normal estimate can be obtained by closed-form solution as classic calibrated photometric stereo~\cite{woodham1980}.

\subsection{Optimality under relaxation}
Our two-step approach can be summarized as follows:
\begin{equation}
 \left\{
\begin{aligned}
    \V{e}^* &= \argmin_{{\vunknown}}{~\mathcal{L}(\vunknown)} \quad
        \mathrm{\st} \quad \V{e}\transp \V{e} = 1 \\ 
        \mathcal{L}(\vunknown) &= 
    \norm{
        \left[
            \begin{array}{c}
        \V{A} \\
        \V{A}'
        \end{array} 
    \right]\vunknown~}_2^2 \\
        \V{e} &= g(\point)
\end{aligned}
\right.,
\end{equation}
where $g(\point): \point \mapsto \vunknown$ denotes the mapping function that corresponds to \eref{eq:scaled_dist_def}.
Our approach can be viewed as optimizing a surface point $\point$ that minimizes the objective function $\mathcal{L}(g(\point))$.
From Eqs.~(\ref{eq:scaled_x})-(\ref{eq:scaled_shifted_z}) in \sref{para:relative_depth_estimation}, the mapping from $\vunknown$ to $\point$, hereafter denoted as $g^{-1}: \vunknown \mapsto \point$, is a one-to-one mapping function.
Since the globally optimal $\vunknown^*$ can be obtained by convex optimization and $g^{-1}$ is an one-to-one mapping function, the obtained surface point $\point^* = g^{-1}(\vunknown^*)$ gives the global minimum of the objective $\mathcal{L}(g(\point))$. Surface normal can also be derived as a closed-form solution using the optimal $\point^*$.

\subsection{Valid light arrangements}
\label{sec:solvable}
In this section, we analyze the solvability of our problem to present the practical light arrangements derived from the constraints.
We here consider the general case where all the lights are not on the same line and discuss the special case later.

\pparag{Scaled distance estimation}
We first present the solvability on scaled distance estimation in all possible combinations. 
Throughout this section, we establish the coefficient matrices $\Veqia, \Veqib, \Veqiia$, and $\Veqiib$ by assembling the four constraints derived by the light fall-off relaxation (Eqs.~(\ref{eq:e_constraints1a})-(\ref{eq:e_constraints2b})), respectively. 
Considering the linearly independent constraints,
the rank of these matrices is as follows:
\begin{equation}
    \left\{
    \begin{array}{lll}
        &\rank{\V{A}} &= 2\npairs - 3 \\
        &\rank{\Veqia} &= \npairs - 2 \quad (\text{non-ring~ light})\\
        &\rank{\Veqib} &=  \npairs - 1 \quad (\text{ring~ light})\\
        &\rank{\Veqiia} &= \npairs - 2  \quad (xyz\text{-axis~ offset}) \\
        &\rank{\Veqiib} &=  \npairs - 1  \quad (z\text{-axis~ offset}) 
      \end{array} 
    \right. . \nonumber
\end{equation}
Here, $\Veqia$ and $\Veqib$ are both obtained by the elementary operation on \eref{eq:sum_of_scaled_dist}, and thus $\Veqia$ is a subset of $\Veqib$. Similarly, $\Veqiia$ is a subset of $\Veqiib$.
Overall, we have four possible cases where (i) non-ring light with the global offset along $xyz$-axis, (ii)~non-ring light with the offset along $z$-axis, (iii) ring light with the offset along $xyz$-axis, and (iv) ring light with the offset along $z$-axis, which lead to the corresponding additional constraints of $\eqia + \eqiia$, $\eqia + \eqiib$, $\eqib + \eqiia$, and $\eqib + \eqiib$, respectively.
\begin{figure}[t]
	\centering
	\includegraphics[width=\linewidth]{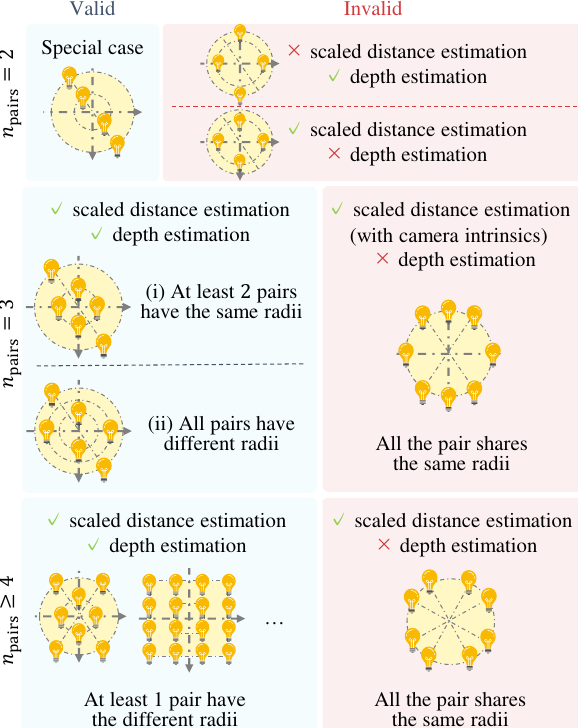}
	\caption{Examples of valid/invalid light arrangements.}
	\label{fig:valid_arrangenemts}
\end{figure}
\begin{figure}[t]
	\centering
	\includegraphics[width=\linewidth]{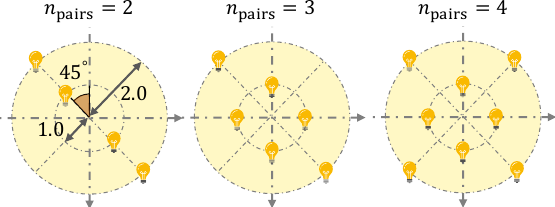}
	\caption{Light arrangements used in the synthetic experiments. In every case, we set the inner and outer radii as $1$ and $2$. }
	\label{fig:light_arrangement_in_exp}
\end{figure}
\begin{figure*}[h!]
	\scriptsize
	\newcommand{\figwidthNormalVisSupp}{0.085}
	\newcommand{\halfFigwidthNormalVisSupp}{0.045}
	\definecolor{lightGray}{gray}{0.6}
	\centering
	\begin{tabular}{@{}cc@{}c@{}c@{}c@{}c@{}c|c@{}c@{}c@{}c@{}c@{}c@{}}
		& & \multicolumn{5}{c}{$\npairs=3$} & \multicolumn{5}{c}{$\npairs=4$} & \\
		& & GT/Input & Ours & Calibrated~\cite{Yvain2018} & fastNFPS~\cite{lichy2022fast} & UniversalPS~\cite{ikehata2022universal} & GT & Ours & Calibrated~\cite{Yvain2018} & fastNFPS~\cite{lichy2022fast} & UniversalPS~\cite{ikehata2022universal} & \\
		\multirow{7}{*}{\rotatebox[origin=c]{90}{\textsc{Bunny}}}
		& 
		\raisebox{\halfFigwidthNormalVisSupp\linewidth}{\rotatebox[origin=c]{90}{Normal map}}
		&
		\includegraphics[width=\figwidthNormalVisSupp\linewidth]{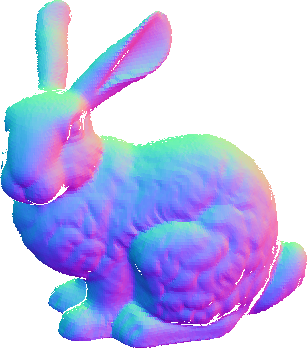}
		&
		\includegraphics[width=\figwidthNormalVisSupp\linewidth]{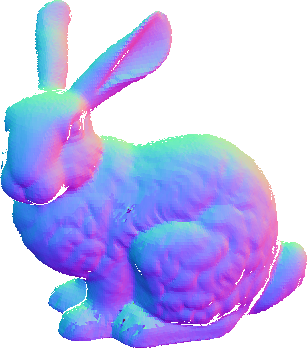}
		&
		\includegraphics[width=\figwidthNormalVisSupp\linewidth]{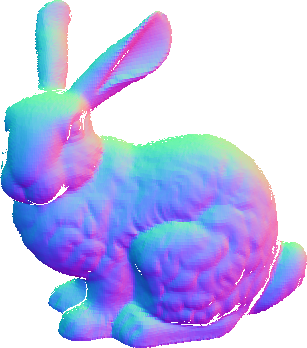}
		&
		\includegraphics[width=\figwidthNormalVisSupp\linewidth]{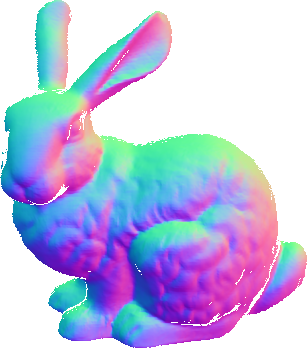}
		&
		\includegraphics[width=\figwidthNormalVisSupp\linewidth]{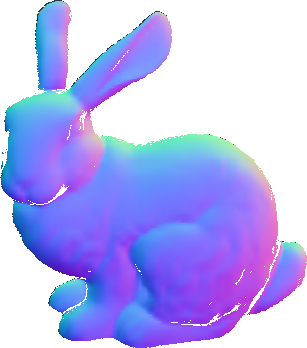}
		&
		\includegraphics[width=\figwidthNormalVisSupp\linewidth]{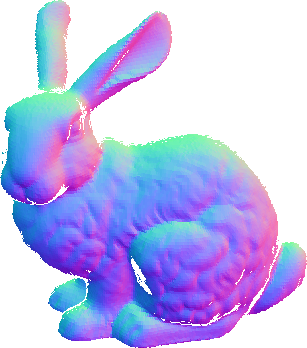}
		&
		\includegraphics[width=\figwidthNormalVisSupp\linewidth]{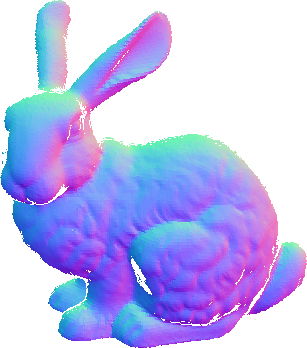}
		&
		\includegraphics[width=\figwidthNormalVisSupp\linewidth]{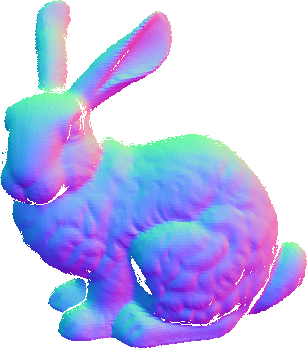}
		&
		\includegraphics[width=\figwidthNormalVisSupp\linewidth]{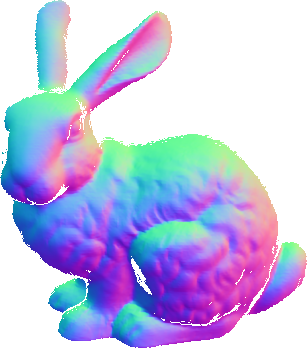}
		&
		\includegraphics[width=\figwidthNormalVisSupp\linewidth]{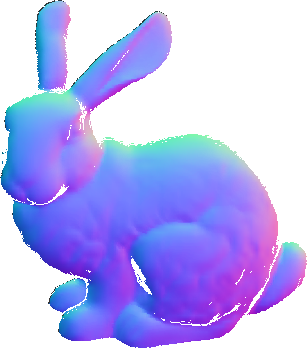}
		&
		\includegraphics[width=0.06\linewidth]{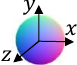}
		\\
		&
		\raisebox{\halfFigwidthNormalVisSupp\linewidth}{\rotatebox[origin=c]{90}{Error map}}
		&
	    \includegraphics[width=\figwidthNormalVisSupp\linewidth]{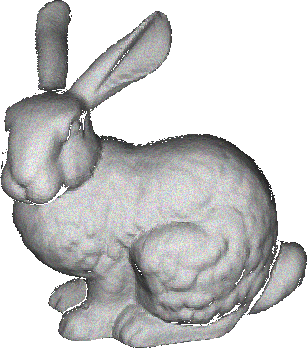}
		&
		\includegraphics[width=\figwidthNormalVisSupp\linewidth]{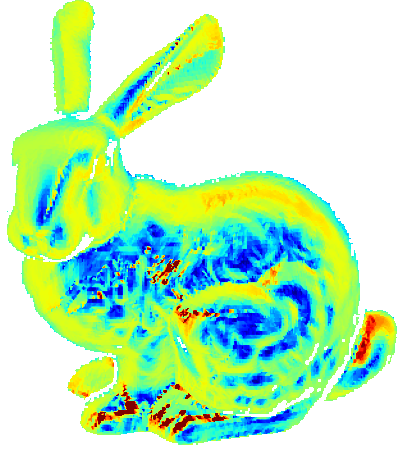}
		&
		\includegraphics[width=\figwidthNormalVisSupp\linewidth]{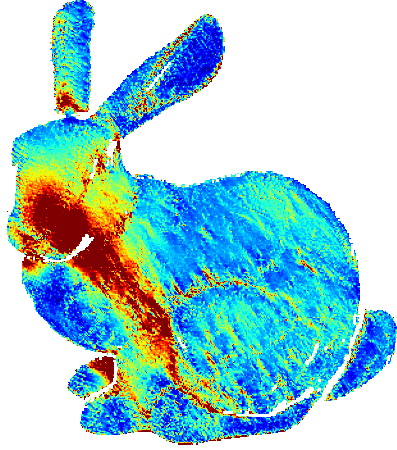}
		&
		\includegraphics[width=\figwidthNormalVisSupp\linewidth]{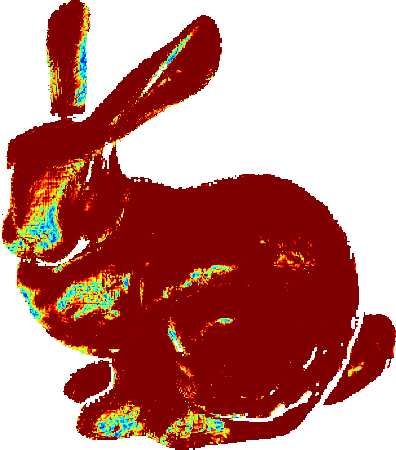}
		&
		\includegraphics[width=\figwidthNormalVisSupp\linewidth]{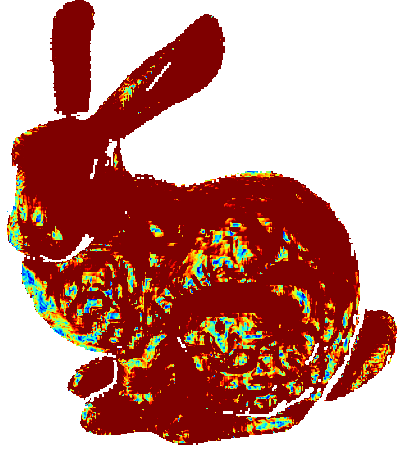}
		&
	    \includegraphics[width=\figwidthNormalVisSupp\linewidth]{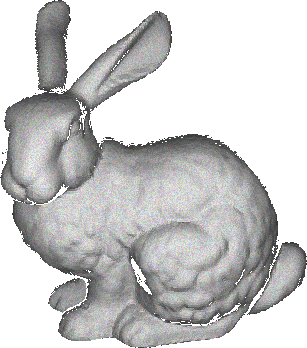}
		&
		\includegraphics[width=\figwidthNormalVisSupp\linewidth]{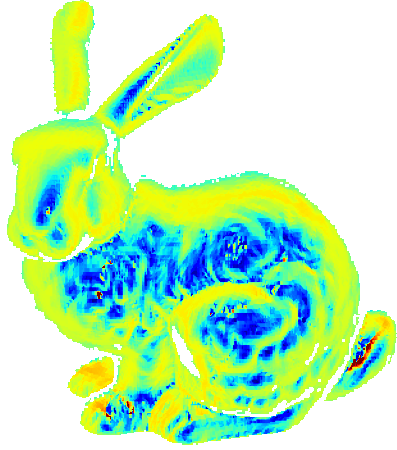}
		&
		\includegraphics[width=\figwidthNormalVisSupp\linewidth]{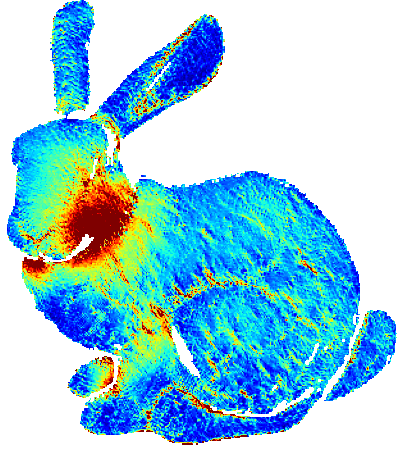}
		&
		\includegraphics[width=\figwidthNormalVisSupp\linewidth]{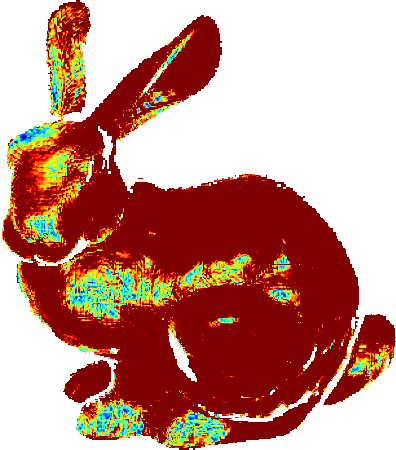}
		&
		\includegraphics[width=\figwidthNormalVisSupp\linewidth]{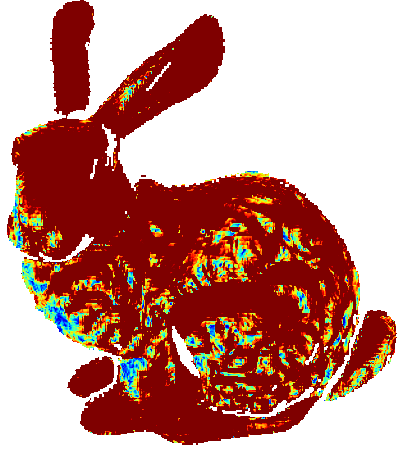}
		&
		\\
		&
		&& 4.974 & 4.590 & 16.312 & 17.719 & & 4.897 & 3.896 & 14.934 & 17.009  
		\\
		&
		\raisebox{\halfFigwidthNormalVisSupp\linewidth}{\rotatebox[origin=c]{90}{Depth map}}	
		&
		\includegraphics[width=\figwidthNormalVisSupp\linewidth]{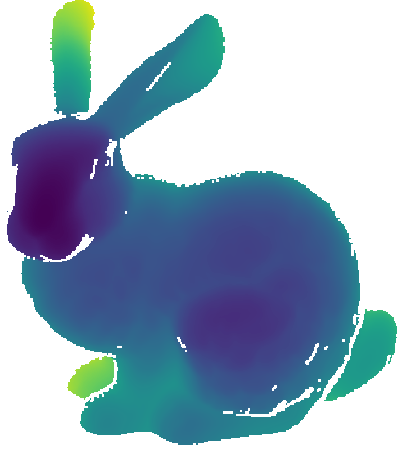}
		&
		\includegraphics[width=\figwidthNormalVisSupp\linewidth]{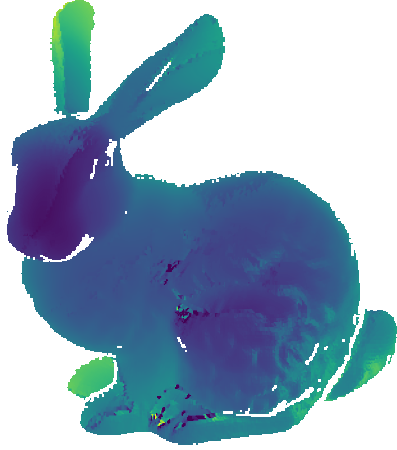}
		&
		\includegraphics[width=\figwidthNormalVisSupp\linewidth]{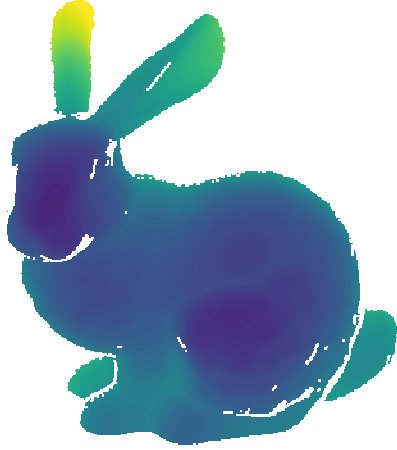}
		&
		\includegraphics[width=\figwidthNormalVisSupp\linewidth]{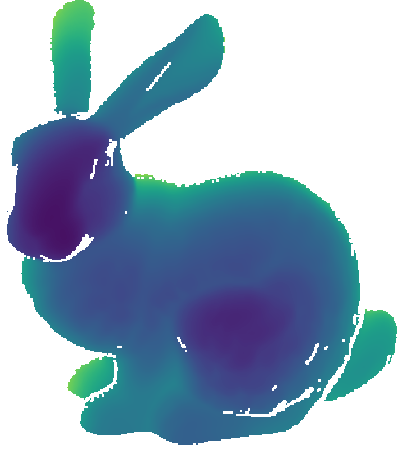}
		& 
		\multirow{2}{*}{
			\rotatebox[origin=c]{-10}{\textcolor{lightGray}{\rule{.1pt}{40pt}}}
		}
		&
		\includegraphics[width=\figwidthNormalVisSupp\linewidth]{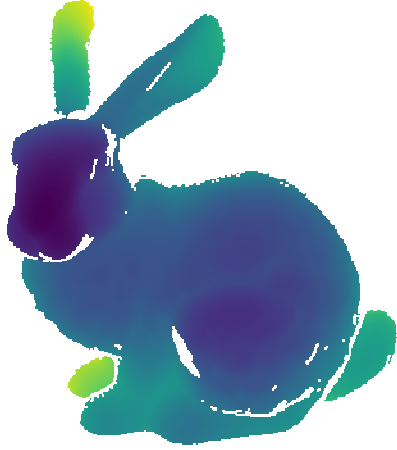}
		&
		\includegraphics[width=\figwidthNormalVisSupp\linewidth]{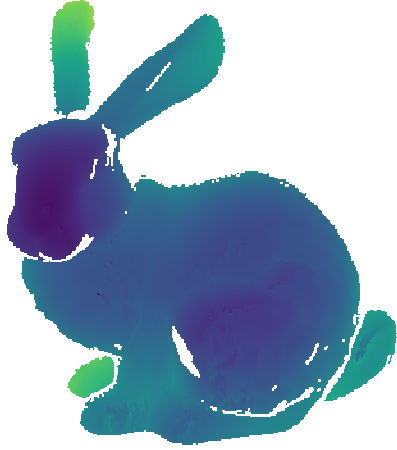}
		&
		\includegraphics[width=\figwidthNormalVisSupp\linewidth]{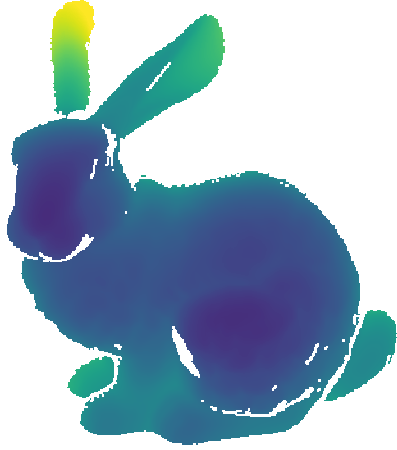}
		&
		\includegraphics[width=\figwidthNormalVisSupp\linewidth]{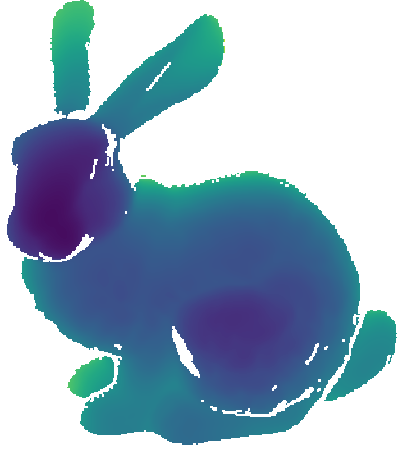}
		& 
		\multirow{2}{*}{
			\rotatebox[origin=c]{-10}{\textcolor{lightGray}{\rule{.1pt}{40pt}}}
		}
		\\
		&
		\raisebox{\halfFigwidthNormalVisSupp\linewidth}{\rotatebox[origin=c]{90}{Error map}}	
		&
		 \includegraphics[width=\figwidthNormalVisSupp\linewidth]{figures/results/oPlusLine45_lambertian_SValbedo_bunny_float32_offset_0.0x0.0x0.5_p3/sample_image.png}
		&
		\includegraphics[width=\figwidthNormalVisSupp\linewidth]{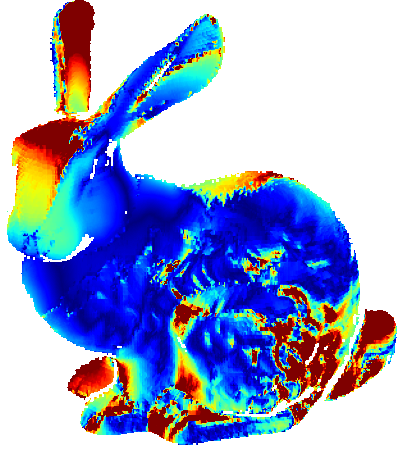}
		&
		\includegraphics[width=\figwidthNormalVisSupp\linewidth]{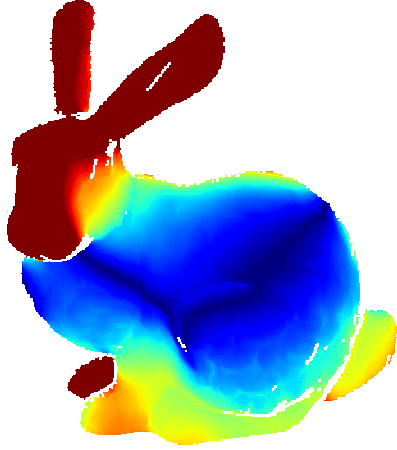}
		&
		\includegraphics[width=\figwidthNormalVisSupp\linewidth]{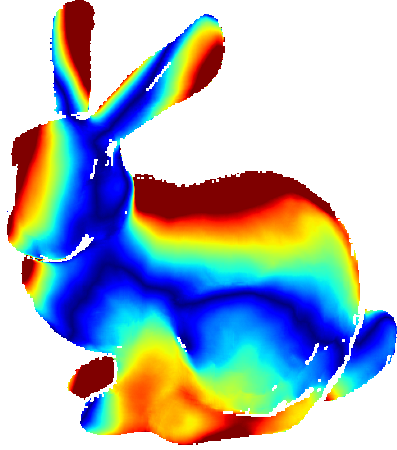}
		&
		&
		\includegraphics[width=\figwidthNormalVisSupp\linewidth]{figures/results/doubleO45_lambertian_SValbedo_bunny_float32_offset_0.0x0.0x0.5_p3/sample_image.png} 
		&
		\includegraphics[width=\figwidthNormalVisSupp\linewidth]{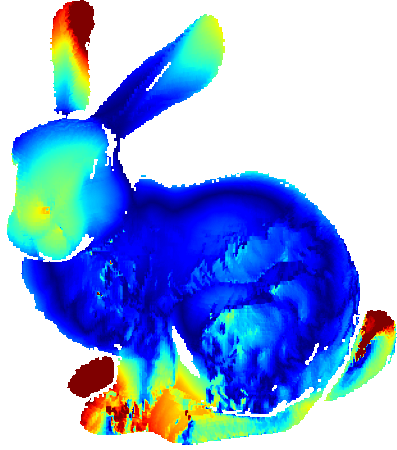}
		&
		\includegraphics[width=\figwidthNormalVisSupp\linewidth]{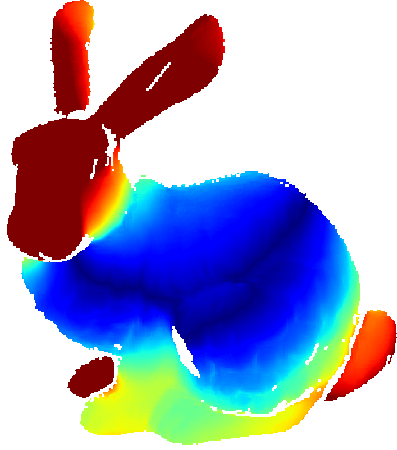}
		&
		\includegraphics[width=\figwidthNormalVisSupp\linewidth]{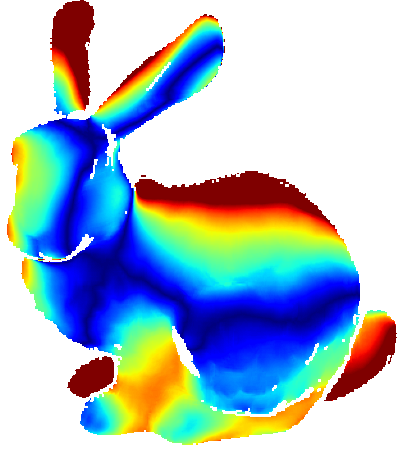}
		&
		&
		\\
		&
		&& 0.026 & 0.032 & 0.031 &  & & 0.017 & 0.031 & 0.029 &\\
		\multirow{6}{*}{\raisebox{\halfFigwidthNormalVisSupp\linewidth}{\rotatebox[origin=c]{90}{\textsc{Crab}}}} & 
		\raisebox{\halfFigwidthNormalVisSupp\linewidth}{\rotatebox[origin=c]{90}{Normal map}}	
		&
		 \includegraphics[width=\figwidthNormalVisSupp\linewidth]{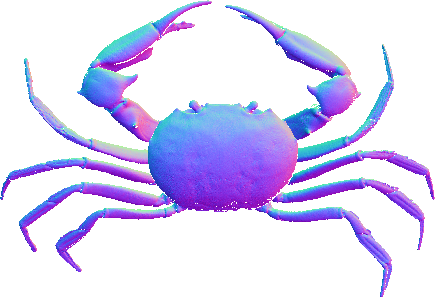}
		 &
		 \includegraphics[width=\figwidthNormalVisSupp\linewidth]{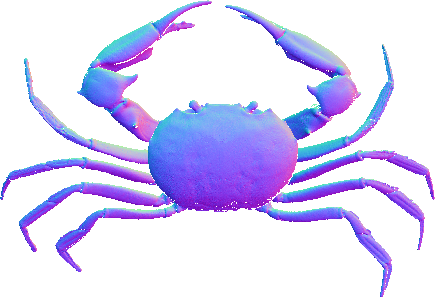}
		 &
		 \includegraphics[width=\figwidthNormalVisSupp\linewidth]{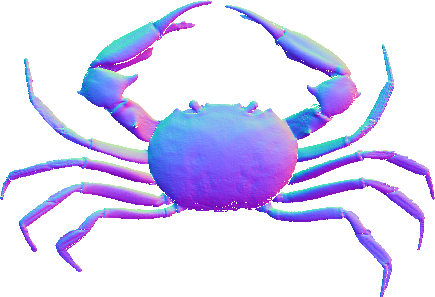}
		 &
		 \includegraphics[width=\figwidthNormalVisSupp\linewidth]{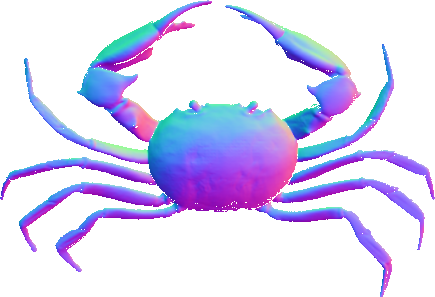}
		 &
		 \includegraphics[width=\figwidthNormalVisSupp\linewidth]{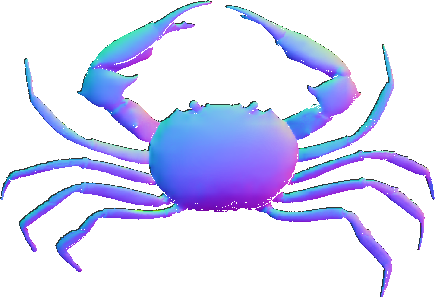}
		 &
		 \includegraphics[width=\figwidthNormalVisSupp\linewidth]{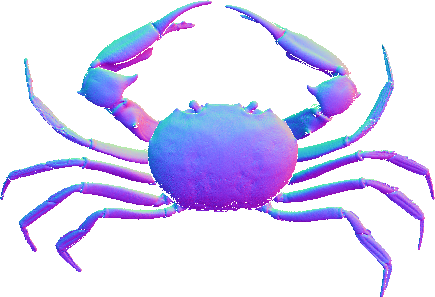}
		 &
		 \includegraphics[width=\figwidthNormalVisSupp\linewidth]{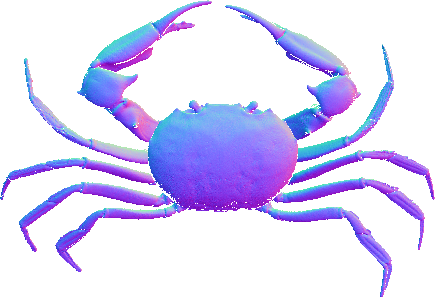}
		 &
		 \includegraphics[width=\figwidthNormalVisSupp\linewidth]{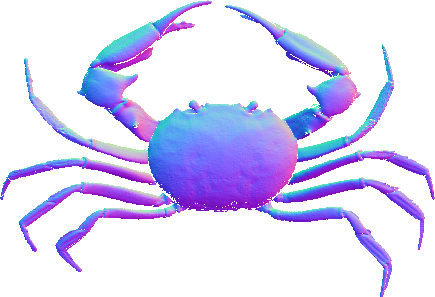}
		 &
		 \includegraphics[width=\figwidthNormalVisSupp\linewidth]{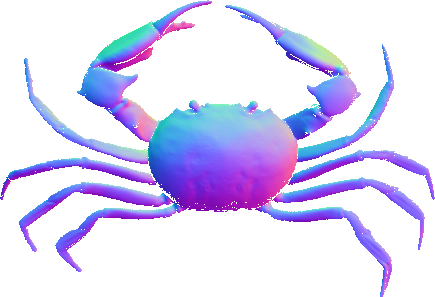}
		 &
		 \includegraphics[width=\figwidthNormalVisSupp\linewidth]{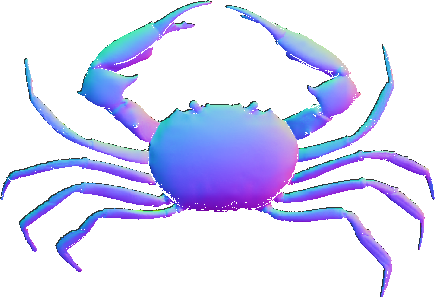}
		 &
		 \\
		 &
		 \raisebox{\halfFigwidthNormalVisSupp\linewidth}{\rotatebox[origin=c]{90}{Error map}}	
		 &
		 \includegraphics[width=\figwidthNormalVisSupp\linewidth]{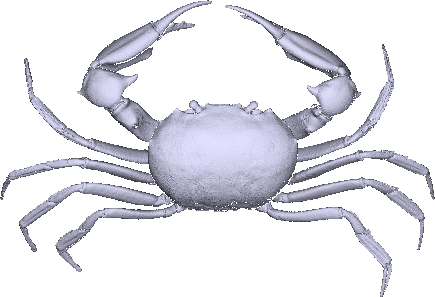}
		 &
		 \includegraphics[width=\figwidthNormalVisSupp\linewidth]{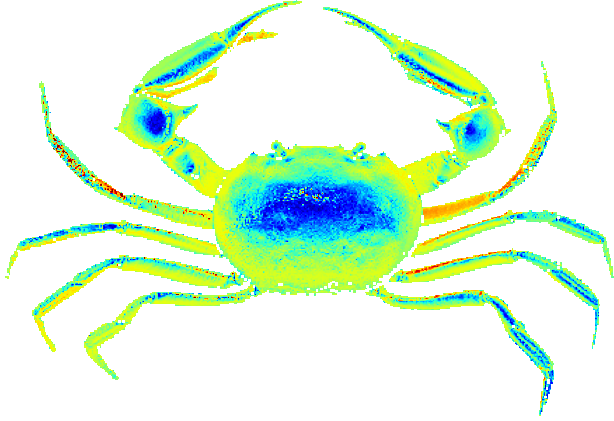}
		 &
		 \includegraphics[width=\figwidthNormalVisSupp\linewidth]{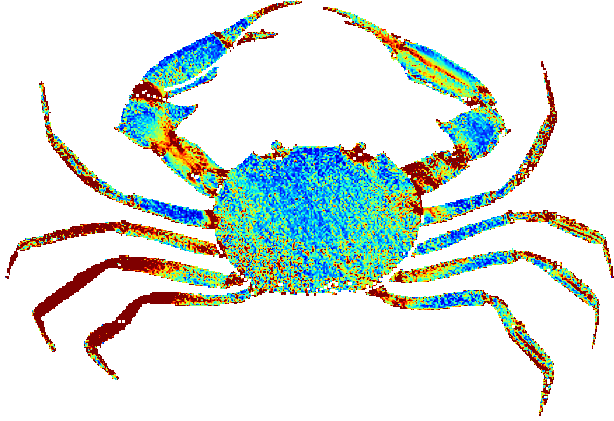}
		 &
		 \includegraphics[width=\figwidthNormalVisSupp\linewidth]{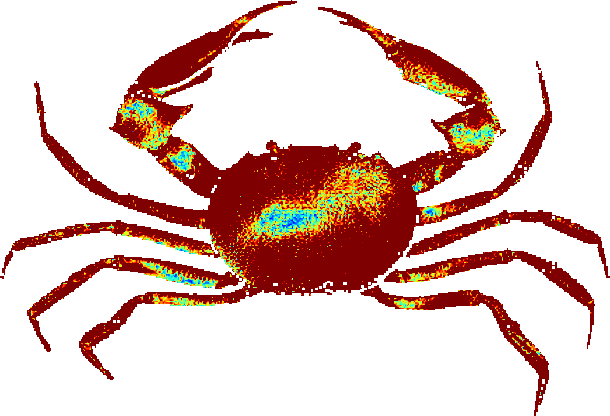}
		 &
		 \includegraphics[width=\figwidthNormalVisSupp\linewidth]{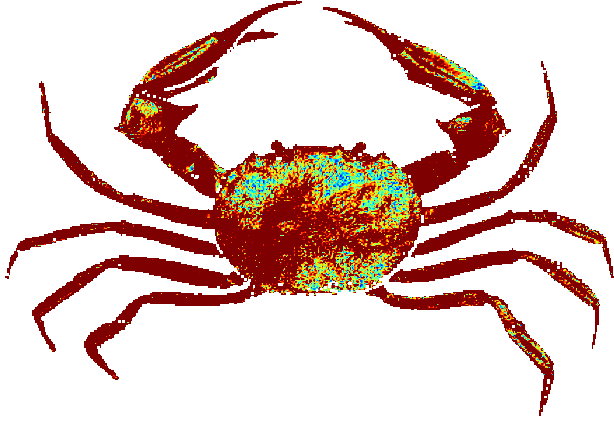}
		 &
		 \includegraphics[width=\figwidthNormalVisSupp\linewidth]{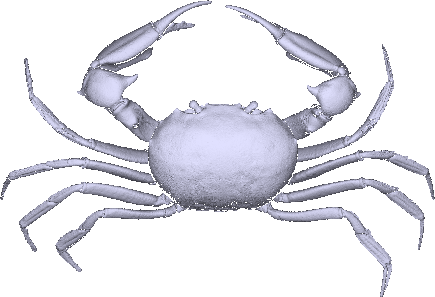}
		 &
		 \includegraphics[width=\figwidthNormalVisSupp\linewidth]{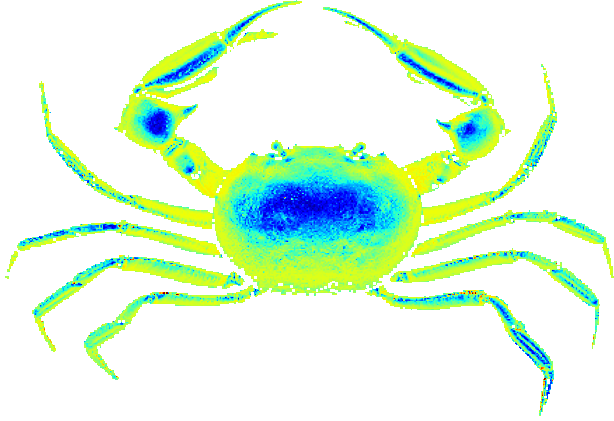}
		 &
		 \includegraphics[width=\figwidthNormalVisSupp\linewidth]{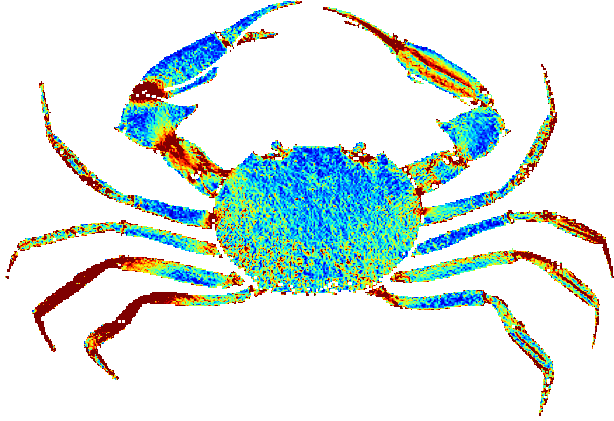}
		 &
		 \includegraphics[width=\figwidthNormalVisSupp\linewidth]{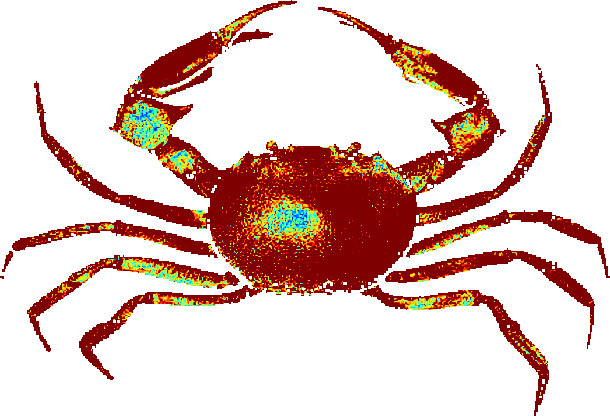}
		 &
		 \includegraphics[width=\figwidthNormalVisSupp\linewidth]{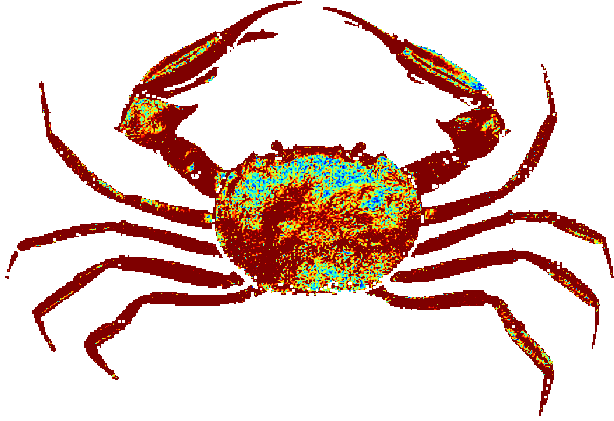}
		 &
		 \\
		 &
		 && 4.907 & 7.384 & 15.514 & 22.176 &  & 4.886 & 6.196 & 14.451 & 20.439 &  \\
				&
				\raisebox{\halfFigwidthNormalVisSupp\linewidth}{\rotatebox[origin=c]{90}{Depth map}}	
				&
				\includegraphics[width=\figwidthNormalVisSupp\linewidth]{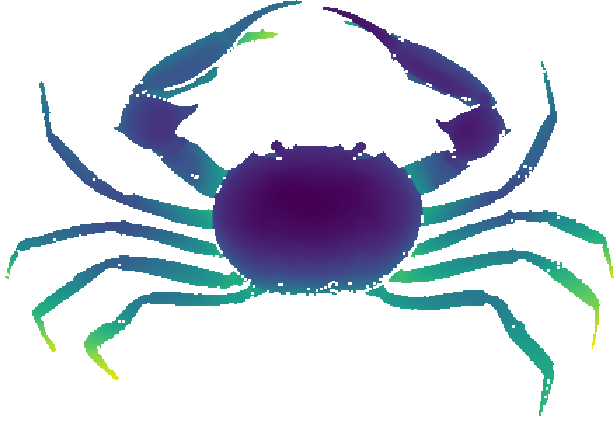}
				&
				\includegraphics[width=\figwidthNormalVisSupp\linewidth]{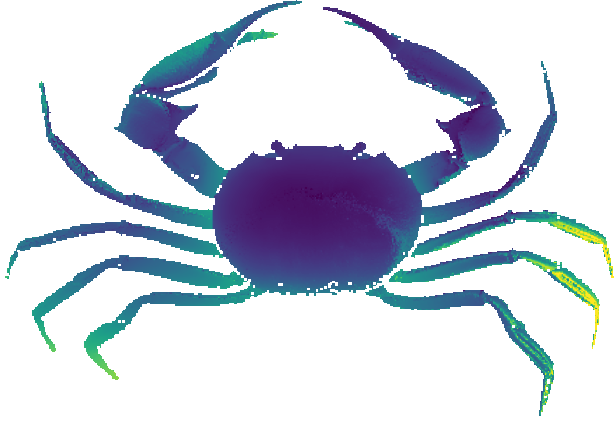}
				&
				\includegraphics[width=\figwidthNormalVisSupp\linewidth]{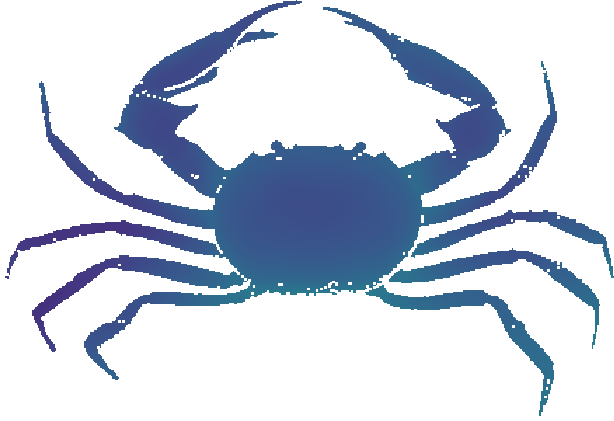}
				&
				\includegraphics[width=\figwidthNormalVisSupp\linewidth]{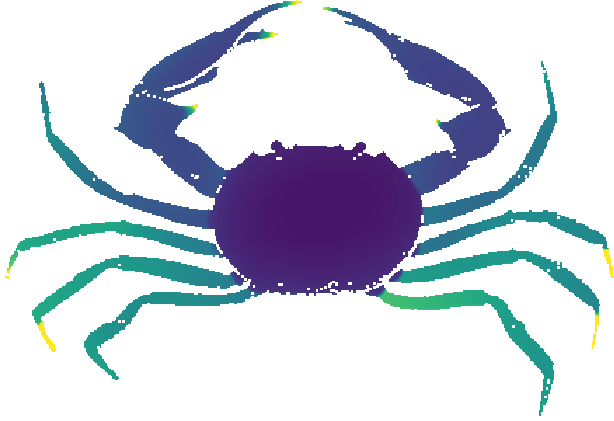}
				& 
				\multirow{2}{*}{
					\rotatebox[origin=c]{-10}{\textcolor{lightGray}{\rule{.1pt}{40pt}}}
				}
				&
				\includegraphics[width=\figwidthNormalVisSupp\linewidth]{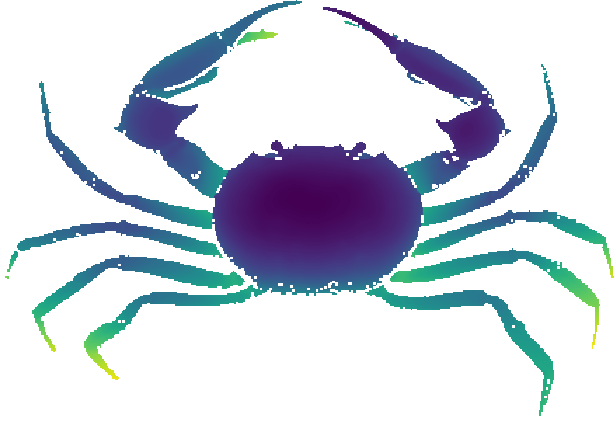}
				&
				\includegraphics[width=\figwidthNormalVisSupp\linewidth]{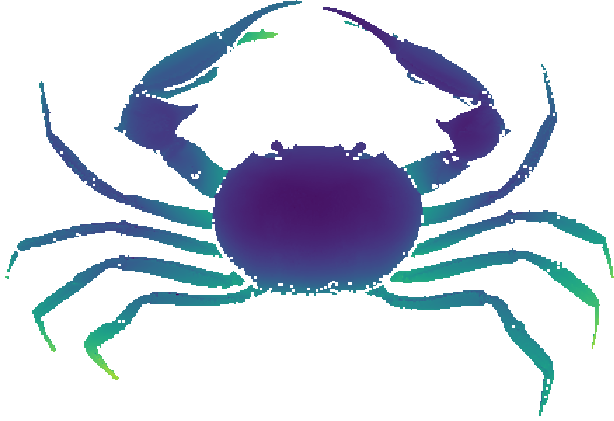}
				&
				\includegraphics[width=\figwidthNormalVisSupp\linewidth]{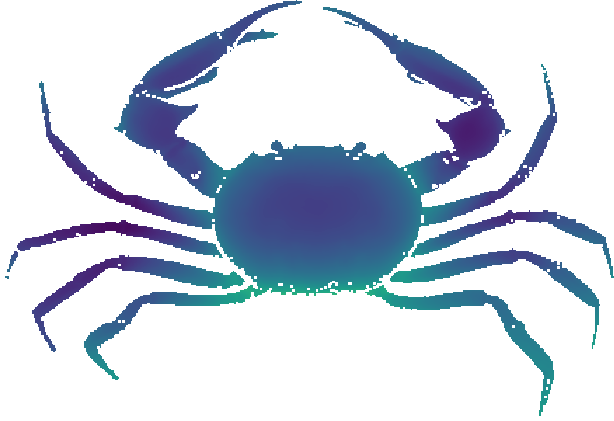}
				&
				\includegraphics[width=\figwidthNormalVisSupp\linewidth]{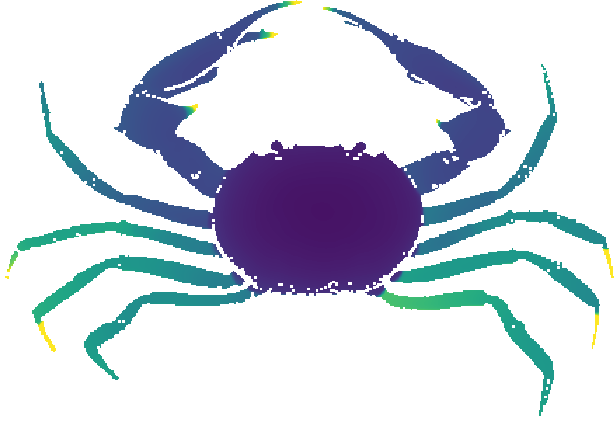}
				& 
				\multirow{2}{*}{
					\rotatebox[origin=c]{-10}{\textcolor{lightGray}{\rule{.1pt}{40pt}}}
				}	
				\\
				&
				\raisebox{\halfFigwidthNormalVisSupp\linewidth}{\rotatebox[origin=c]{90}{Error map}}	
				&
		 		\includegraphics[width=\figwidthNormalVisSupp\linewidth]{figures/results/oPlusLine45_lambertian_crab_float32_offset_0.0x0.0x0.5_p3/sample_image.png}
				&
				\includegraphics[width=\figwidthNormalVisSupp\linewidth]{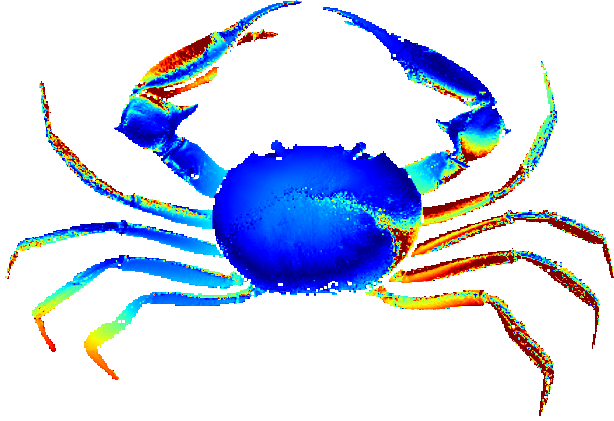}
				&
				\includegraphics[width=\figwidthNormalVisSupp\linewidth]{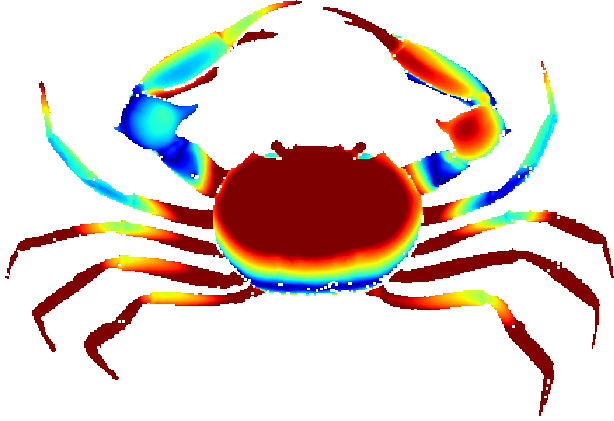}
				&
				\includegraphics[width=\figwidthNormalVisSupp\linewidth]{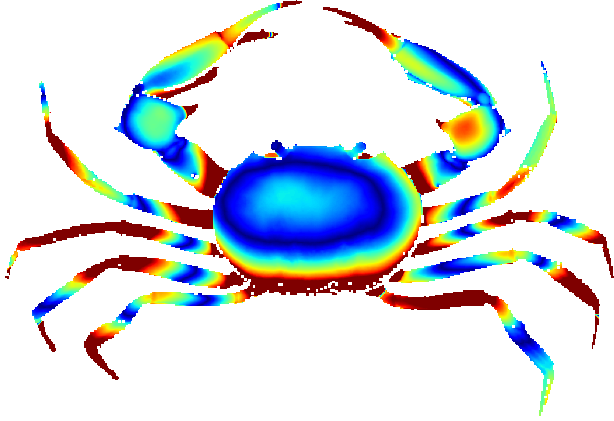}
				&
				&
				\includegraphics[width=\figwidthNormalVisSupp\linewidth]{figures/results/doubleO45_lambertian_crab_float32_offset_0.0x0.0x0.5_p3/sample_image.png} 
				&
				\includegraphics[width=\figwidthNormalVisSupp\linewidth]{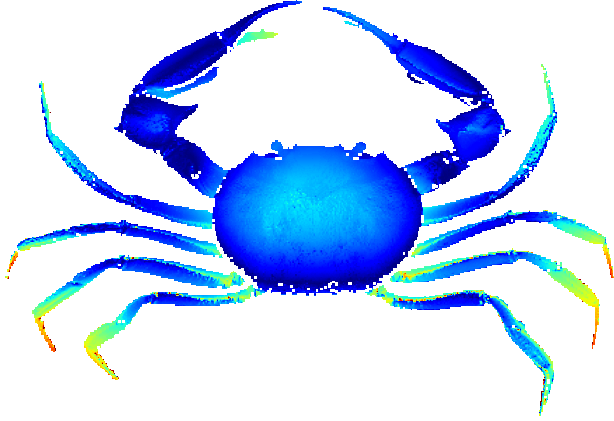}
				&
				\includegraphics[width=\figwidthNormalVisSupp\linewidth]{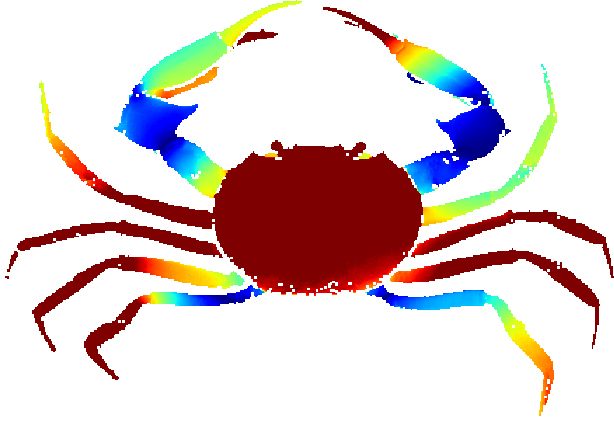}
				&
				\includegraphics[width=\figwidthNormalVisSupp\linewidth]{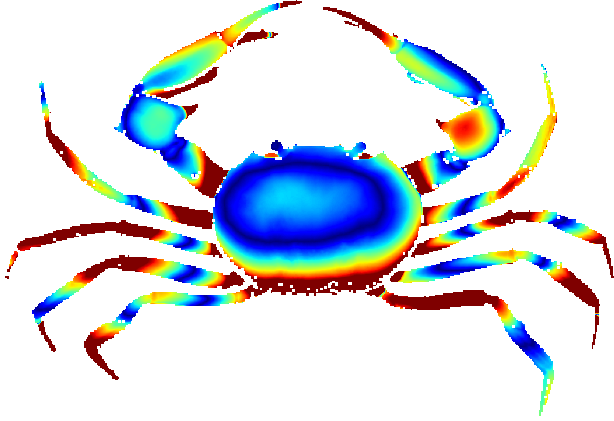}
				&
				&
				\\
				& 
			&	& 0.024 & 0.058 & 0.037 &  &  & 0.013 & 0.055 & 0.038 &  \\
		 		&&\rotatedColorbar{Normal error}{0.065}{$10^\circ$}{-15} & 
		 		\rotatedColorbar{Depth error}{0.065}{$0.01$}{-15} & 
	\end{tabular}
	\caption{Estimation results of our method and comparison methods, with the global offset along $z$-axis $\lo=\left[0.0,0.0,0.5\right]$ added to the center of the symmetric light pairs. 
	The left-hand side and right-hand side columns show the estimations from $\npairs=3$ and $\npairs=4$ cases, respectively. We show the normal maps and depth maps along with the corresponding error maps, as well as a mean angular~(for normal)/relative absolute errors~(for depth) at the bottom of each error map.}
	\label{fig:synth_results_z_shift}
	\vspace*{-10pt}
\end{figure*}
 
\begin{figure*}[h!]
	\scriptsize
	\newcommand{\figwidthNormalVisSupp}{0.085}
	\newcommand{\halfFigwidthNormalVisSupp}{0.045}
	\definecolor{lightGray}{gray}{0.6}
	\centering
	\begin{tabular}{@{}cc@{}c@{}c@{}c@{}c@{}c@{}|c@{}c@{}c@{}c@{}c@{}c@{}}
		& & \multicolumn{5}{c}{$\npairs=3$} & \multicolumn{5}{c}{$\npairs=4$} & \\
		& & GT/Input & Ours & Calibrated~\cite{Yvain2018} & fastNFPS~\cite{lichy2022fast} & UniversalPS~\cite{ikehata2022universal} & GT & Ours & Calibrated~\cite{Yvain2018} & fastNFPS~\cite{lichy2022fast} & UniversalPS~\cite{ikehata2022universal} & \\
		\multirow{7}{*}{\raisebox{\halfFigwidthNormalVisSupp\linewidth}{\rotatebox[origin=c]{90}{\textsc{Bunny}}}} 
		& 
		\raisebox{\halfFigwidthNormalVisSupp\linewidth}{\rotatebox[origin=c]{90}{Normal map}}
		&
		\includegraphics[width=\figwidthNormalVisSupp\linewidth]{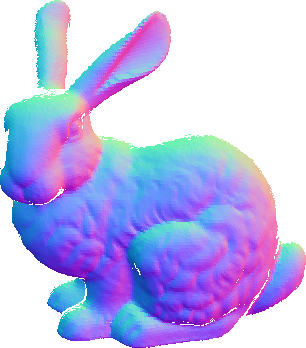}
		&
		\includegraphics[width=\figwidthNormalVisSupp\linewidth]{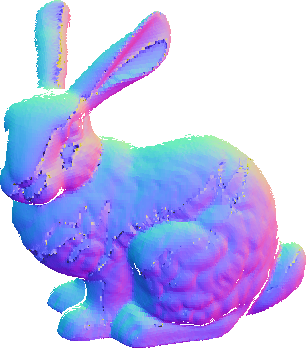}
		&
		\includegraphics[width=\figwidthNormalVisSupp\linewidth]{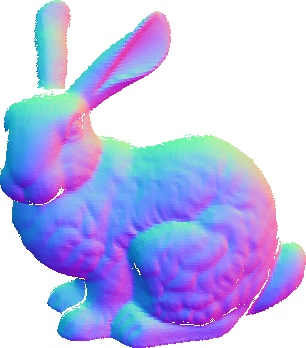}
		&
		\includegraphics[width=\figwidthNormalVisSupp\linewidth]{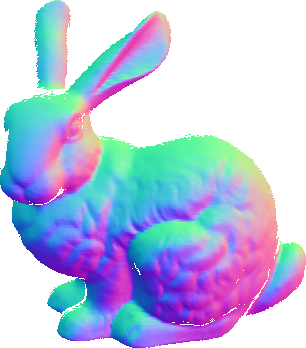}
		&
		\includegraphics[width=\figwidthNormalVisSupp\linewidth]{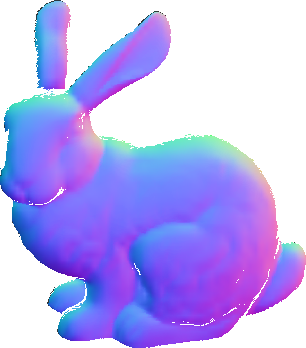}
		&
		\includegraphics[width=\figwidthNormalVisSupp\linewidth]{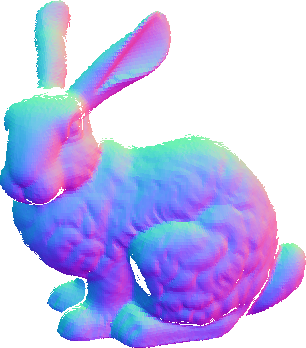}
		&
		\includegraphics[width=\figwidthNormalVisSupp\linewidth]{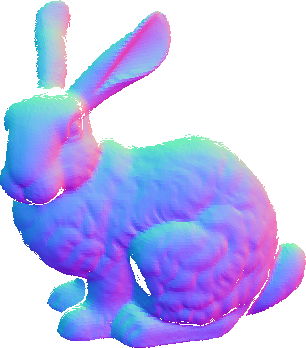}
		&
		\includegraphics[width=\figwidthNormalVisSupp\linewidth]{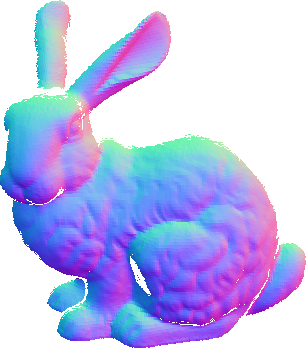}
		&
		\includegraphics[width=\figwidthNormalVisSupp\linewidth]{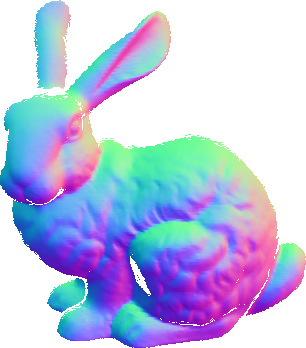}
		&
		\includegraphics[width=\figwidthNormalVisSupp\linewidth]{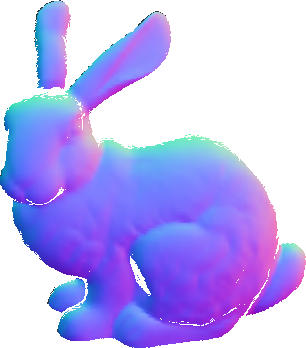}
		&
		\\
		&
		\raisebox{\halfFigwidthNormalVisSupp\linewidth}{\rotatebox[origin=c]{90}{Error map}}
		&
	    \includegraphics[width=\figwidthNormalVisSupp\linewidth]{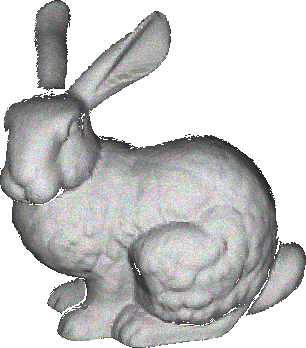}
		&
		\includegraphics[width=\figwidthNormalVisSupp\linewidth]{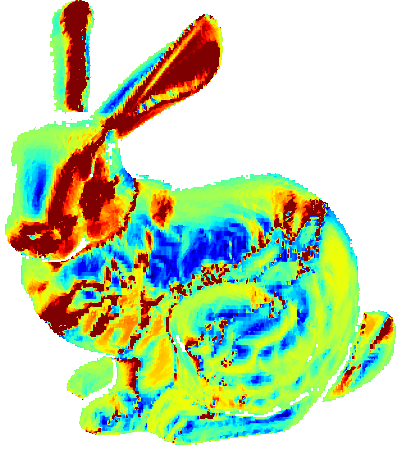}
		&
		\includegraphics[width=\figwidthNormalVisSupp\linewidth]{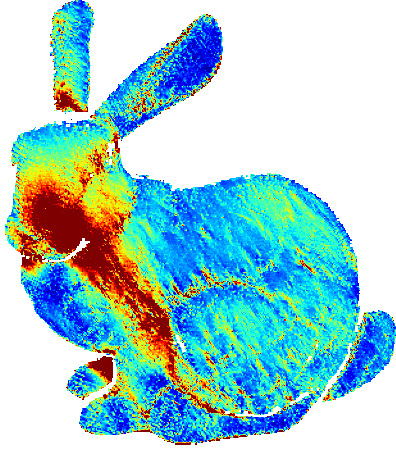}
		&
		\includegraphics[width=\figwidthNormalVisSupp\linewidth]{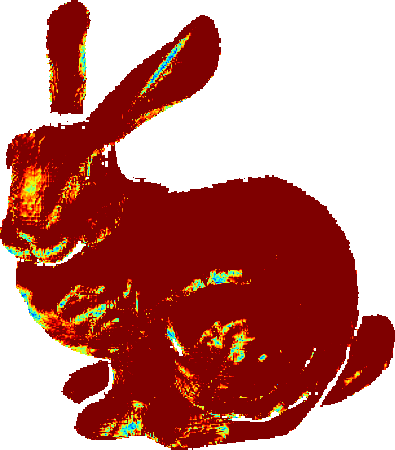}
		&
		\includegraphics[width=\figwidthNormalVisSupp\linewidth]{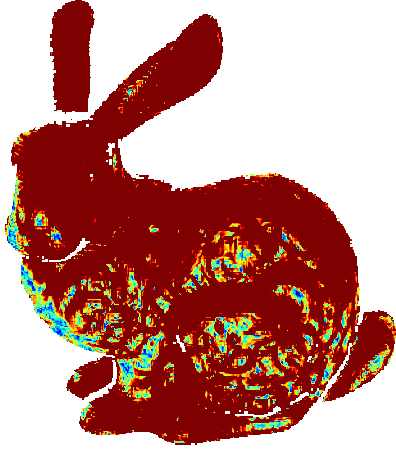}
		&
	    \includegraphics[width=\figwidthNormalVisSupp\linewidth]{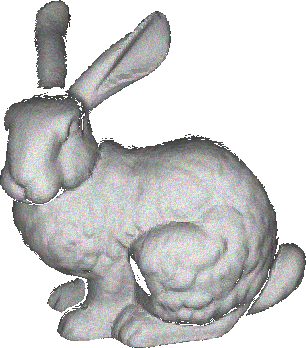}
		&
		\includegraphics[width=\figwidthNormalVisSupp\linewidth]{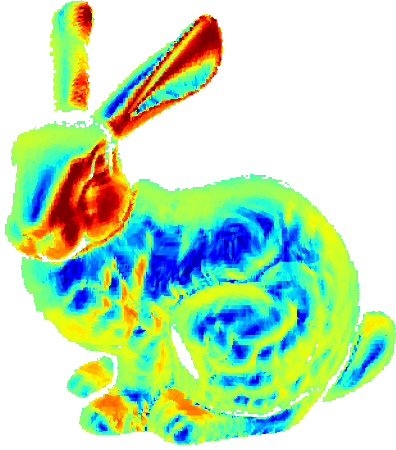}
		&
		\includegraphics[width=\figwidthNormalVisSupp\linewidth]{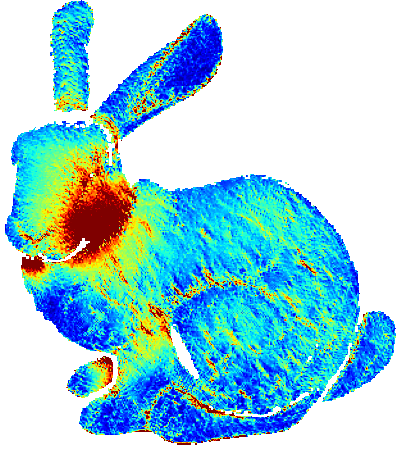}
		&
		\includegraphics[width=\figwidthNormalVisSupp\linewidth]{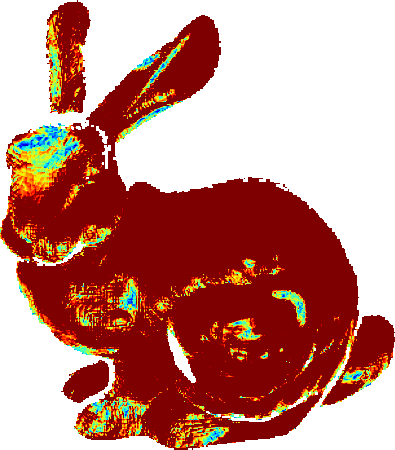}
		&
		\includegraphics[width=\figwidthNormalVisSupp\linewidth]{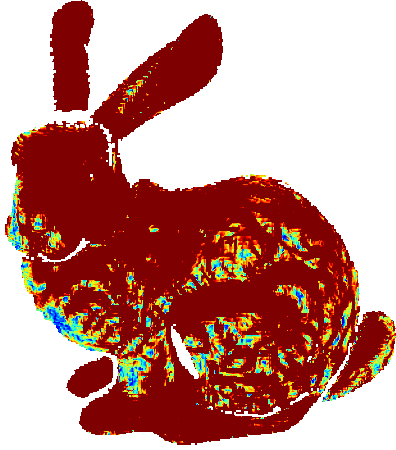}
		&
		\\
		& & & 7.416 & 4.704 & 18.487 & 18.089 &   & 5.116 & 4.122 & 16.641 & 17.536 & \\
		 &
		 \raisebox{\halfFigwidthNormalVisSupp\linewidth}{\rotatebox[origin=c]{90}{Depth map}}	
		 &
		\includegraphics[width=\figwidthNormalVisSupp\linewidth]{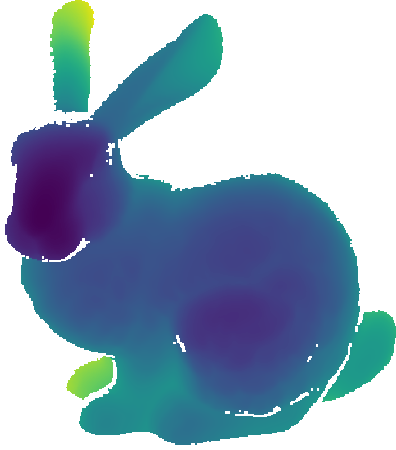}
		&
		\includegraphics[width=\figwidthNormalVisSupp\linewidth]{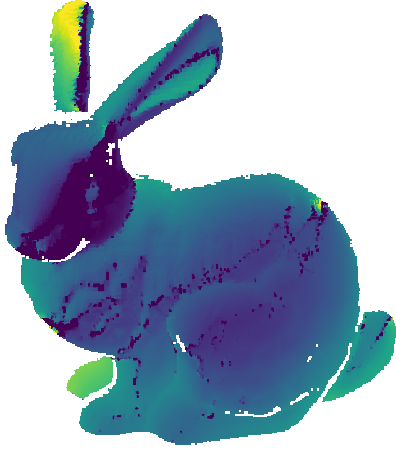}
		&
		\includegraphics[width=\figwidthNormalVisSupp\linewidth]{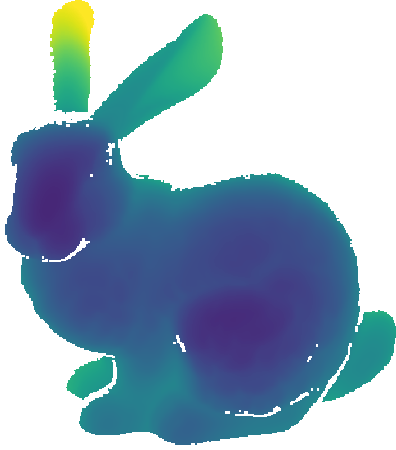}
		&
		\includegraphics[width=\figwidthNormalVisSupp\linewidth]{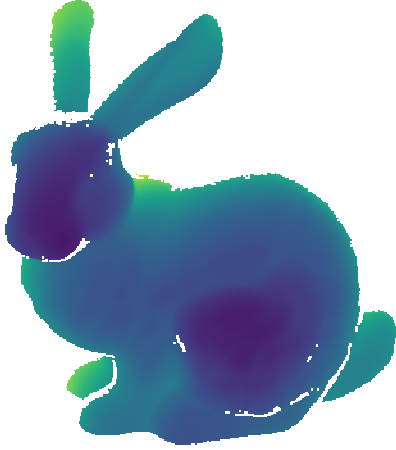}
		& 
		\multirow{2}{*}{
			\rotatebox[origin=c]{-10}{\textcolor{lightGray}{\rule{.1pt}{40pt}}}
		}
		&
		\includegraphics[width=\figwidthNormalVisSupp\linewidth]{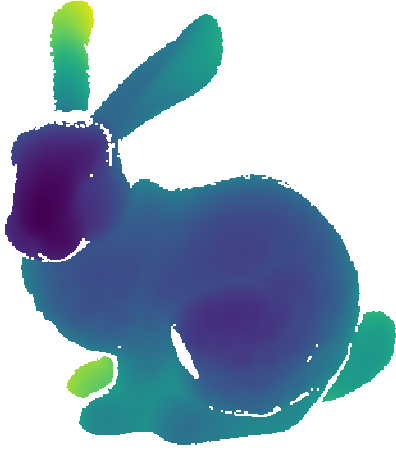}
		&
		\includegraphics[width=\figwidthNormalVisSupp\linewidth]{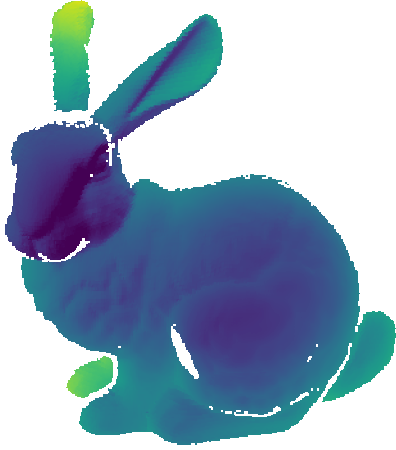}
		&
		\includegraphics[width=\figwidthNormalVisSupp\linewidth]{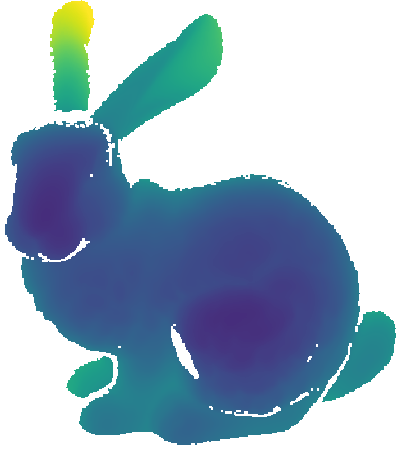}
		&
		\includegraphics[width=\figwidthNormalVisSupp\linewidth]{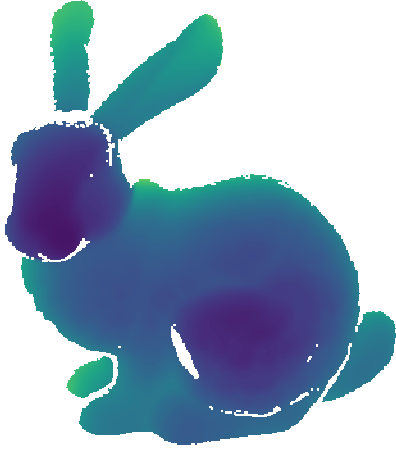}
		& 
		\multirow{2}{*}{
			\rotatebox[origin=c]{-10}{\textcolor{lightGray}{\rule{.1pt}{40pt}}}
		}
		\\
		&
		\raisebox{\halfFigwidthNormalVisSupp\linewidth}{\rotatebox[origin=c]{90}{Error map}}	
		&
		 \includegraphics[width=\figwidthNormalVisSupp\linewidth]{figures/results/oPlusLine45_lambertian_SValbedo_bunny_float32_offset_0.3x0.4x0.5_p3/sample_image.png}
		&
		\includegraphics[width=\figwidthNormalVisSupp\linewidth]{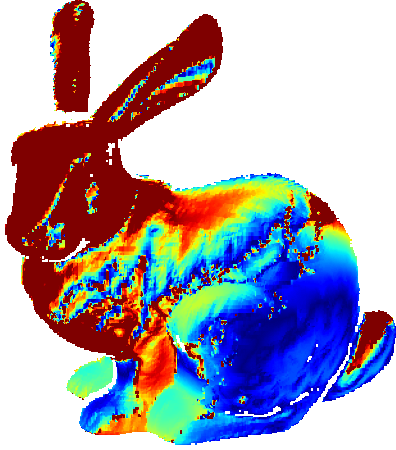}
		&
		\includegraphics[width=\figwidthNormalVisSupp\linewidth]{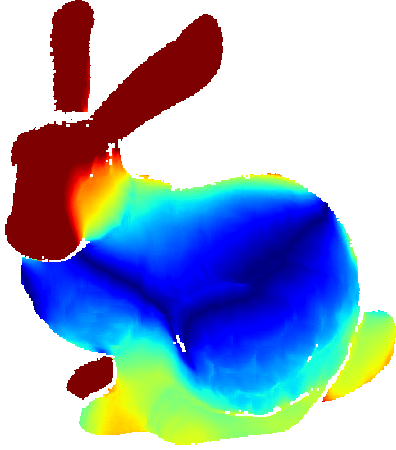}
		&
		\includegraphics[width=\figwidthNormalVisSupp\linewidth]{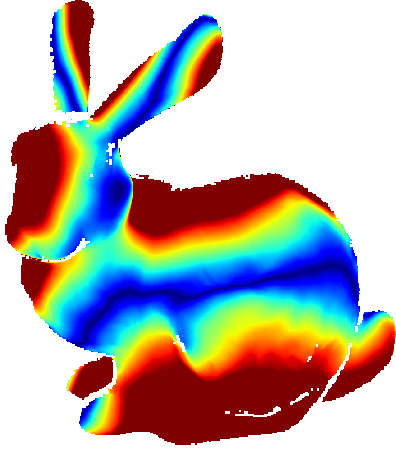}
		&
		&
		\includegraphics[width=\figwidthNormalVisSupp\linewidth]{figures/results/doubleO45_lambertian_SValbedo_bunny_float32_offset_0.3x0.4x0.5_p3/sample_image.png} 
		&
		\includegraphics[width=\figwidthNormalVisSupp\linewidth]{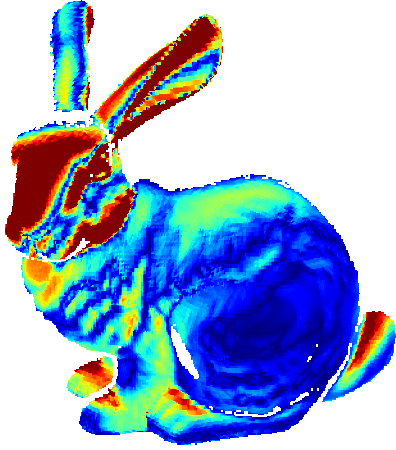}
		&
		\includegraphics[width=\figwidthNormalVisSupp\linewidth]{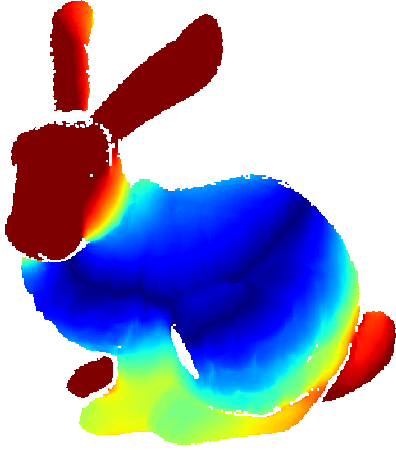}
		&
		\includegraphics[width=\figwidthNormalVisSupp\linewidth]{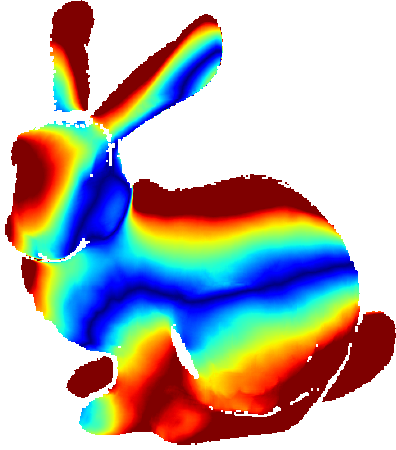}
		&
		&
		\\
		&
		&& 0.077 & 0.033 & 0.042 &  &  & 0.025 & 0.033 & 0.042 & \\
		\multirow{6}{*}{\raisebox{\halfFigwidthNormalVisSupp\linewidth}{\rotatebox[origin=c]{90}{\textsc{Crab}}}} & 
		\raisebox{\halfFigwidthNormalVisSupp\linewidth}{\rotatebox[origin=c]{90}{Normal map}}	
		&
		 \includegraphics[width=\figwidthNormalVisSupp\linewidth]{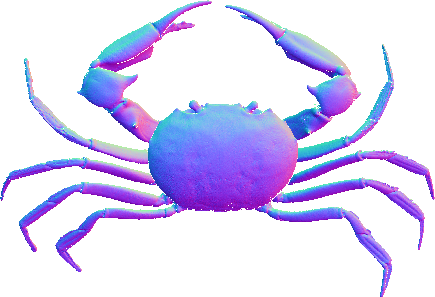}
		 &
		 \includegraphics[width=\figwidthNormalVisSupp\linewidth]{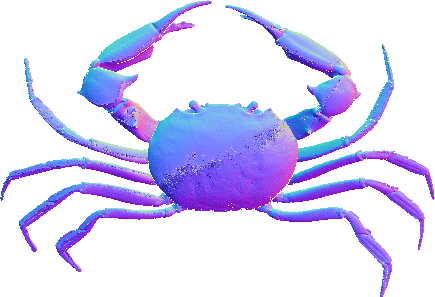}
		 &
		 \includegraphics[width=\figwidthNormalVisSupp\linewidth]{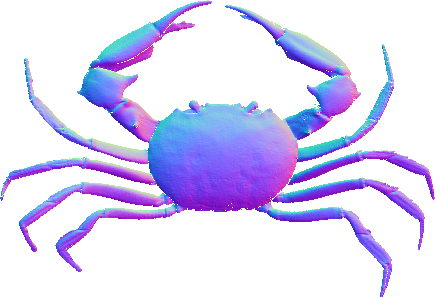}
		 &
		 \includegraphics[width=\figwidthNormalVisSupp\linewidth]{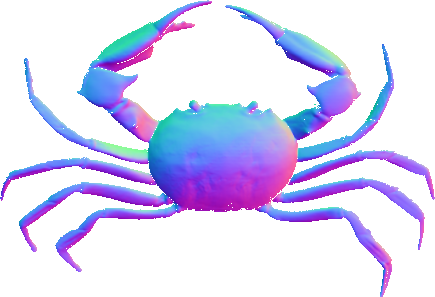}
		 &
		 \includegraphics[width=\figwidthNormalVisSupp\linewidth]{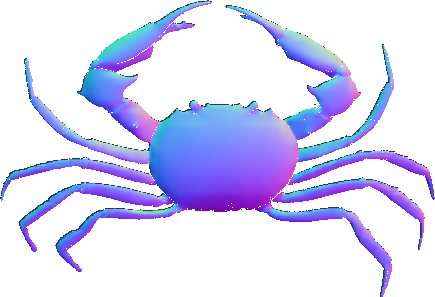}
		 &
		 \includegraphics[width=\figwidthNormalVisSupp\linewidth]{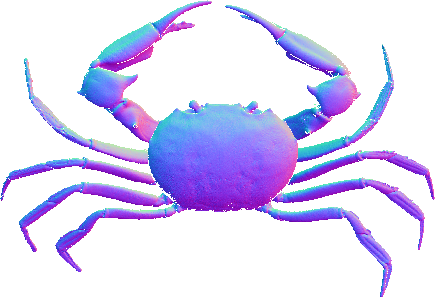}
		 &
		 \includegraphics[width=\figwidthNormalVisSupp\linewidth]{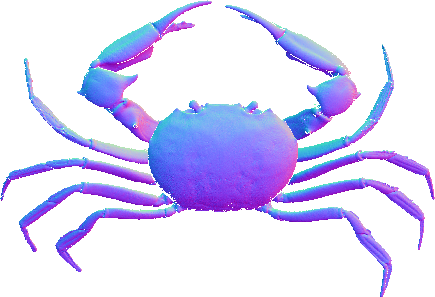}
		 &
		 \includegraphics[width=\figwidthNormalVisSupp\linewidth]{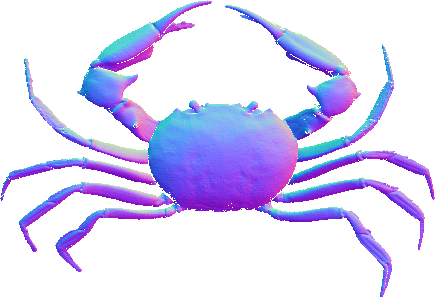}
		 &
		 \includegraphics[width=\figwidthNormalVisSupp\linewidth]{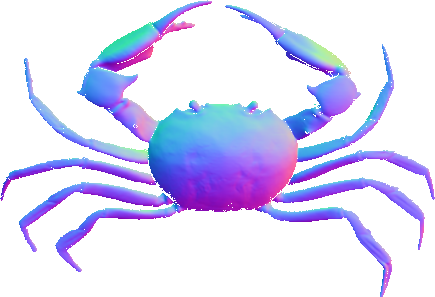}
		 &
		 \includegraphics[width=\figwidthNormalVisSupp\linewidth]{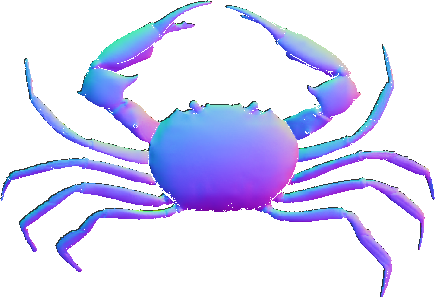}
		 &
		 \\
		 &
		 \raisebox{\halfFigwidthNormalVisSupp\linewidth}{\rotatebox[origin=c]{90}{Error map}}	
		 &
		 \includegraphics[width=\figwidthNormalVisSupp\linewidth]{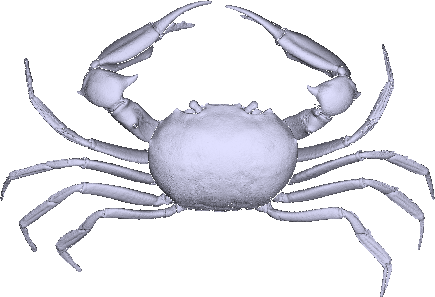}
		 &
		 \includegraphics[width=\figwidthNormalVisSupp\linewidth]{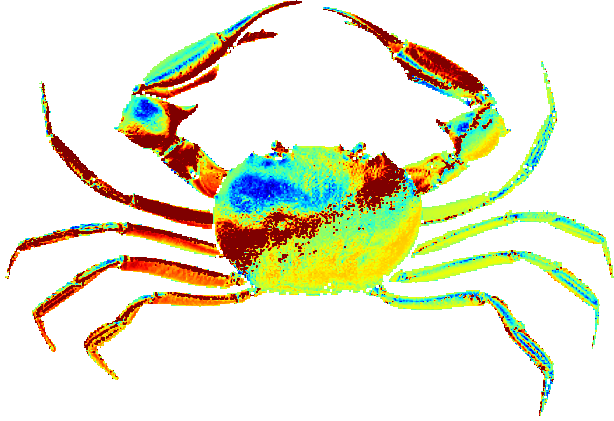}
		 &
		 \includegraphics[width=\figwidthNormalVisSupp\linewidth]{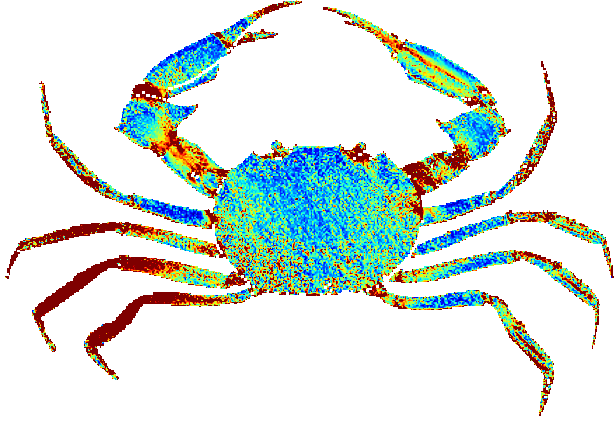}
		 &
		 \includegraphics[width=\figwidthNormalVisSupp\linewidth]{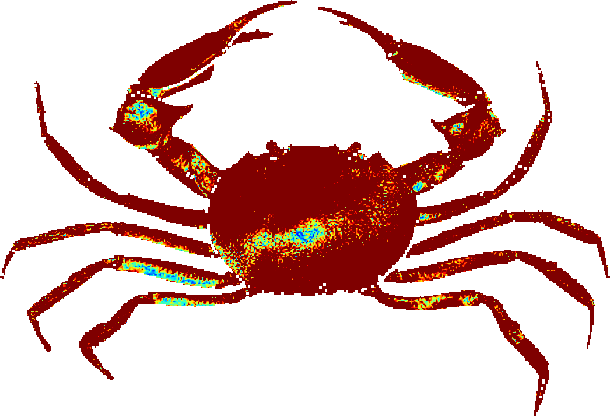}
		 &
		 \includegraphics[width=\figwidthNormalVisSupp\linewidth]{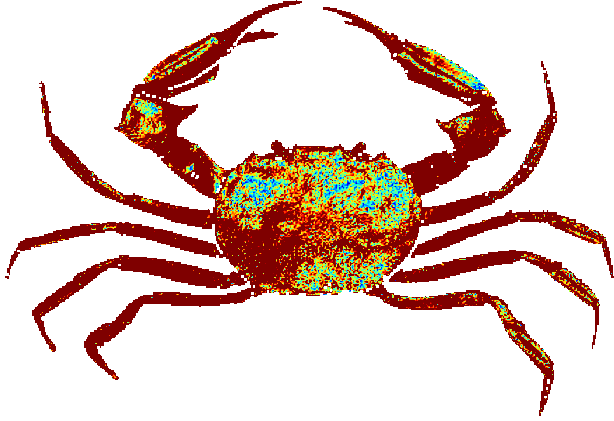}
		 &
		 \includegraphics[width=\figwidthNormalVisSupp\linewidth]{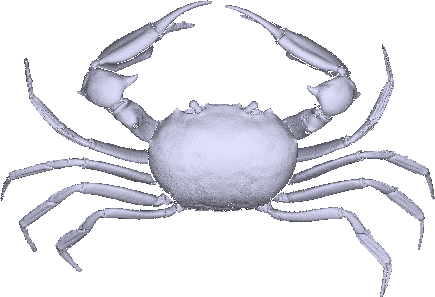}
		 &
		 \includegraphics[width=\figwidthNormalVisSupp\linewidth]{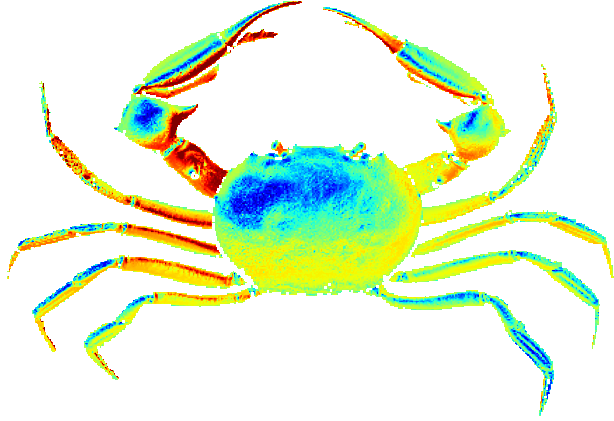}
		 &
		 \includegraphics[width=\figwidthNormalVisSupp\linewidth]{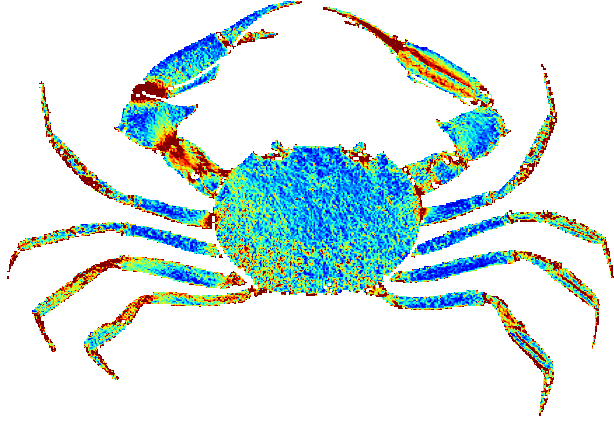}
		 &
		 \includegraphics[width=\figwidthNormalVisSupp\linewidth]{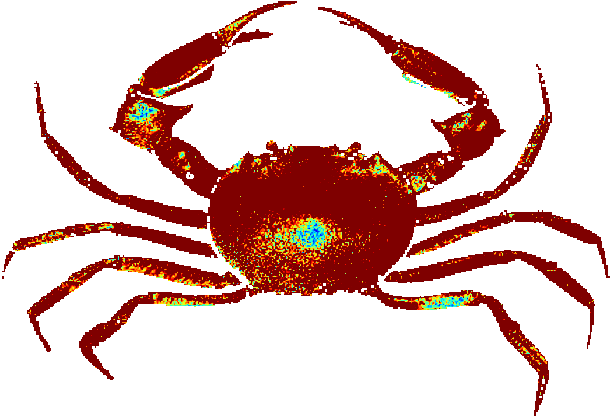}
		 &
		 \includegraphics[width=\figwidthNormalVisSupp\linewidth]{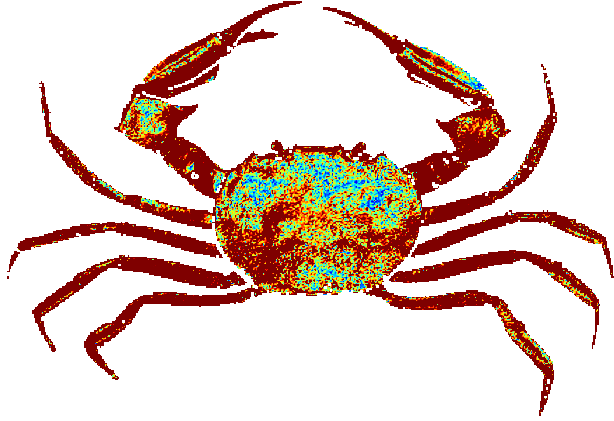}
		 &
		 \\
		& & & 8.555 & 7.375 & 17.692 & 20.708  &  &  5.504 & 5.612 & 16.257 & 18.873 & \\
		&
				\raisebox{\halfFigwidthNormalVisSupp\linewidth}{\rotatebox[origin=c]{90}{Depth map}}	
		&
		 \includegraphics[width=\figwidthNormalVisSupp\linewidth]{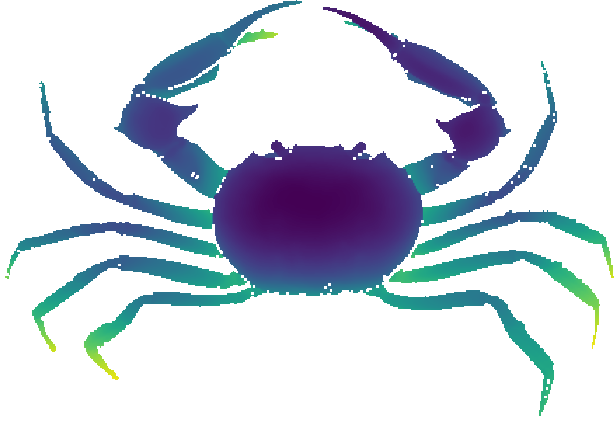}
		 &
		 \includegraphics[width=\figwidthNormalVisSupp\linewidth]{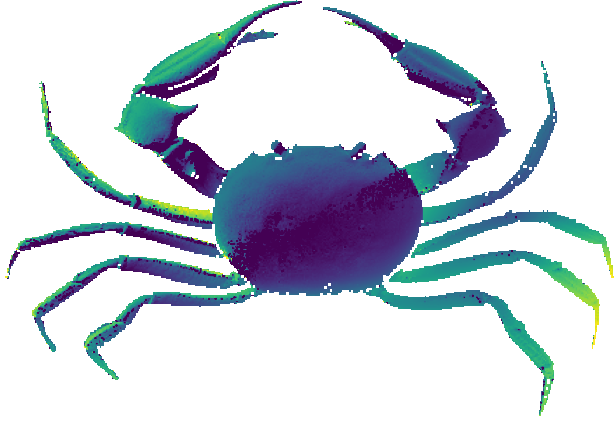}
		 &
		 \includegraphics[width=\figwidthNormalVisSupp\linewidth]{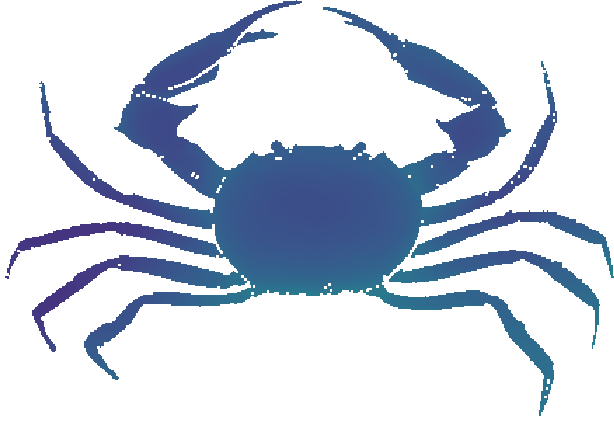}
		 &
		 \includegraphics[width=\figwidthNormalVisSupp\linewidth]{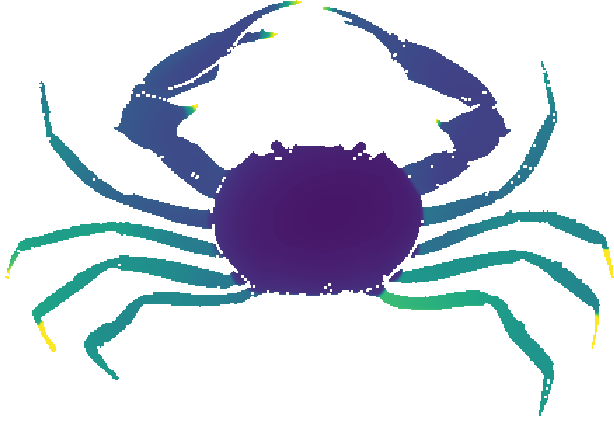}
		 & 
		 \multirow{2}{*}{
			\rotatebox[origin=c]{-10}{\textcolor{lightGray}{\rule{.1pt}{40pt}}}
		} 
		 &
		 \includegraphics[width=\figwidthNormalVisSupp\linewidth]{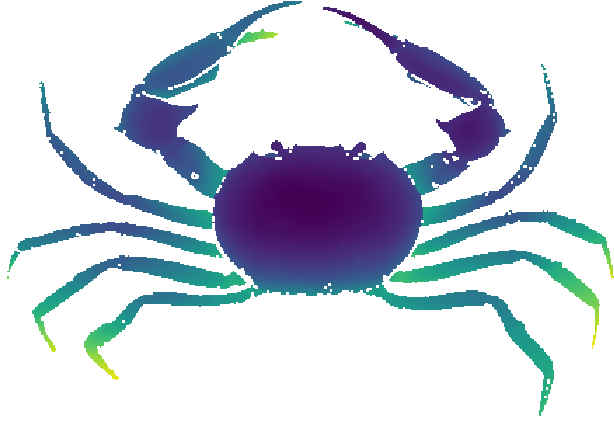}
		 &
		 \includegraphics[width=\figwidthNormalVisSupp\linewidth]{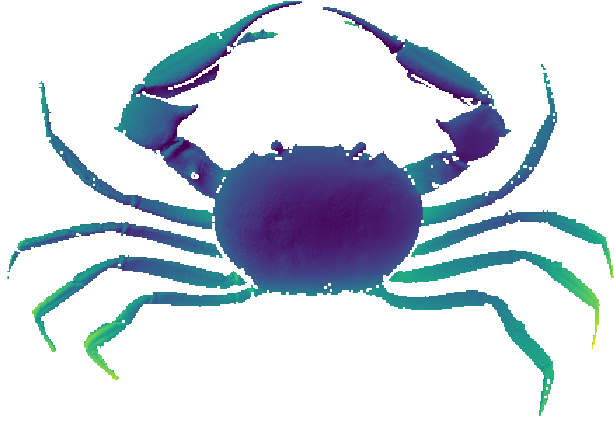}
		 &
		 \includegraphics[width=\figwidthNormalVisSupp\linewidth]{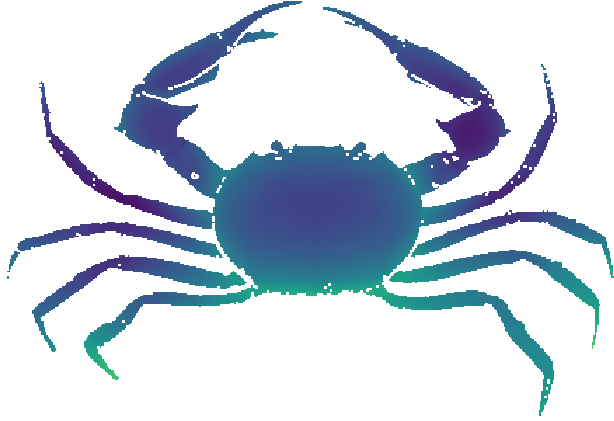}
		 &
		 \includegraphics[width=\figwidthNormalVisSupp\linewidth]{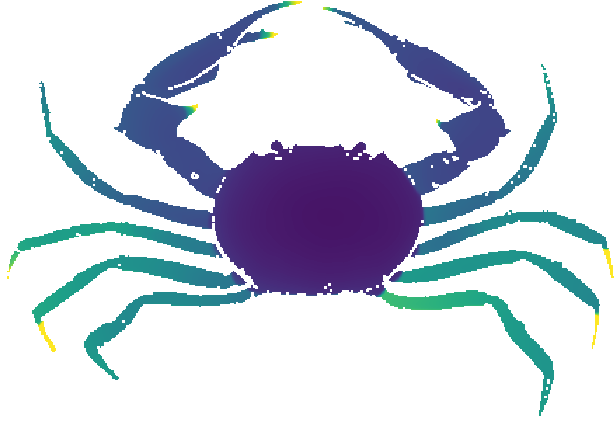}
		 & 
		 \multirow{2}{*}{
			\rotatebox[origin=c]{-10}{\textcolor{lightGray}{\rule{.1pt}{40pt}}}
		}
		 \\
		 &
				\raisebox{\halfFigwidthNormalVisSupp\linewidth}{\rotatebox[origin=c]{90}{Error map}}	
		&
		 
		  \includegraphics[width=\figwidthNormalVisSupp\linewidth]{figures/results/oPlusLine45_lambertian_crab_float32_offset_0.3x0.4x0.5_p3/sample_image.png}
		 &
		 \includegraphics[width=\figwidthNormalVisSupp\linewidth]{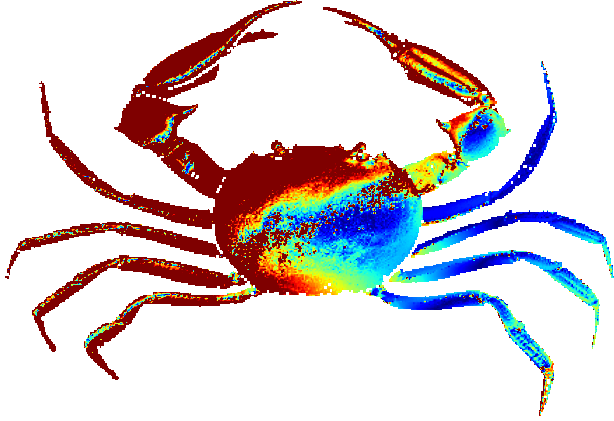}
		 &
		 \includegraphics[width=\figwidthNormalVisSupp\linewidth]{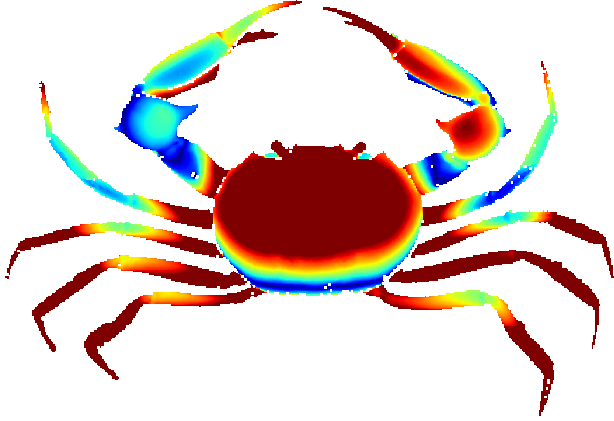}
		 &
		 \includegraphics[width=\figwidthNormalVisSupp\linewidth]{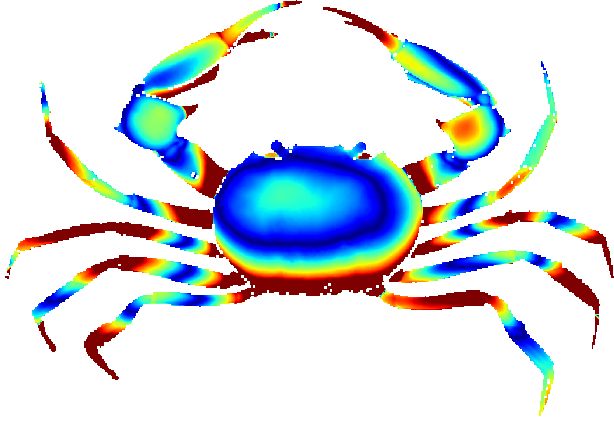}
		 &
		 &
		 \includegraphics[width=\figwidthNormalVisSupp\linewidth]{figures/results/doubleO45_lambertian_crab_float32_offset_0.3x0.4x0.5_p3/sample_image.png} 
		 &
		 \includegraphics[width=\figwidthNormalVisSupp\linewidth]{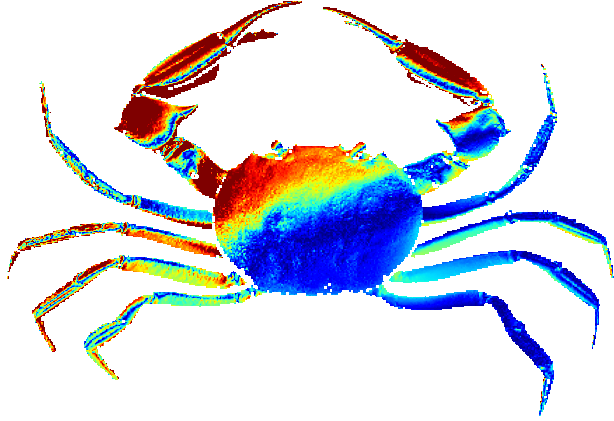}
		 &
		 \includegraphics[width=\figwidthNormalVisSupp\linewidth]{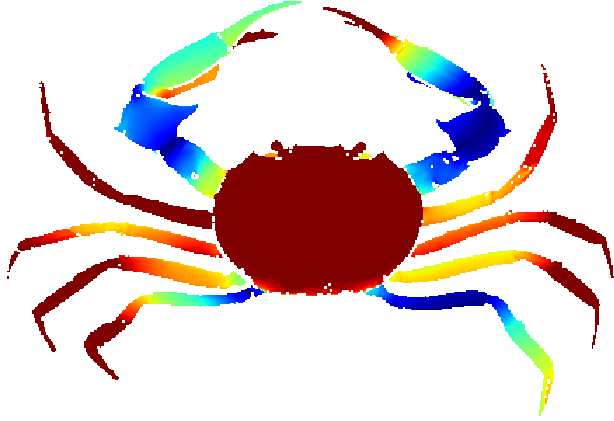}
		 &
		 \includegraphics[width=\figwidthNormalVisSupp\linewidth]{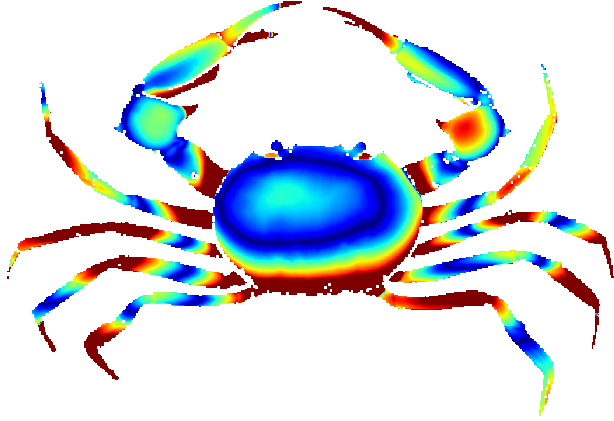}
		 &
		 &
		 \\
		 & 
		 && 0.088 & 0.058 & 0.037 &   &   & 0.030 & 0.050 & 0.038 & \\
		& \\
		 &&\rotatedColorbar{Normal error}{0.065}{$10^\circ$}{-15} & 
		 \rotatedColorbar{Depth error}{0.065}{$0.01$}{-15} & 
	\end{tabular}
	\caption{Estimation results of our method and comparison methods, with the global offset along all the $xyz$-axis $\lo=\left[0.3,0.4,0.5\right]$ added to the center of the symmetric light pairs. 
	 }
	\vspace*{-10pt}
	\label{fig:synth_results_xyz_shift}
\end{figure*}

\begin{figure}[h!]
	\scriptsize
	\newcommand{\figwidthNormalVisSupp}{0.23}
	\newcommand{\halfFigwidthNormalVisSupp}{0.115}
	\centering
	\begin{tabular}{@{}c@{}c@{}c@{}c@{}c@{}c@{}}
		& GT/Input & Ours & Calibrated~\cite{Yvain2018} & fastNFPS~\cite{lichy2022fast} &  \\
		\multicolumn{5}{c}{\textsc{Bunny}} \\
		\raisebox{\halfFigwidthNormalVisSupp\linewidth}{\rotatebox[origin=c]{90}{Depth map}}
		&
		\includegraphics[width=\figwidthNormalVisSupp\linewidth]{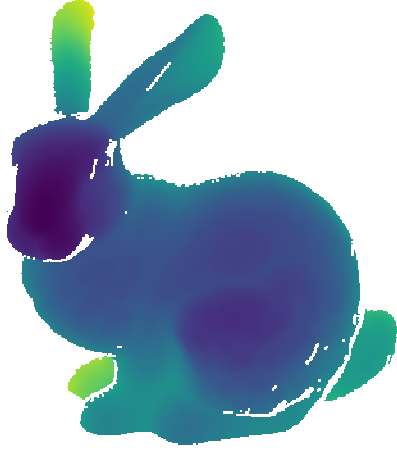}
		&
		\includegraphics[width=\figwidthNormalVisSupp\linewidth]{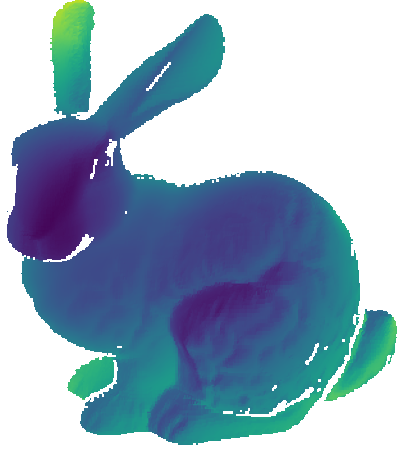}
		&
		\includegraphics[width=\figwidthNormalVisSupp\linewidth]{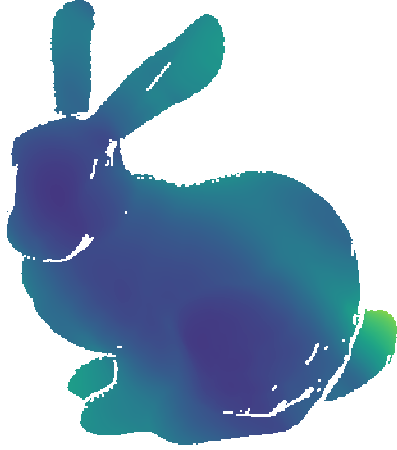}
		&
		\includegraphics[width=\figwidthNormalVisSupp\linewidth]{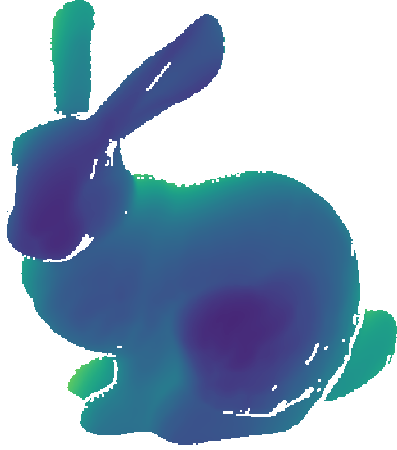}
		& 
		\\
		\raisebox{\halfFigwidthNormalVisSupp\linewidth}{\rotatebox[origin=c]{90}{Error map}}
		&
		\includegraphics[width=\figwidthNormalVisSupp\linewidth]{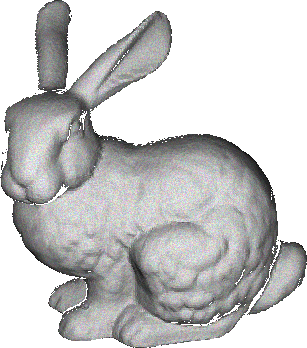}
		&
		\includegraphics[width=\figwidthNormalVisSupp\linewidth]{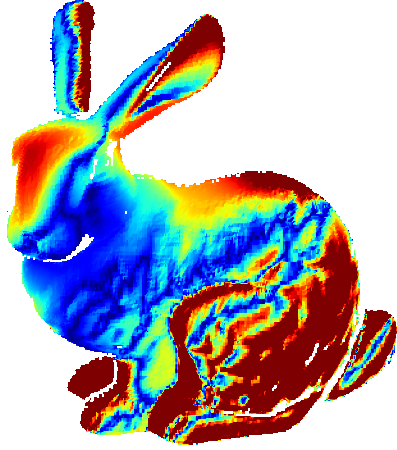}
		&
		\includegraphics[width=\figwidthNormalVisSupp\linewidth]{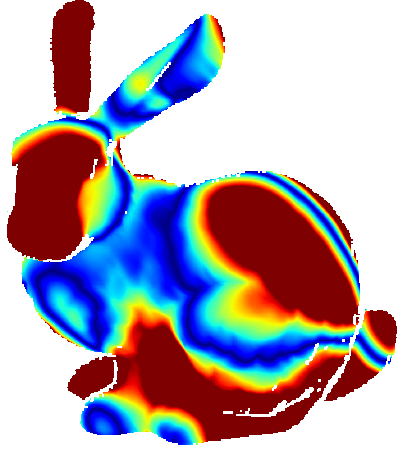}
		&
		\includegraphics[width=\figwidthNormalVisSupp\linewidth]{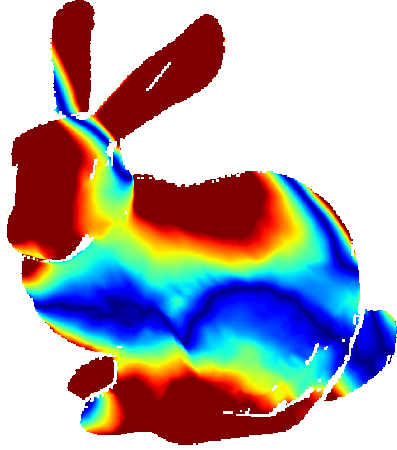}
		&
		\colorbar{\figwidthNormalVisSupp}{$0.01$}{0.18}
		\\
		 & & 0.038 & 0.055 & 0.050 & \\
		 \multicolumn{5}{c}{\textsc{Crab}} \\
		 \raisebox{0.09\linewidth}{\rotatebox[origin=c]{90}{Depth map}}
		 &
		\includegraphics[width=\figwidthNormalVisSupp\linewidth]{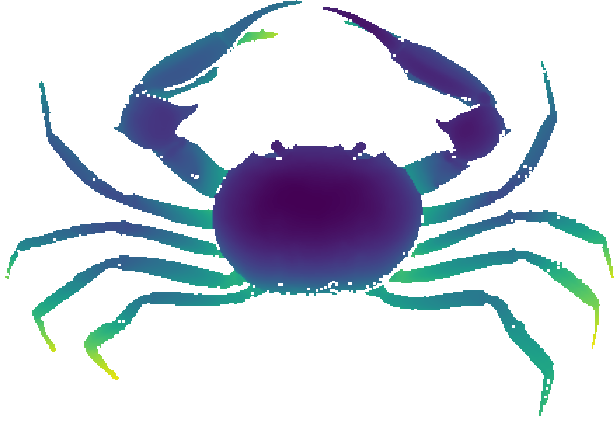}
		&
		\includegraphics[width=\figwidthNormalVisSupp\linewidth]{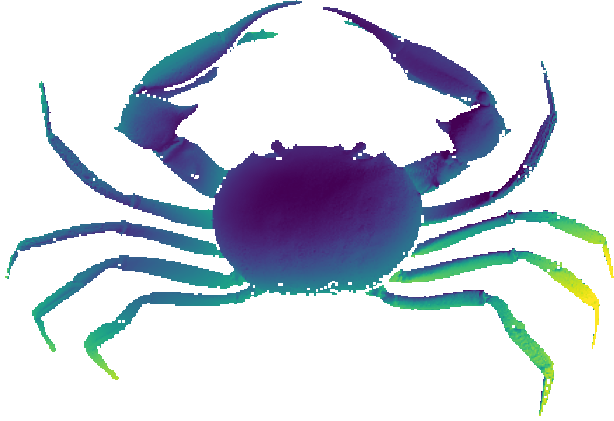}
		&
		\includegraphics[width=\figwidthNormalVisSupp\linewidth]{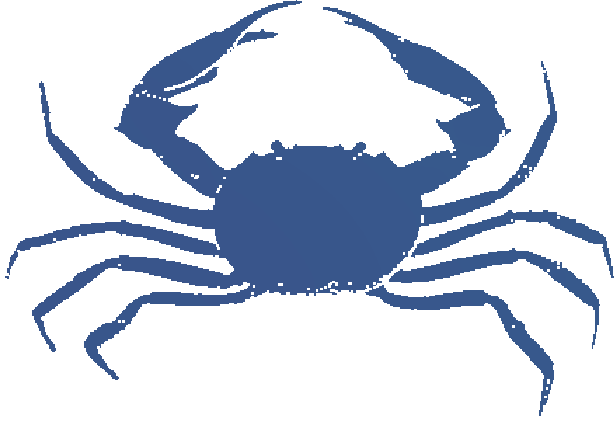}
		&
		\includegraphics[width=\figwidthNormalVisSupp\linewidth]{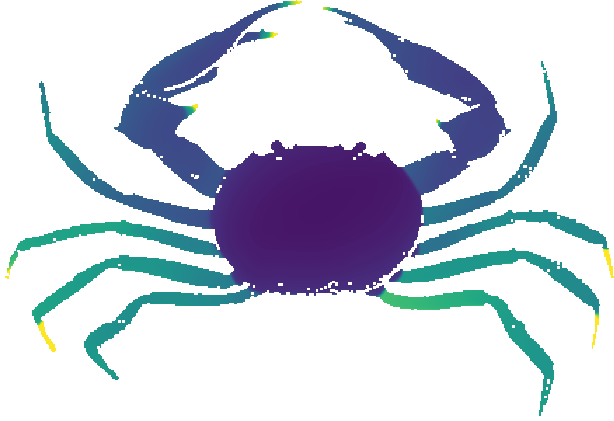}
		& 
		\\
		\raisebox{0.09\linewidth}{\rotatebox[origin=c]{90}{Error map}}
		&
		\includegraphics[width=\figwidthNormalVisSupp\linewidth]{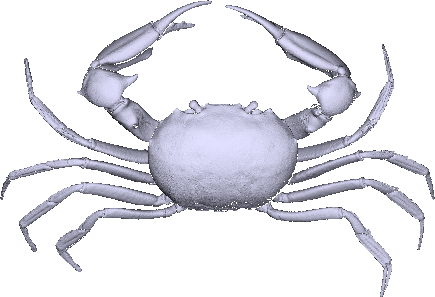}
		&
		\includegraphics[width=\figwidthNormalVisSupp\linewidth]{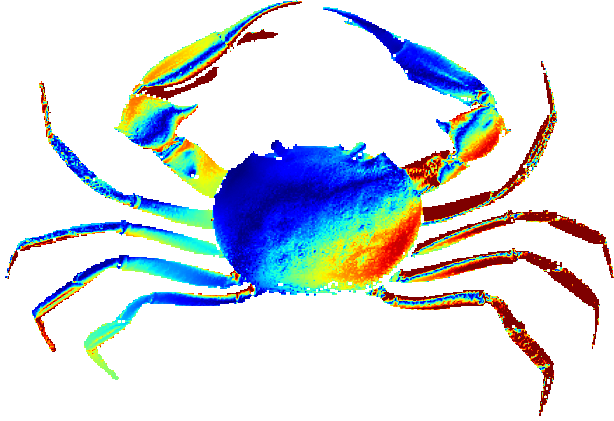}
		&
		\includegraphics[width=\figwidthNormalVisSupp\linewidth]{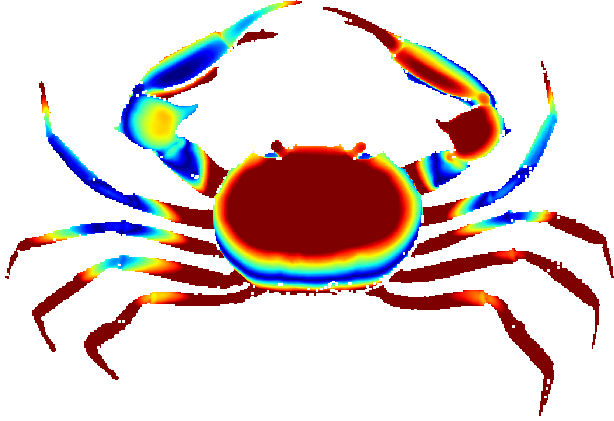}
		&
		\includegraphics[width=\figwidthNormalVisSupp\linewidth]{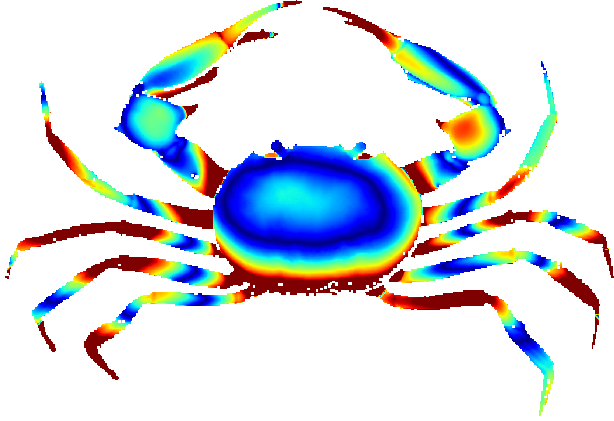}
		&
		 \colorbar{\halfFigwidthNormalVisSupp}{$0.01$}{0.08}
		\\
		&  & 0.031 & 0.059 & 0.037 & \\
	\end{tabular}
	\caption{Depth estimation results with $\npairs=2$ case. We add the global offset along the $z$-axis as $\lo=\left[0.0,0.0,0.5\right]$. 
	We show a mean relative absolute error at the bottom of each error map.
	}
	\vspace*{-10pt}
	\label{fig:synth_depth_results_fourOnLine_z_shift}
\end{figure}

Since $\V{A}$, $\V{A}'_{\eqia/\eqib}$ and $\V{A}'_{\eqiia/\eqiib}$ are linearly independent of each other, we can estimate $\V{e}$ for all the possible light arrangements if the sum of the rank of constraints from \mbox{$\V{A}$, $\V{A}'_{\eqia/\eqib}$ and $\V{A}'_{\eqiia/\eqiib}$} is at least $2\npairs - 1$; namely, we can estimate $\V{e}$ with  $\npairs \ge 3$ for (i)$\sim$ (iii) cases and $\npairs \ge 2$ for (iv) case.
In summary, the minimum setting for scaled distance estimation for non-ring light cases \ie (i) and (iii) is that we have at least $\npairs = 3$. Although the scaled distance can be estimated in the case of a ring light, the depth cannot be estimated as described in the following section.

\pparag{Surface position estimation}
We then consider the solvability of surface point estimation given scaled distances.
At the end of this part, we show that the following condition should be satisfied.
\begin{condition}
At least two pairs with different radii must exist for the surface position estimation.
\end{condition}
\noindent
In short, this is due to the fact that the denominator of $\rho^{-1}r^2$ in \eref{eq:e_constraints1a} goes to $0$ when all radii are equal, so the surface point obtained from Eqs.~(\ref{eq:scaled_x})-(\ref{eq:scaled_shifted_z}), which includes $\rho^{-1}r^2$, cannot be recovered either.
The condition can also be explained by its geometric meaning.

Apollonius shows that the set of points that have a given ratio of distances to the fixed points lies in the circle~\cite{ogilvy1990excursions}.
The aforementioned theorem also holds in $\mathbb{R}^n$ case~(see Appendix: Proposition 1.); \iow with the two fixed points $\in \doubleR^{3}$ and its corresponding ratios, we get an Apollonius \emph{sphere}.

Now we have the known relative light positions $\frac{\vs_* - \lo}{r}$ and the estimated distance ratios $e_*$. Therefore, given two randomly selected lights, we can obtain an Apollonius sphere, where the unknown surface point $\point'_r$ lies.
Thus, the problem of solving for a surface point, given known relative light positions and their distance to the estimated shifted point, can be cast into the problem of estimating the intersection of Apollonius spheres using combinations of two lights chosen from ${2\npairs}$. Consequently, a point \point~ can be recovered if the spheres have a unique intersection. More details along with the proof of each proposition can be found in the supplementary material.

To facilitate the discussion, we present the following proposition, the proof of which is given in the appendix.
\setcounter{proposition}{1}
\begin{proposition}
    When the radii of the symmetric lights $r_*$ of all pairs are the same, all the possible Apollonius spheres are centered on a line, whose direction is $\left[-y',x',0\right]$. 
    \label{prop:spheres_on_line}
\end{proposition}
Assuming that the estimation of the scaled distances has no error, then the Apollonius spheres always have some common intersections.
If we let all 3D spheres lie on a line, the intersection obviously has an ambiguity along the 2D circle, and thus the unique surface point $\point'$ cannot be recovered. 
That is, if all the light pairs have a unique radius, \eg arranged on a ring, our \emph{relatively-calibrated} near-light photometric stereo problem cannot be solved. 
In contrast, we empirically observed symmetric light arrangements sharing the same center with at least two different radii can disambiguate the problem.
The examples of valid/invalid light arrangements are shown in \fref{fig:valid_arrangenemts}, which satisfies the condition for both scaled distance estimation and surface point estimation.
\begin{figure*}
	\centering	
	\newcommand{\figwidthNormalVisSupp}{0.09}
	\newcommand{\halfFigwidthNormalVisSupp}{0.04}
	\newcommand{\basename}{doubleO45_lambertian_bunny_float32_offset_0.0x0.0x4.0_p3}
	\resizebox{1.04\textwidth}{!}{
	\begin{tabular}{@{}c@{}c@{}c@{}|c@{}c@{}|c@{}c@{}|c@{}c@{}|c@{}c@{}c}
		\multicolumn{12}{c}{$\xleftarrow{
			\makebox[0.95\linewidth]{Median of light-to-surface distance $l$}
		}$} \\
		&
		\multicolumn{2}{c}{$l=6$} & \multicolumn{2}{c}{$l=5$} & \multicolumn{2}{c}{$l=4$} & \multicolumn{2}{c}{$l=3$} & \multicolumn{2}{c}{$l=2$} & 	
		\\
		\raisebox{\halfFigwidthNormalVisSupp\linewidth}{\rotatebox[origin=c]{90}{Input image}}
		&
		\multicolumn{2}{c}{\includegraphics[width=\figwidthNormalVisSupp\linewidth]{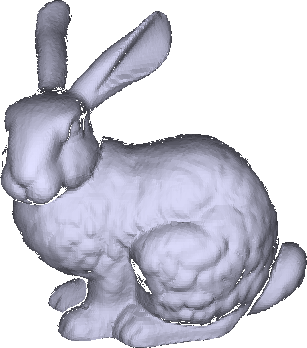}}
		&
		\multicolumn{2}{c}{\includegraphics[width=\figwidthNormalVisSupp\linewidth]{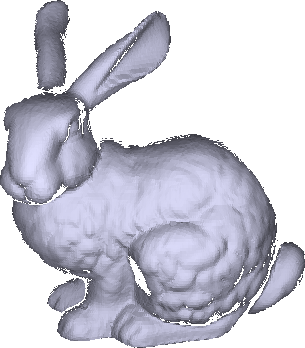}}
		&
		\multicolumn{2}{c}{\includegraphics[width=\figwidthNormalVisSupp\linewidth]{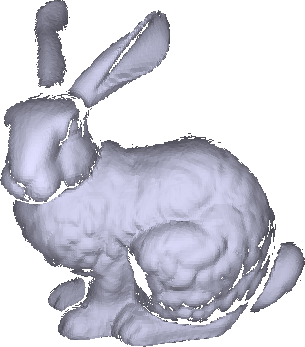}}
		&
		\multicolumn{2}{c}{\includegraphics[width=\figwidthNormalVisSupp\linewidth]{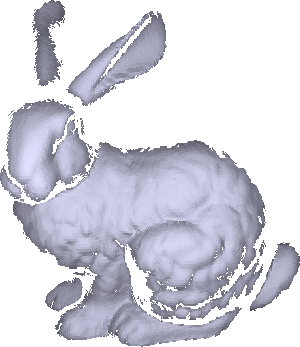}}
		&
		\multicolumn{2}{c}{\includegraphics[width=\figwidthNormalVisSupp\linewidth]{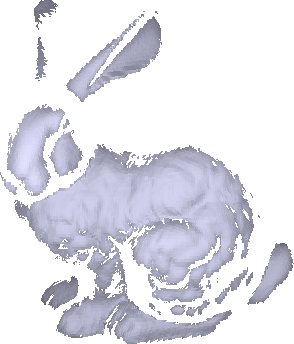}}
		& 
		\\
		& Ours & Calibrated~\cite{Yvain2018}& Ours & Calibrated~\cite{Yvain2018}&  Ours & Calibrated~\cite{Yvain2018}& Ours & Calibrated~\cite{Yvain2018}&Ours & Calibrated~\cite{Yvain2018}&  \\ 
		\raisebox{\halfFigwidthNormalVisSupp\linewidth}{\rotatebox[origin=c]{90}{Normal map}}
		&
		\includegraphics[width=\figwidthNormalVisSupp\linewidth]{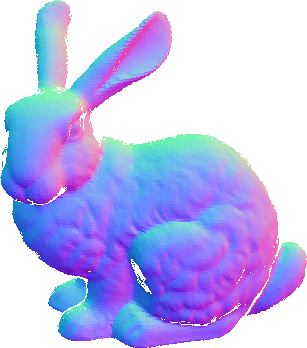}
		&
		\includegraphics[width=\figwidthNormalVisSupp\linewidth]{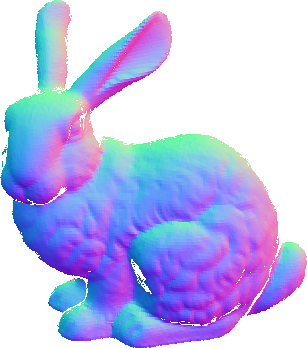}
		&
		\includegraphics[width=\figwidthNormalVisSupp\linewidth]{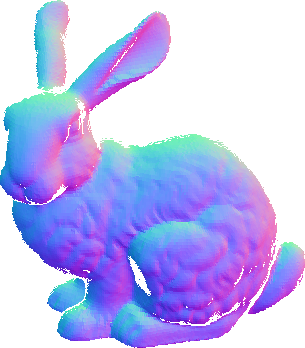}
		&
		\includegraphics[width=\figwidthNormalVisSupp\linewidth]{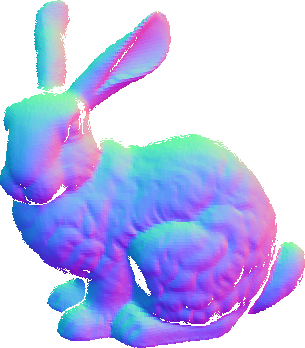}
		&
		\includegraphics[width=\figwidthNormalVisSupp\linewidth]{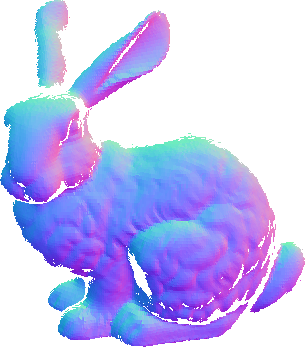}
		&
		\includegraphics[width=\figwidthNormalVisSupp\linewidth]{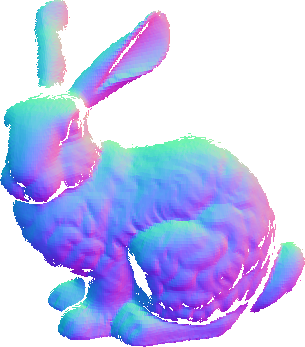}
		&
		\includegraphics[width=\figwidthNormalVisSupp\linewidth]{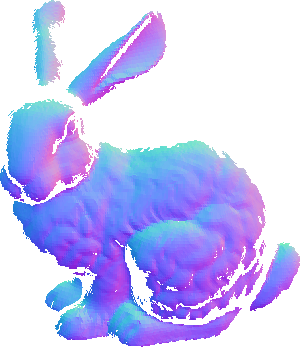}
		&
		\includegraphics[width=\figwidthNormalVisSupp\linewidth]{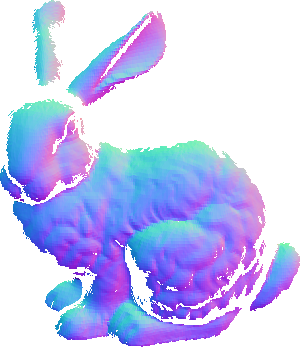}
		&
		\includegraphics[width=\figwidthNormalVisSupp\linewidth]{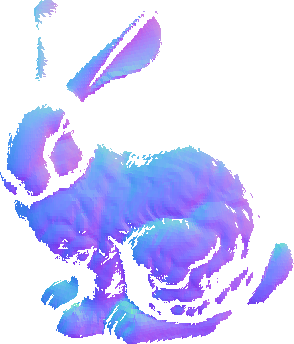}
		&
		\includegraphics[width=\figwidthNormalVisSupp\linewidth]{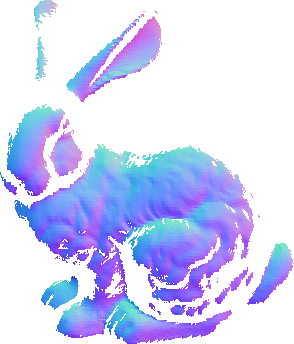}
		& 
		\\
		\raisebox{\halfFigwidthNormalVisSupp\linewidth}{\rotatebox[origin=c]{90}{Error map}}
		&
		\includegraphics[width=\figwidthNormalVisSupp\linewidth]{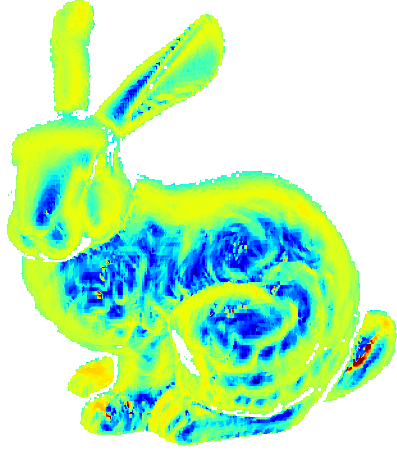}
		&
		\includegraphics[width=\figwidthNormalVisSupp\linewidth]{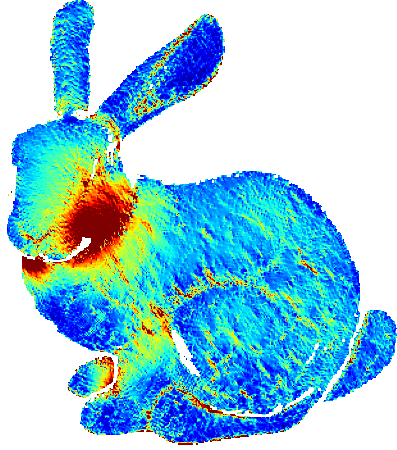}
		&
		\includegraphics[width=\figwidthNormalVisSupp\linewidth]{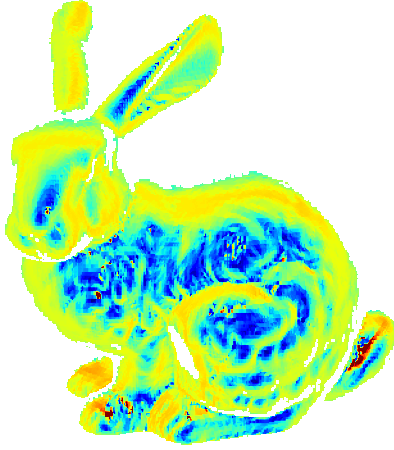}
		&
		\includegraphics[width=\figwidthNormalVisSupp\linewidth]{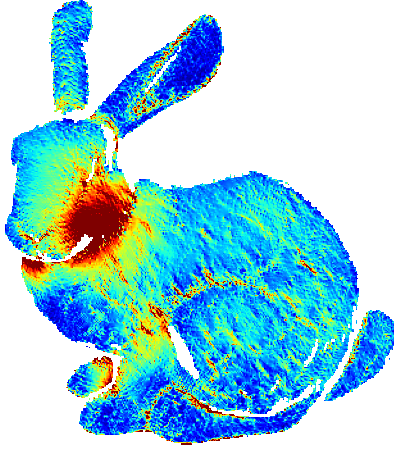}
		&
		\includegraphics[width=\figwidthNormalVisSupp\linewidth]{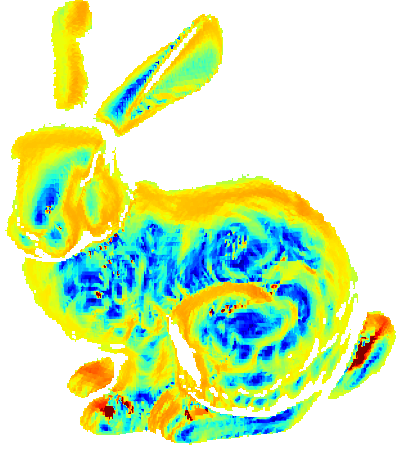}
		&
		\includegraphics[width=\figwidthNormalVisSupp\linewidth]{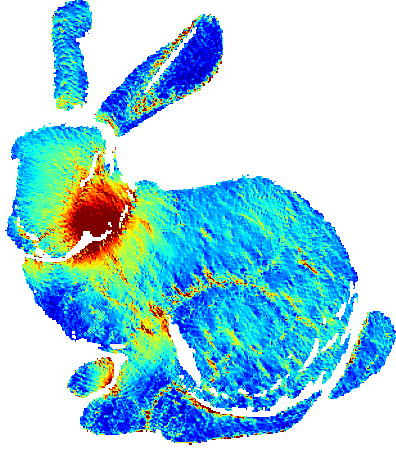}
		&
		\includegraphics[width=\figwidthNormalVisSupp\linewidth]{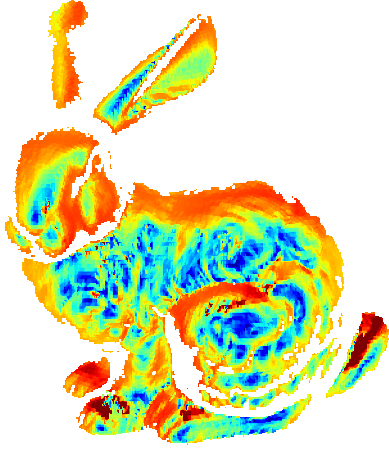}
		&
		\includegraphics[width=\figwidthNormalVisSupp\linewidth]{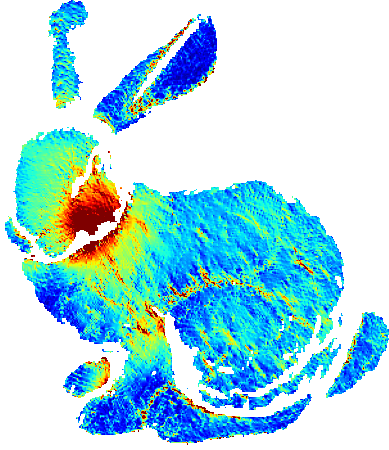}
		&
		\includegraphics[width=\figwidthNormalVisSupp\linewidth]{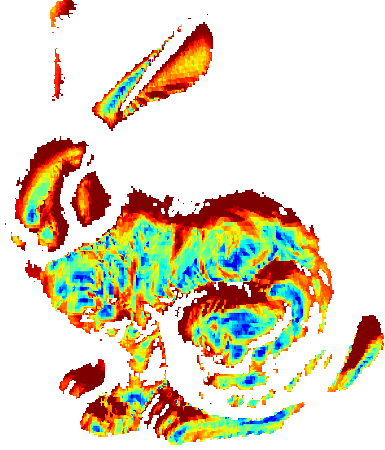}
		&
		\includegraphics[width=\figwidthNormalVisSupp\linewidth]{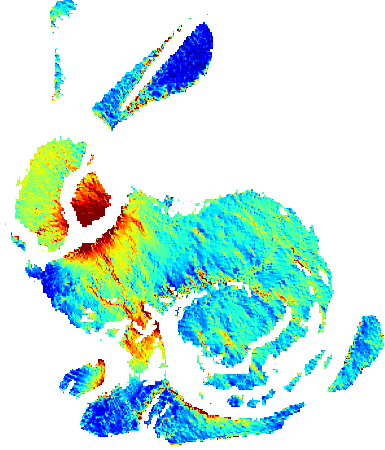}
		&
		\colorbar{\figwidthNormalVisSupp}{$10^\circ$}{0.047}
		\\
		& 4.814 & 3.900  & 5.021 & 3.981  & 5.374 & 3.859 & 6.013 & 3.908  & 7.429 & 4.443
	\end{tabular}
	}
	\caption{Estimation results with varying light-to-surface distances.
	We show the estimated surface normal with angular error maps, where the distance from the light to the object is set to $2$, $3$, $4$, $5$, and $6$ under the setting of $r=1$. }
	\label{fig:different_distance_experiment}
\end{figure*}

\pparag{Special case}
Finally, we consider the solvability condition when all lights are on the same line. 
In this case, we can estimate the surface points with $\ge 4$ lights and the known camera intrinsics, assuming that the $x-$ and $y-$ components of the light offset $\lo$ are all $0$.
With a similar rank analysis as in the general case, we can confirm that $\ge 4$ lights are needed for the scaled distance estimation. 
From the scaled distances we have an inhomogeneous constraint on $x_r'$ and $y_r'$ from \eref{eq:scaled_x}. Additionally, when $\lo = \left[0,0,s_z\right]\transp$, the ratio $y_r'/x_r'$ can be obtained using a normalized camera coordinate $\V{p}$ as $v'/u'$. From the above two constraints we can estimate $x_r'$ and $y_r'$, and $z_r'$ can also be obtained by following \eref{eq:scaled_shifted_z}. 


\section{Experiments}
In this section, we evaluate our method using both synthetic and real-world scenes. 
We first outline our experimental settings and then present the results of our evaluations.

\subsection{Experimental settings}

\pparag{Baselines}
For comparison, we evaluate our method with three state-of-the-art techniques: Calibrated~\cite{Yvain2018}, FastNFPS~\cite{lichy2022fast}, and UniversalPS~\cite{ikehata2022universal}. Calibrated is a near-light calibrated photometric stereo method that assumes Lambertian surfaces.FastNFPS and UniversalPS are both learning-based approaches. FastNFPS is a near-light uncalibrated photometric stereo method for non-Lambertian scenes, and UniversalPS is an uncalibrated method designed for both near and distant light conditions.

For Calibrated, we use the ground-truth light positions as input. Since Calibrated requires initial guesses of the depth, we use the median value of the ground-truth depth for synthetic experiments and the approximate distance between the camera and the target object for real-world experiments. Since UniversalPS only outputs normal maps, we evaluate its performance using the estimated normal maps.

\pparag{Evaluation procedures}
For quantitative evaluation in the synthetic experiment, we align the estimated depth maps from each method by scaling and shifting so that the estimated depth fits the ground truth.
This is necessary because FastNFPS and our method estimate depth maps with scale and shift ambiguities. Although Calibrated outputs absolute depth, the results can shift depending on the initialization. To ensure a fair comparison, we apply the same alignment to all methods.
We use an angular error for surface normal, and a relative absolute error for depth as evaluation metrics. Relative absolute error is computed as the ratio of raw absolute error over the ground-truth depth. 

\subsection{Evaluation with synthetic scenes}
In this section, we quantitatively evaluate the baseline method and ours on a synthetic dataset.

\pparag{Dataset}
We render synthetic scenes using two Lambertian objects, \textsc{Bunny} with spatially-varying albedo and \textsc{Crab} with uniform albedo. We use a camera whose focal length, sensor size, and resolution are set to $\SI{85}{\milli\metre}$, $36\times24~\si{\milli\metre}$, and $720\times480$, respectively. 
In the image rendering process, the physically-correct light fall-off, \ie proportional to squared distance, is included, and global illumination effects, such as cast shadows and inter-reflections, are not included. We also mask the attached shadow area only in the synthetic experiments. The rendered images are stored in single-precision.

\Fref{fig:light_arrangement_in_exp} shows the light arrangements used in our synthetic experiments with varying the number of symmetric light pairs from $2$ to $4$.
For each light arrangement, we evaluate the results with a global offset along the $z$-axis and along all three axes $x,y,z$-axis with respect to the camera's optical center. The proposed and the uncalibrated methods assume unknown offsets, while the other methods use the ground truth light positions.
The object is put $6$ distance away from the camera center, where the inner radius of the symmetric pairs is $1.$

\pparag{Results}
\Fref{fig:synth_results_z_shift} shows the estimated results of proposed and comparison methods for our synthetic scenes with the global offset along the $z$-axis.
For both scenes with different numbers of symmetric light pairs, the proposed method achieves results that are comparable to those of the Calibrated in terms of both normal and depth map estimation. 
One possible reason for the relatively large errors observed in FastNFPS and UniversalPS is the small number of light sources.
Furthermore, we observe that the Calibrated, which assumes a continuous surface, fails to accurately recover the depth map for scenes with discontinuities, such as the limbs of the \textsc{Crab}. In contrast, the proposed method, which employs per-pixel estimation, achieves better results in handling these discontinuities.

\Fref{fig:synth_results_xyz_shift} shows the estimated results with the global offset along all three axes $x,y,z$- axis. In the scenes with $\npairs=3$, although our method also achieves comparable results, it fails depth recovery in some areas, \eg right side of the \textsc{Bunny}'s face. This is because the $\npairs=3$ case is the minimal condition required to solve our homogeneous system, and the solution should be sensitive to the approximation error from light fall-off relaxation. In contrast, in the scenes with $\npairs=4$, the proposed method achieves almost similar accuracy compared to the Calibrated although the proposed method assumes an unknown global offset.

\Fref{fig:synth_depth_results_fourOnLine_z_shift} shows the estimated results for $\npairs=2$, where the proposed method can only estimate depth maps.
The proposed method outperforms Calibrated even though it uses the given light positions and initial depth, due to the difficulty in solving the non-convex optimization problem with a small number of lights.
\begin{figure}[t]
	\centering
	\includegraphics[width=0.95\linewidth]{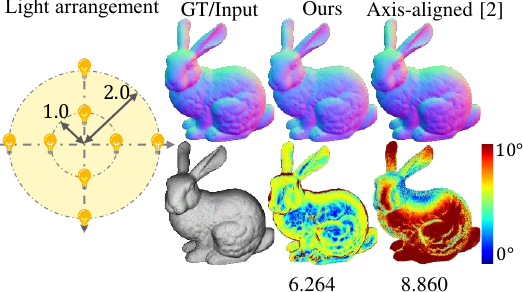}
	\caption{(Left) Axis-aligned symmetric light arrangement used in this experiment. (Right) Estimated normal maps from the proposed method~(Ours) and axis-aligned symmetric light method~\cite{Minami2022}.}
	\label{fig:compare_to_symps}
    \vspace*{-10pt}
\end{figure}

\begin{figure}[t]
	\centering
	\includegraphics[width=0.9\linewidth]{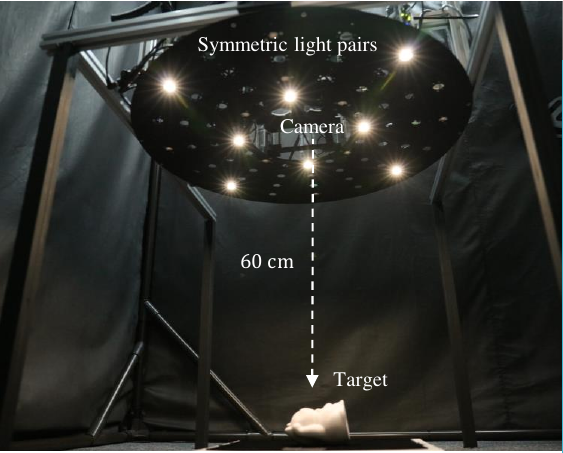}
	\caption{Our capturing setup. We put the object approximately $\SI{60}{\centi\metre}$ away from the camera. Inner and outer radii are $\SI{12.5}{\centi\metre}$ and $\SI{27.5}{\centi\metre},$ respectively.
		We use the images taken under each of the lights shown in the figure.}
	\label{fig:experimental_setting}
    \vspace*{-20pt}
\end{figure}
\begin{figure*}[h!]
	\newcommand{\figwidthNormalVisSupp}{0.13}
	\newcommand{\halfFigwidthNormalVisSupp}{0.065}
	\centering
	\begin{tabular}{@{}c@{}c@{}c@{}c@{}c@{}c@{}c@{}c@{}c@{}c@{}}
		&  \textsc{Cat} & \textsc{Sheep} & \textsc{SV-Cat} & \textsc{Sailor} & \textsc{Santa} & \textsc{Bunny} & \textsc{Buddha} & \textsc{Donut} & \textsc{Ball} 	
		\\
		\raisebox{\halfFigwidthNormalVisSupp\linewidth}{\rotatebox[origin=c]{90}{Input image}} &
		\includegraphics[height=\figwidthNormalVisSupp\linewidth]{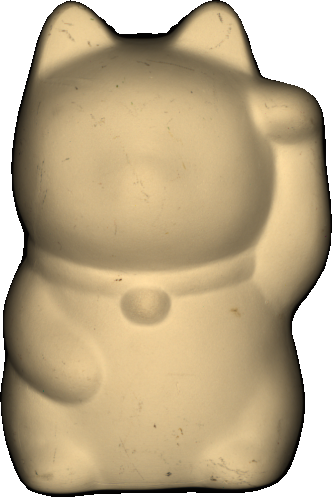} &
		\includegraphics[height=\figwidthNormalVisSupp\linewidth]{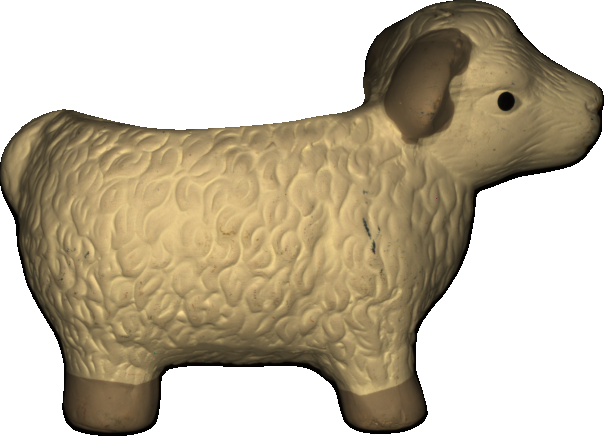} &
		\includegraphics[height=\figwidthNormalVisSupp\linewidth]{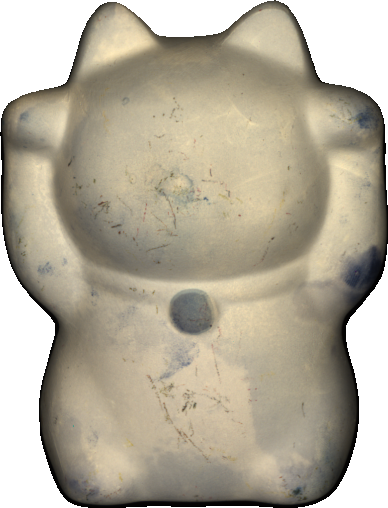}&
		\includegraphics[height=\figwidthNormalVisSupp\linewidth]{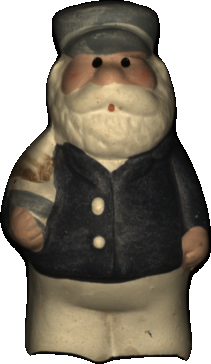}&
		\includegraphics[height=\figwidthNormalVisSupp\linewidth]{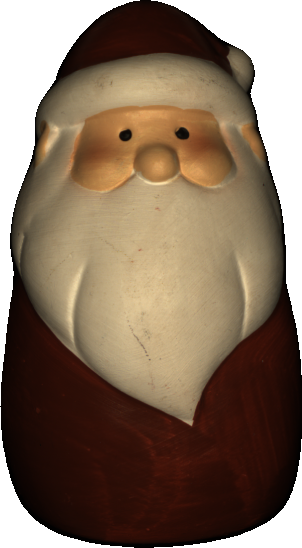}&
		\includegraphics[height=\figwidthNormalVisSupp\linewidth]{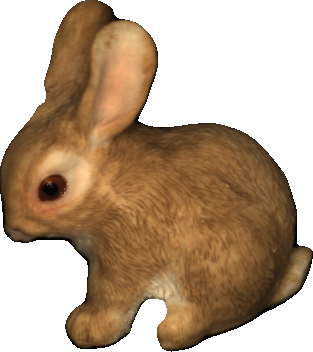}&
		\includegraphics[height=\figwidthNormalVisSupp\linewidth]{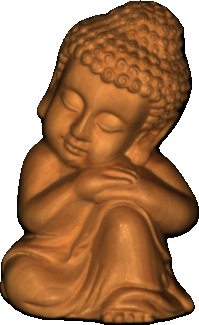}&
		\includegraphics[height=\figwidthNormalVisSupp\linewidth]{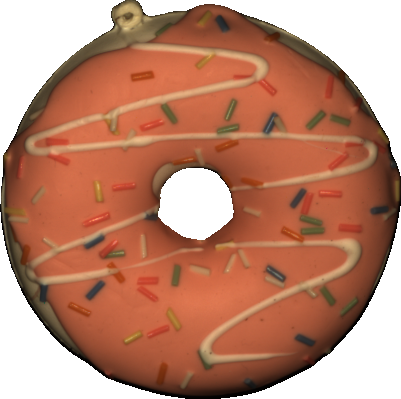}&
		\includegraphics[height=\figwidthNormalVisSupp\linewidth]{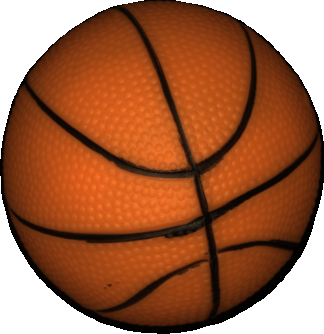}\\

		\raisebox{\halfFigwidthNormalVisSupp\linewidth}{\rotatebox[origin=c]{90}{Ours}} & 
		\includegraphics[height=\figwidthNormalVisSupp\linewidth]{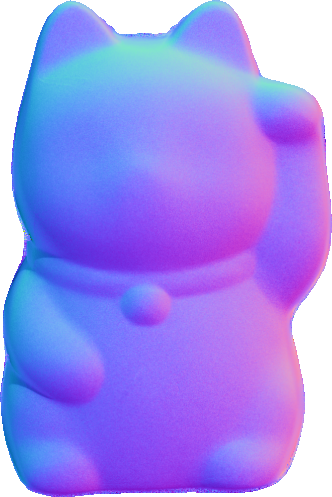} &
		\includegraphics[height=\figwidthNormalVisSupp\linewidth]{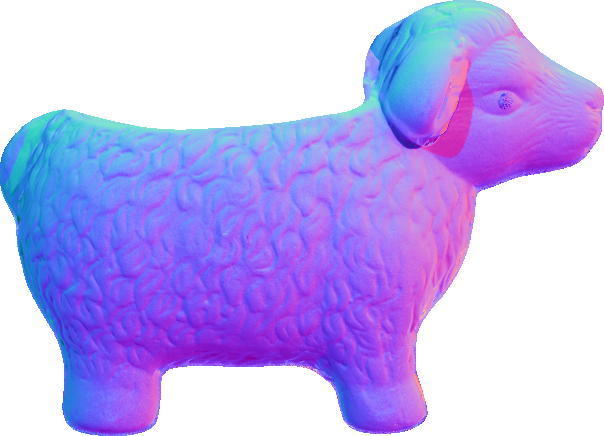} &
		\includegraphics[height=\figwidthNormalVisSupp\linewidth]{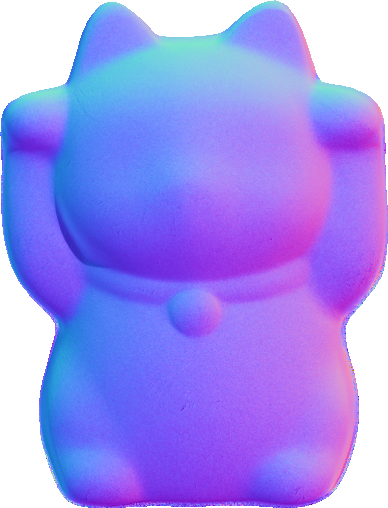}&
		\includegraphics[height=\figwidthNormalVisSupp\linewidth]{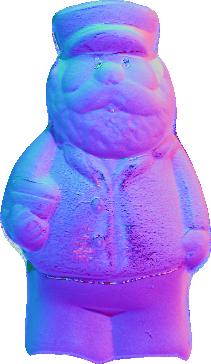}&
		\includegraphics[height=\figwidthNormalVisSupp\linewidth]{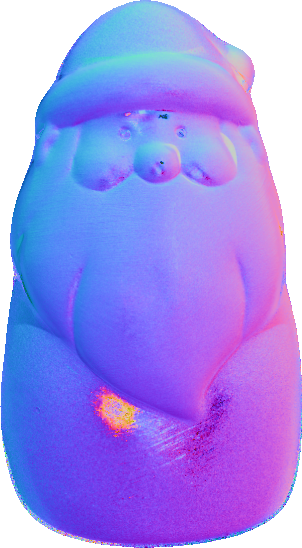}&
		\includegraphics[height=\figwidthNormalVisSupp\linewidth]{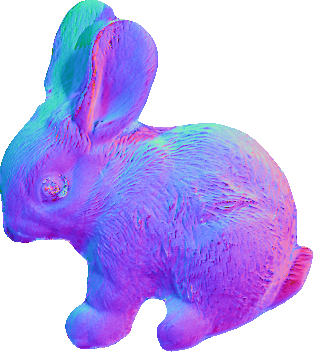}&
		\includegraphics[height=\figwidthNormalVisSupp\linewidth]{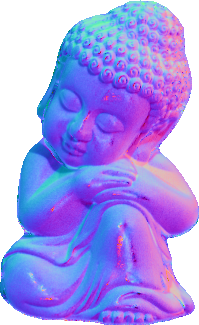}&
		\includegraphics[height=\figwidthNormalVisSupp\linewidth]{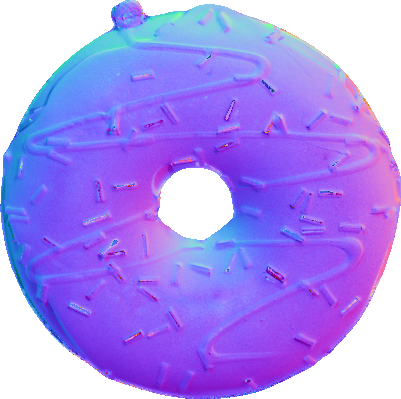}&
		\includegraphics[height=\figwidthNormalVisSupp\linewidth]{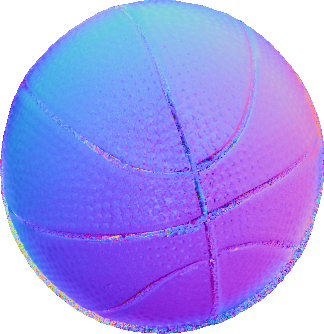}\\
		
		\raisebox{\halfFigwidthNormalVisSupp\linewidth}{\rotatebox[origin=c]{90}{Calibrated~\cite{Yvain2018}}} & 
		\includegraphics[height=\figwidthNormalVisSupp\linewidth]{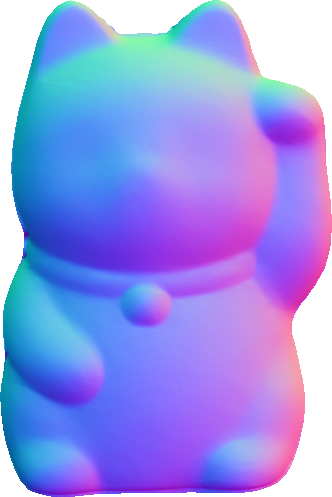} &
		\includegraphics[height=\figwidthNormalVisSupp\linewidth]{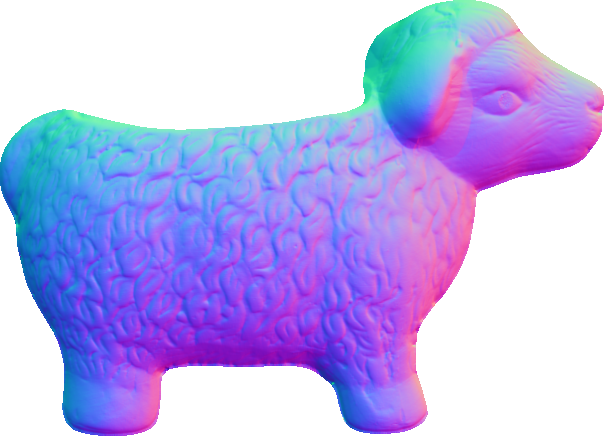} &
		\includegraphics[height=\figwidthNormalVisSupp\linewidth]{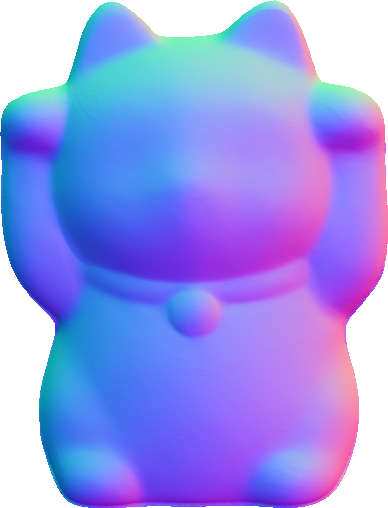}&
		\includegraphics[height=\figwidthNormalVisSupp\linewidth]{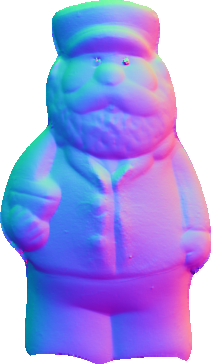}&
		\includegraphics[height=\figwidthNormalVisSupp\linewidth]{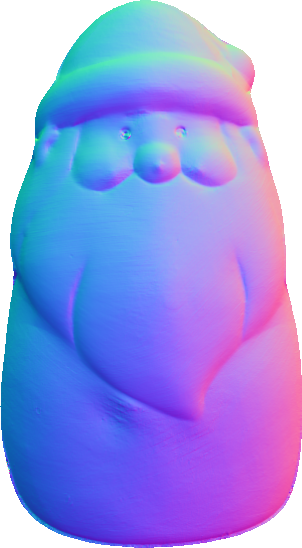}&
		\includegraphics[height=\figwidthNormalVisSupp\linewidth]{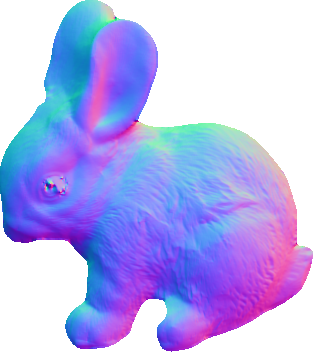}&
		\includegraphics[height=\figwidthNormalVisSupp\linewidth]{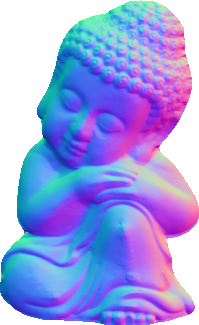}&
		\includegraphics[height=\figwidthNormalVisSupp\linewidth]{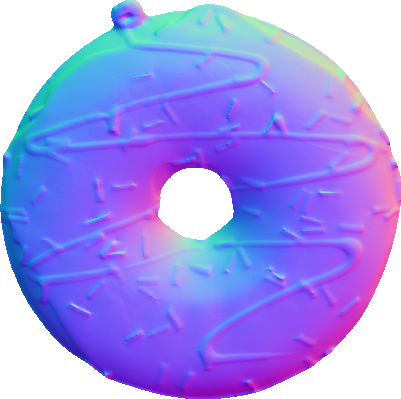}&
		\includegraphics[height=\figwidthNormalVisSupp\linewidth]{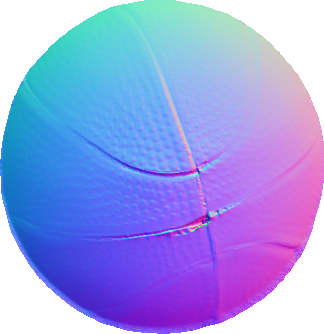}\\
		
		\raisebox{\halfFigwidthNormalVisSupp\linewidth}{\rotatebox[origin=c]{90}{FastNFPS~\cite{lichy2022fast}}} & 
		\includegraphics[height=\figwidthNormalVisSupp\linewidth]{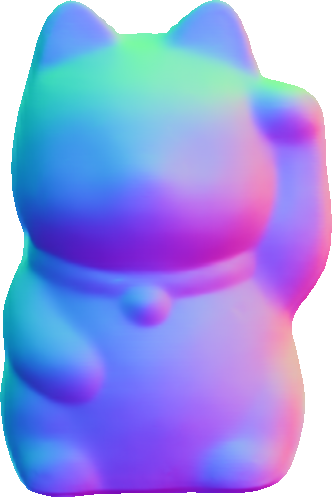} &
		\includegraphics[height=\figwidthNormalVisSupp\linewidth]{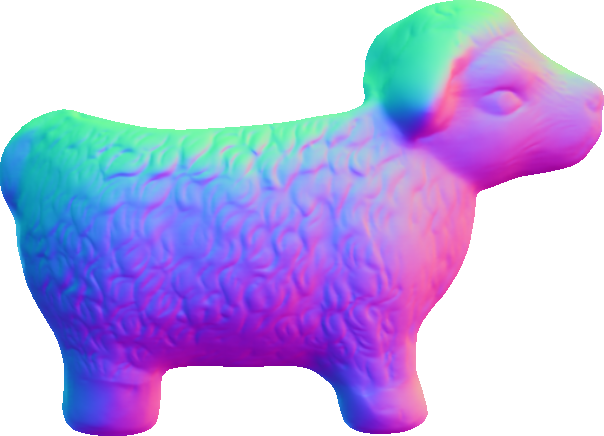} &
		\includegraphics[height=\figwidthNormalVisSupp\linewidth]{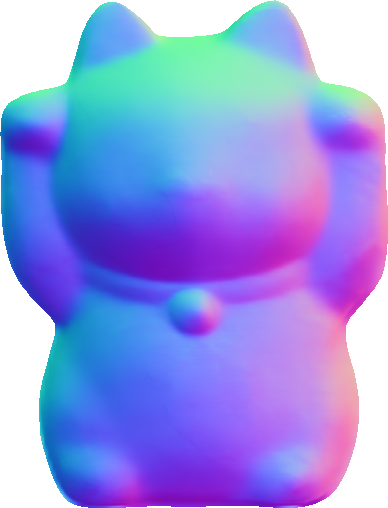}&
		\includegraphics[height=\figwidthNormalVisSupp\linewidth]{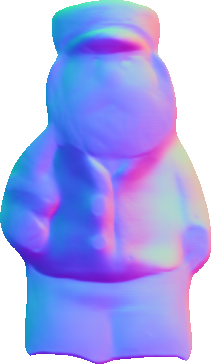}&
		\includegraphics[height=\figwidthNormalVisSupp\linewidth]{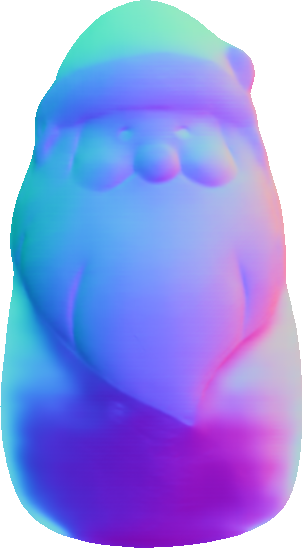}&
		\includegraphics[height=\figwidthNormalVisSupp\linewidth]{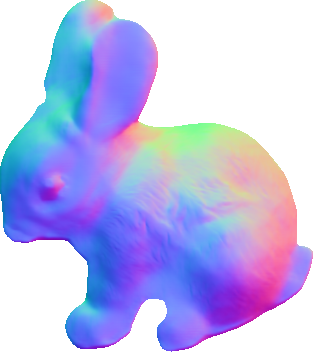}&
		\includegraphics[height=\figwidthNormalVisSupp\linewidth]{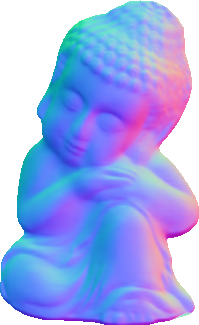}&
		\includegraphics[height=\figwidthNormalVisSupp\linewidth]{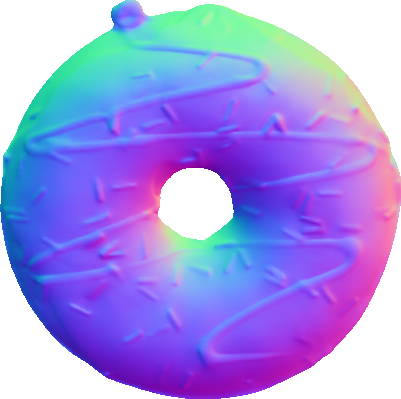}&
		\includegraphics[height=\figwidthNormalVisSupp\linewidth]{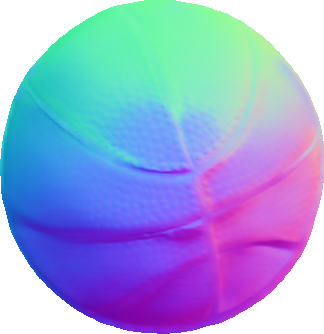}\\
		
		\raisebox{\halfFigwidthNormalVisSupp\linewidth}{\rotatebox[origin=c]{90}{UniversalPS~\cite{ikehata2022universal}}} & 
		\includegraphics[height=\figwidthNormalVisSupp\linewidth]{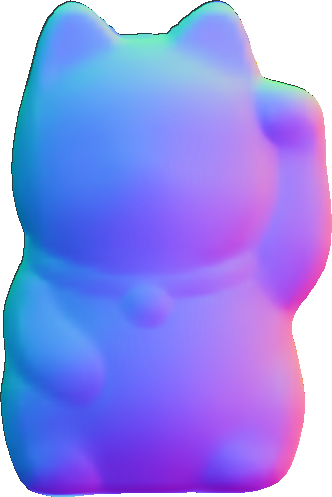} &
		\includegraphics[height=\figwidthNormalVisSupp\linewidth]{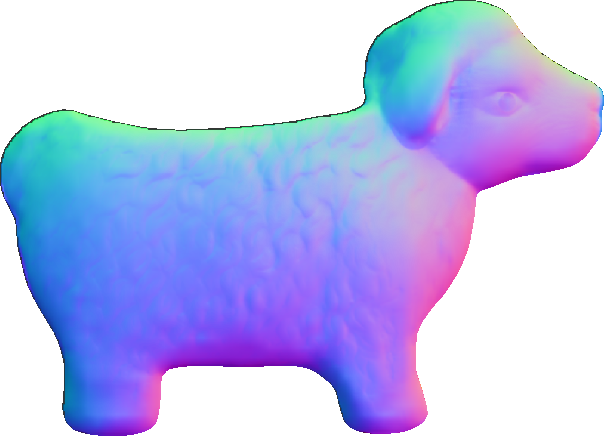} &
		\includegraphics[height=\figwidthNormalVisSupp\linewidth]{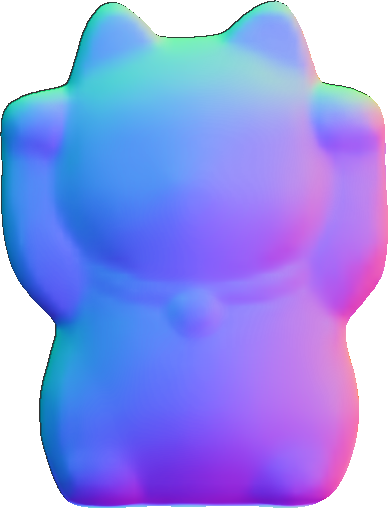}&
		\includegraphics[height=\figwidthNormalVisSupp\linewidth]{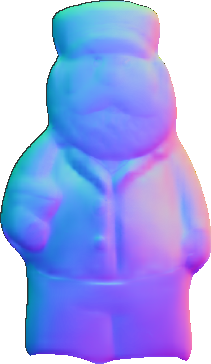}&
		\includegraphics[height=\figwidthNormalVisSupp\linewidth]{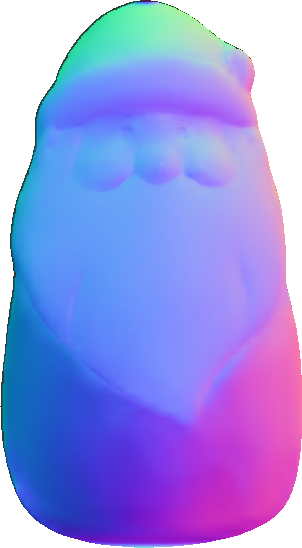}&
		\includegraphics[height=\figwidthNormalVisSupp\linewidth]{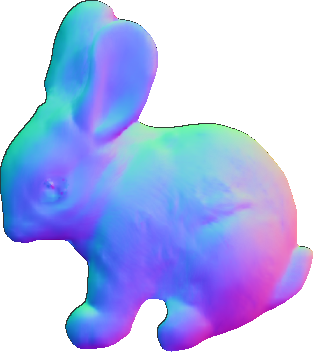}&
		\includegraphics[height=\figwidthNormalVisSupp\linewidth]{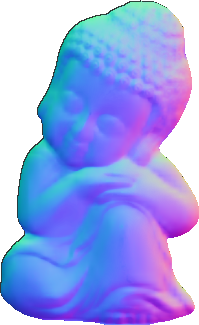}&
		\includegraphics[height=\figwidthNormalVisSupp\linewidth]{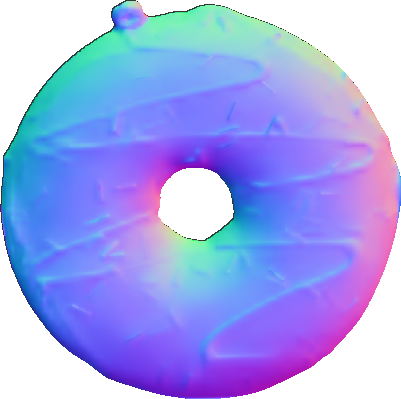}&
		\includegraphics[height=\figwidthNormalVisSupp\linewidth]{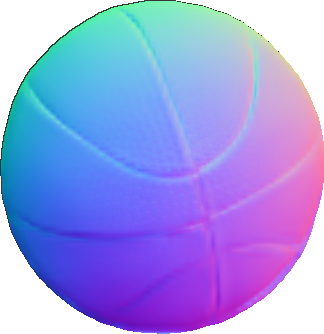}\\
			
	\end{tabular}
	\caption{Estimated normal maps for our real-world dataset. For each scene, we show one of the input images and estimate normal maps by the proposed method~(Ours) and the comparison methods.
	}
	\label{fig:realworld_experiments}
\end{figure*}

\begin{figure*}[h!]
    \scriptsize
\vspace{-5pt}
\newcommand{\figwidthNormalVisSupp}{0.098}
\newcommand{\halfFigwidthNormalVisSupp}{0.045}
\newcommand{\figSize}{0.098}
\definecolor{lightGray}{gray}{0.6}
\centering
\begin{tabular}{@{}cc@{}c@{}c@{}c@{}c@{}c@{}|@{}
		c@{}c@{}c@{}c@{}c@{}c@{}	
		}
	& \multicolumn{5}{c}{\bf{Surface normal}} & & & \multicolumn{5}{c}{\bf{Depth}}  \\
	& 
	GT/Input & Ours & Calibrated~\cite{Yvain2018} & fastNFPS~\cite{lichy2022fast} & UniversalPS~\cite{ikehata2022universal} & &&  GT & Ours & Calibrated~\cite{Yvain2018} & fastNFPS~\cite{lichy2022fast} & UniversalPS~\cite{ikehata2022universal} 
	\\
	\raisebox{\halfFigwidthNormalVisSupp\linewidth}{\rotatebox[origin=c]{90}{Estimated}}
	&
	\includegraphics[width=\figSize\linewidth]{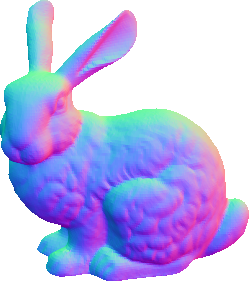}
	&
	\includegraphics[width=\figSize\linewidth]{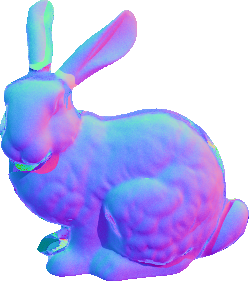}
	&
	\includegraphics[width=\figSize\linewidth]{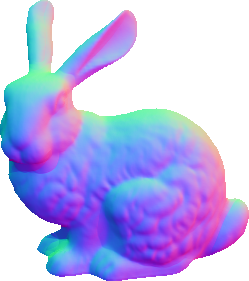}
	&
	\includegraphics[width=\figSize\linewidth]{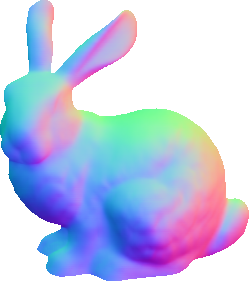}
	&
	\includegraphics[width=\figSize\linewidth]{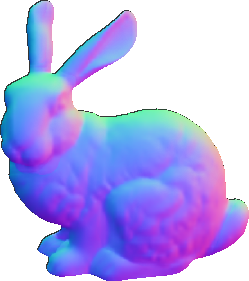}
	&
	&
	&
	\includegraphics[width=\figSize\linewidth]{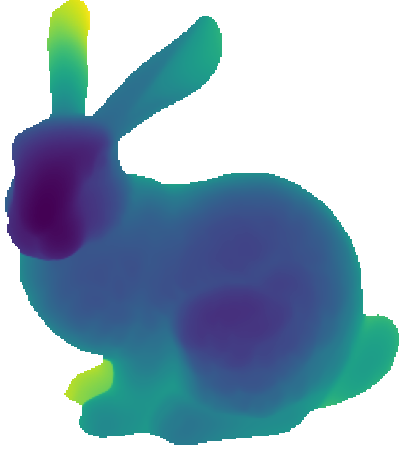}
	&
	\includegraphics[width=\figSize\linewidth]{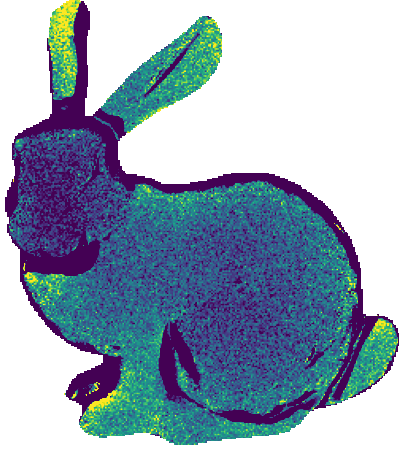}
	&
	\includegraphics[width=\figSize\linewidth]{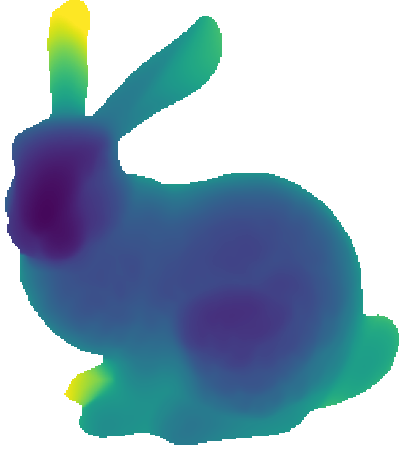}
	&
	\includegraphics[width=\figSize\linewidth]{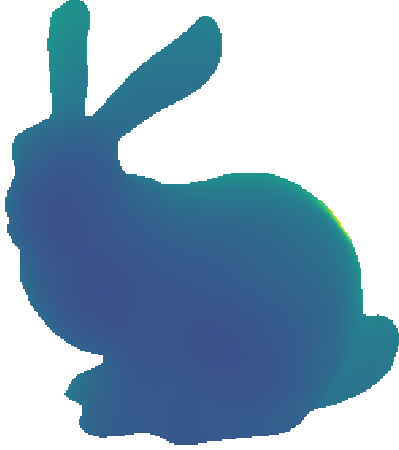}
	& 
	\multirow{2}{*}{
		\hspace{5pt}
		\rotatebox[origin=c]{-10}{\textcolor{lightGray}{\rule{.1pt}{20pt}}}
	}
	\\
	\raisebox{\halfFigwidthNormalVisSupp\linewidth}{\rotatebox[origin=c]{90}{Error map}}
	&
	\includegraphics[width=\figSize\linewidth]{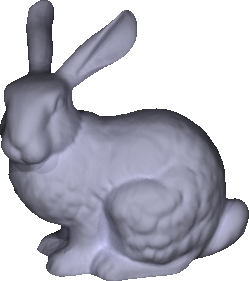}
	&
	\includegraphics[width=\figSize\linewidth]{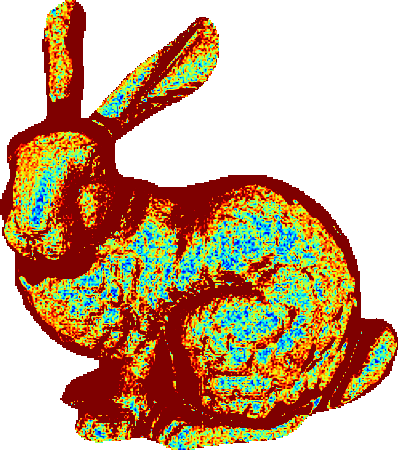}
	&
	\includegraphics[width=\figSize\linewidth]{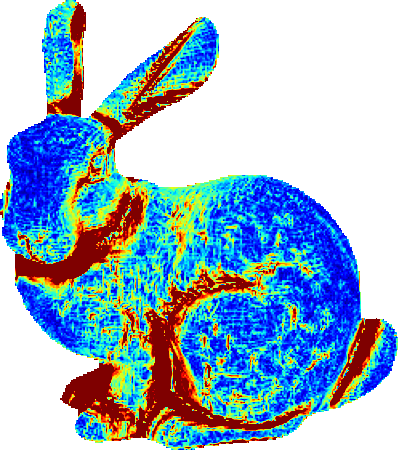}
	&
	\includegraphics[width=\figSize\linewidth]{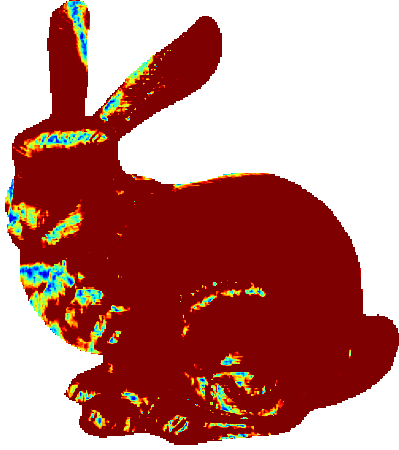}
	&
	\includegraphics[width=\figSize\linewidth]{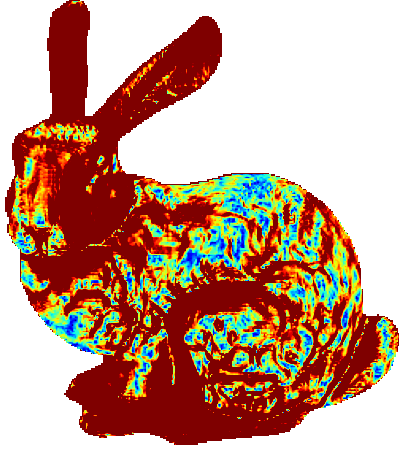}
	& & 
	&
		\includegraphics[width=\figSize\linewidth]{figures/results/synth_mimic_real/sample_image.png}
	&
	\includegraphics[width=\figSize\linewidth]{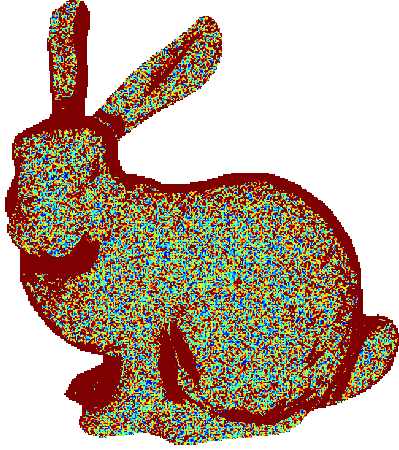}
	&
	\includegraphics[width=\figSize\linewidth]{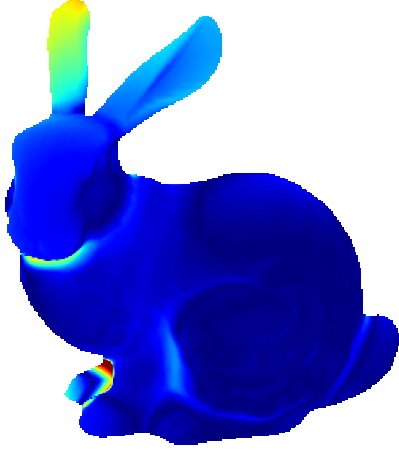}
	&
	\includegraphics[width=\figSize\linewidth]{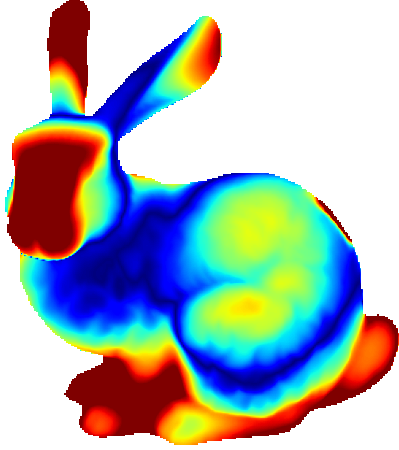}
	\\
	\vspace*{-10pt}
	& 
	& 
	\begin{tabular}{c}
		15.532 \\ ~/ 9.706
	\end{tabular} & 
	\begin{tabular}{c}
		6.034  \\ ~/ 4.452
	\end{tabular} & 
	\begin{tabular}{c}
		22.095\\ ~/ 23.127
	\end{tabular} & 
	\begin{tabular}{c}
		14.240 \\ ~/ 12.084 
	\end{tabular} 
	&  &  & &
	\begin{tabular}{c}
		25.184 \\ ~/ 5.411 
	\end{tabular} & 
	\begin{tabular}{c}
		0.549    \\ ~/  0.470
	\end{tabular} & 
	\begin{tabular}{c}
		3.306 \\ ~/ 4.282
	\end{tabular} & 
	\begin{tabular}{c}
		INVALID
	\end{tabular}
	\vspace*{15pt}
	\\
	&& \multicolumn{3}{c}{\rotatedColorbar{Normal error}{0.08}{$10^\circ$}{-20}} 

	&&&&&
	\multicolumn{3}{c}{\rotatedColorbar{Depth error}{0.08}{$0.01$}{-20}} & 	
\end{tabular}

    \caption{Estimation results of our method and comparison methods on our synthetic scenes, which are rendered under the realistic image setting described in \sref{sec:noisy_setting}. We show a mean angular~(for normal)/relative absolute errors~(for depth) at the bottom of each error map. The two values separated by ``/'' indicate the errors without and with masking the pixels where our depth estimation errors are larger than $0.01$.}
    \vspace*{-10pt}
    \label{fig:synth_results_mimic_real}
    \noindent
\end{figure*}
\subsection{Effect of varying light-to-surface distance}

The light fall-off relaxation may fail when the light-to-surface distance is too short.
We evaluate the effect of varying light-to-surface distances using synthetic scenes.
In this experiment, we use $\npairs=4$ lights as shown in \fref{fig:light_arrangement_in_exp}, and we change the positions of light sources along the $z$-axis.
\Fref{fig:different_distance_experiment} shows the estimated results.
While the shorter cases have a larger error, we can obtain comparable results when we use approximately $l \ge 4$, where $l$ is the distance from the light sources to the target object.
This result suggests that, for example, if we use the radius of the inner symmetric light pair $s \approx \SI{12.5}{\centi\metre}$, we should put the target object more than $\SI{50}{\centi\metre}$ far away from the lights,
which is comparable to the conclusion in the supplementary material.
The error by the relaxation is attributed to the difference in the input images, \iow the difference in the light fall-off terms. To assess the further analysis of relaxation-accuracy relations, we evaluate the relative error of the light fall-off terms in the supplementary material. Please refer to it for more details.

\subsection{Comparison with axis-aligned method}

We further evaluate the estimation results of our method and the axis-aligned symmetric light method~\cite{Minami2022}.
The light arrangement used in this experiment and the estimated normal maps by both methods are shown in \fref{fig:compare_to_symps}. 
The proposed method outperforms the axis-aligned symmetric light method, which assumes distant lights, due to its explicit consideration of the near-light effects.

\subsection{Noise/Shadow effect}
	To evaluate the method in a more realistic setting, we render the images with settings as close as possible to real-world experiments, with quantization of $12$ bit unsigned integer same as the real-world setting, global illumination effects enabled including cast shadows and inter-reflections, and a shot and readout noise following existing work~\cite{cao2023physics} as synthetic noises.
    The parameters for shot noise and the standard deviation for readout noise are fitted using real-world images captured by the same camera used in real-world experiments.
	\Fref{fig:synth_results_mimic_real} shows the estimated results with noisy images captured under $\npairs=4$ symmetric lights. Our surface normal estimates outperform the uncalibrated PS methods, in most of the area. The depth map from fastNFPS has a flattened shape, while ours can estimate a reasonable shape except for the boundary area. 
	\label{sec:noisy_setting}
\subsection{Experiment with real-world scenes}
In this section, we show evaluations using our real-world dataset.
To capture the dataset, we use the device shown in \Fref{fig:experimental_setting}, which consists of a CCD camera\footnote{BFS-U3-28S5C-C, Teledyne FLIR LLC} equipped with $\SI{25}{\milli\metre}$ lens and eight LEDs\footnote{XLamp CXA1304, Cree, Inc.}.
The LEDs are mounted on a board precisely manufactured by a CNC machine, ensuring the accurate placement of each LED. We employ a constant current circuit in our device for all light sources to emit uniform radiance. For further detail, please refer to the supplementary material.
Since the camera is fixed to the board via rigs, it is difficult to align the optical center of the symmetric light pairs with the camera. Therefore, in this experiment, the proposed method assumes the existence of an unknown global offset and solves for it using four symmetric light pairs.
The camera's intrinsics are calibrated using OpenCV\footnote{OpenCV 4.5, https://opencv.org, Retrieved Apr. 14, 2023.} for comparison methods.
For the target objects, we use nine objects with an almost diffuse surface.

\pparag{Results}
\label{para:realworld_results}
\Fref{fig:realworld_experiments} shows the estimated normal maps by the comparison and proposed methods.
Overall, the proposed method achieves almost comparable results to Calibrated, even though the proposed method does not use the exact light positions. Both the learning-based results in over-smoothed results for most of the objects. 
Both the Calibrated and our method output reasonable shape estimations for almost pure Lambertian scenes, such as \textsc{Cat} and \textsc{Sheep}, and, for the other scenes, both the Calibrated and ours are affected by the specular observations due to the deviation from the Lambertian assumption. For example, in \textsc{Santa}'s lower body, we can see that both methods failed to recover the correct shape.
Other than \textsc{Cat}, the surfaces contain spatially-varying albedo, and the proposed method can naturally handle it because of our per-pixel formulation.
Comparing the proposed and comparison methods, we can see that the proposed method is affected by cast shadows more than the calibrated method, for example, around the \textsc{Rabbit}'s ear part.
This is because the proposed method computes the differences of symmetric pair observations, which is sensitive to the pair that one observation is shadowed. 
The results of the proposed method are slightly noisy because of the per-pixel estimation. This can be improved by considering the smoothness across the neighboring pixels.

\section{Conclusion}
We have proposed a near-light photometric stereo method using origin symmetric lights.
Unlike the previous studies dealing with non-convex optimization problems,
we show the symmetric arrangements along with the light fall-off relaxation casts the near-light photometric stereo problem to a per-pixel linear estimation, which results in a closed-form solution for surface normal and depth recovery that is globally optimal under light fall-off relaxation.
The experiments show that our method enables comparable results to the state-of-the-art calibrated near-light photometric stereo method, without explicit calibration or initial depth input.

\ifpeerreview \else
\section*{Acknowledgments}
This work was supported by JSPS KAKENHI Grant Number JP23H05491.
\fi

\bibliographystyle{IEEEtran}
\bibliography{ref}

\ifpeerreview \else





\begin{IEEEbiography}
    [{\includegraphics[width=1in,height=1.25in,clip,keepaspectratio]{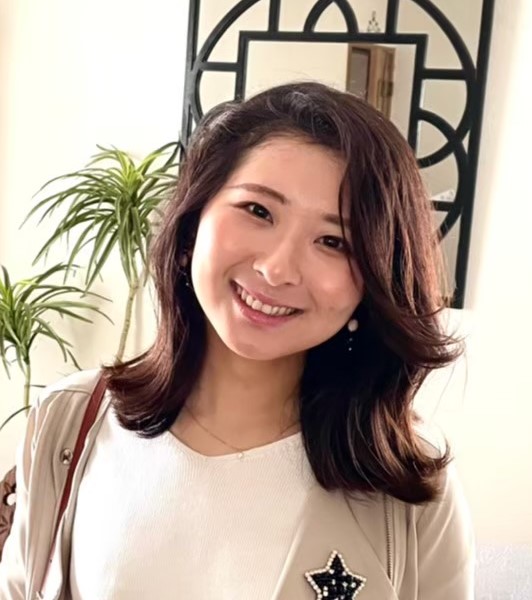}}]
    {Lilika Makabe} received her B.S. and M.S. degrees in computer science from Osaka University, Japan, in 2020 and 2022, respectively, where she is currently working toward a Ph.D degree. Her research interests include computational photography and physics-based vision.
    \end{IEEEbiography}
        
\begin{IEEEbiography}[{\includegraphics[width=1in,height=1.25in,clip,keepaspectratio]{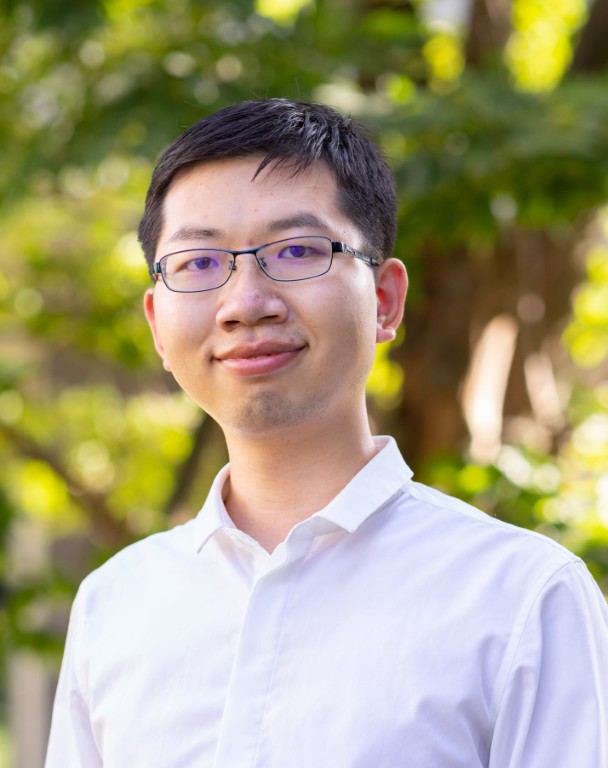}}]{Heng Guo}
	received his B.E. and M.S. degrees in signal and information processing from University of Electronic Science and Technology of China, and Ph.D. degree from Osaka University, in 2015, 2018, and 2022. He is currently a specially-appointed assistant professor at Osaka University. His research interests include physics-based vision and machine learning.
\end{IEEEbiography}

\begin{IEEEbiography}[{\includegraphics[width=1in,height=1.25in,clip,keepaspectratio]{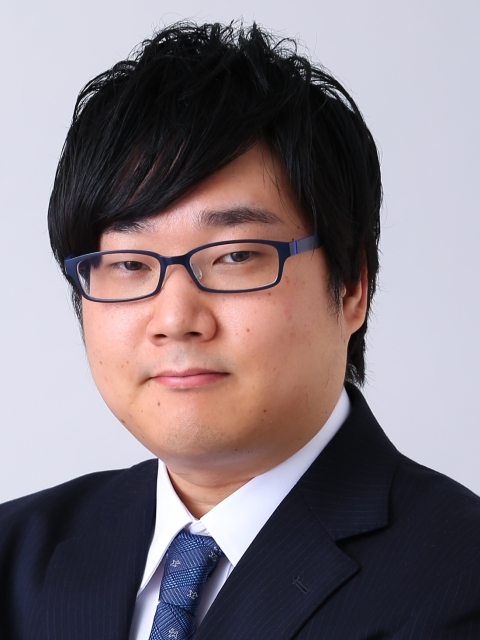}}]{Hiroaki Santo} received his M.S. and Ph.D. degrees in information science from Osaka University, Japan, in 2018 and 2021, respectively. He is currently an assistant professor with the Department of Multimedia Engineering, Graduate School of Information Science and Technology, Osaka University. His research interests include computer vision and machine learning.
\end{IEEEbiography}
   
\begin{IEEEbiography}
    [{\includegraphics[width=1in,height=1.25in,clip,keepaspectratio]{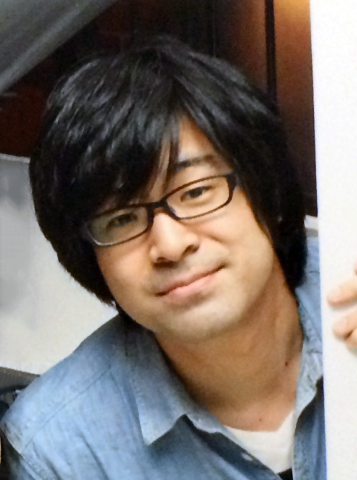}}]
    {Fumio Okura} received the M.S. and Ph.D. degrees in engineering from the Nara Institute of Science and Technology, in 2011 and 2014, respectively. He has been an assistant professor with the Institute of Scientific and Industrial Research, Osaka University, until 2020. He is now an associate professor with the Graduate School of Information Science and Technology, Osaka University. His research interest includes the boundary domain between computer vision and computer graphics.
\end{IEEEbiography}

\begin{IEEEbiography}
    [{\includegraphics[width=1in,height=1.25in,clip,keepaspectratio]{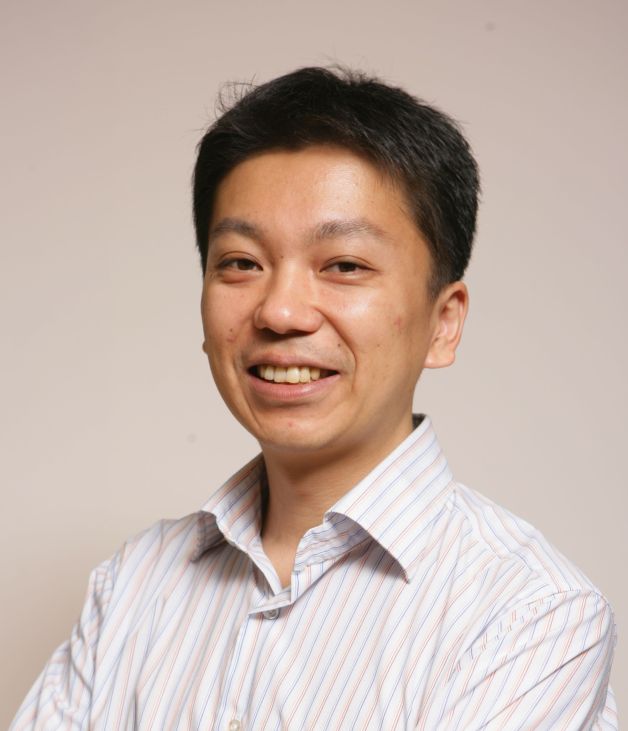}}]
    {Yasuyuki Matsushita} received his B.S., M.S. and Ph.D. degrees in EECS from the University of Tokyo in 1998, 2000, and 2003, respectively. From April 2003 to March 2015, he was with Visual Computing group at Microsoft Research Asia. In April 2015, he joined Osaka University as a professor. His research area includes computer vision, machine learning and optimization. He is/was an Editor-in-Chief for International Journal of Computer Vision and on the editorial board of IEEE Transactions on Pattern Analysis and Machine Intelligence (TPAMI), The Visual Computer journal, IPSJ Transactions on Computer Vision Applications (CVA), and Encyclopedia of Computer Vision. He served/is serving as a Program Co-Chair of PSIVT 2010, 3DIMPVT 2011, ACCV 2012, ICCV 2017, and a General Co-Chair for ACCV 2014 and ICCV 2021. He is a senior member of IEEE.
\end{IEEEbiography}


\fi

\end{document}